\documentclass[a4paper,fleqn]{cas-dc}

\usepackage[numbers]{natbib}

\usepackage{caption}
\usepackage{amsmath, amssymb} 
\usepackage{graphicx}
\usepackage{float}
\usepackage{subcaption}

\usepackage[ruled,linesnumbered]{algorithm2e}
\usepackage{algpseudocode}
\usepackage{xcolor}
\usepackage{etoolbox}
\usepackage{multirow}
\usepackage{hyperref}
\usepackage{xr}
\usepackage{color}
\usepackage{comment}

\renewcommand{\figurename}{Fig.}
\def\tsc#1{\csdef{#1}{\textsc{\lowercase{#1}}\xspace}}
\tsc{WGM}
\tsc{QE}

\begin{document}
\let\WriteBookmarks\relax
\def\floatpagepagefraction{1}
\def\textpagefraction{.001}

\shorttitle{}

\shortauthors{}

\title [mode = title]{HPSO: Particle Swarm Optimization with Hypergraph-Based Topology}

\author[1,2]{Wenbin~Pei}[orcid=0000-0002-8259-2614]

\cormark[1]

\ead{peiwenbin@dlut.edu.cn}

\affiliation[1]{organization={School of Computer Science and Technology, Dalian University of Technology},
            city={Dalian},
            postcode={116024}, 
            country={China}}

\affiliation[2]{organization={Key Laboratory of Social Computing and Cognitive Intelligence (Dalian University of Technology), Ministry of Education},
            country={China}}

\author[1,2]{Xi~Luo}

\author[3]{Bing~Xue}

\affiliation[3]{organization={Centre for Data Science and Artificial Intelligence \& School of Engineering and Computer Science, Victoria University of Wellington},
            city={Wellington},
            postcode={6140}, 
            country={New Zealand}}

\author[3]{Mengjie~Zhang}

\author[1,2,4]{Qiang~Zhang}[orcid=0000-0003-0609-0337]

\cormark[1]

\ead{zhangq26@126.com}

\affiliation[4]{organization={National and Local Joint Engineering Laboratory of Computer Aided Design, Dalian University},
            city={Dalian},
            postcode={116622}, 
            country={China}}

\begin{abstract}
Particle swarm optimization (PSO) has been widely applied to solve complex optimization problems from real-world applications due to its efficient exploration of large solution spaces and the ability to converge towards optimal solutions without requiring gradient information. Common swarm topologies in standard PSO and its variants, e.g., Ring and Star, can be regarded as graphs, where each edge connects only two particles.
Such topology structures allow direct interactions only between connected particle pairs, and thus often fail to directly capture the higher-order social relationships that are necessary for navigating complex search landscapes. Therefore, this article proposes a novel PSO variant termed Hypergraph-assisted Particle Swarm Optimization (HPSO). In HPSO, the topology of the particles in a swarm is modeled by a hypergraph, in which hyperedges are used to connect multiple particles. This allows multiple particles within a hyperedge to interact directly. Furthermore, an adaptive hypergraph updating strategy is designed to periodically reconstruct the topology based on cumulative average particle displacement, thereby maintaining swarm diversity throughout the evolutionary process. In the experiments, the effectiveness of HPSO is verified on the IEEE CEC'17 benchmark suite, and the results demonstrate that HPSO achieves promising performance across various types of functions. Furthermore, the ablation experiment demonstrates that HPSO has excellent search capabilities.
\end{abstract}

\begin{keywords}
 Optimization \sep Particle Swarm Optimization \sep Hypergraph Topology \sep Adaptive Hypergraph Update.  
\end{keywords}

\maketitle

\section{Introduction}

Particle swarm optimization (PSO) has been acknowledged as one of the most prominent optimization algorithms, drawing inspiration from the collective social behavior observed in biological populations, such as bird flocks and fish schools \cite{Gad2022}. 
Owing to its computational efficiency and rapid convergence toward promising regions of the search space without requiring gradient information \cite{HMPPSO}, PSO has been successfully introduced to resolve intricate engineering tasks across a wide range of domains \cite{PSO-HV}, including power systems \cite{power_system, power_generation_system}, structural design \cite{structural_design}, and signal processing \cite{signal_processing}. However, the use of PSO to solve increasingly complex real-world engineering problems presents new challenges, such as exponentially increasing search space, dynamic environments, and the need to balance exploration with exploitation \cite{EAPSO}. To address these challenges, extensive literature  have demonstrated that swarm topology is a critical factor influencing the search performance of PSO \cite{ImpactCommunicationTopology2019a,pop_topologies_for_pso,population_structure, FIPS}.

\begin{figure}
\centering
\includegraphics[scale=0.3]{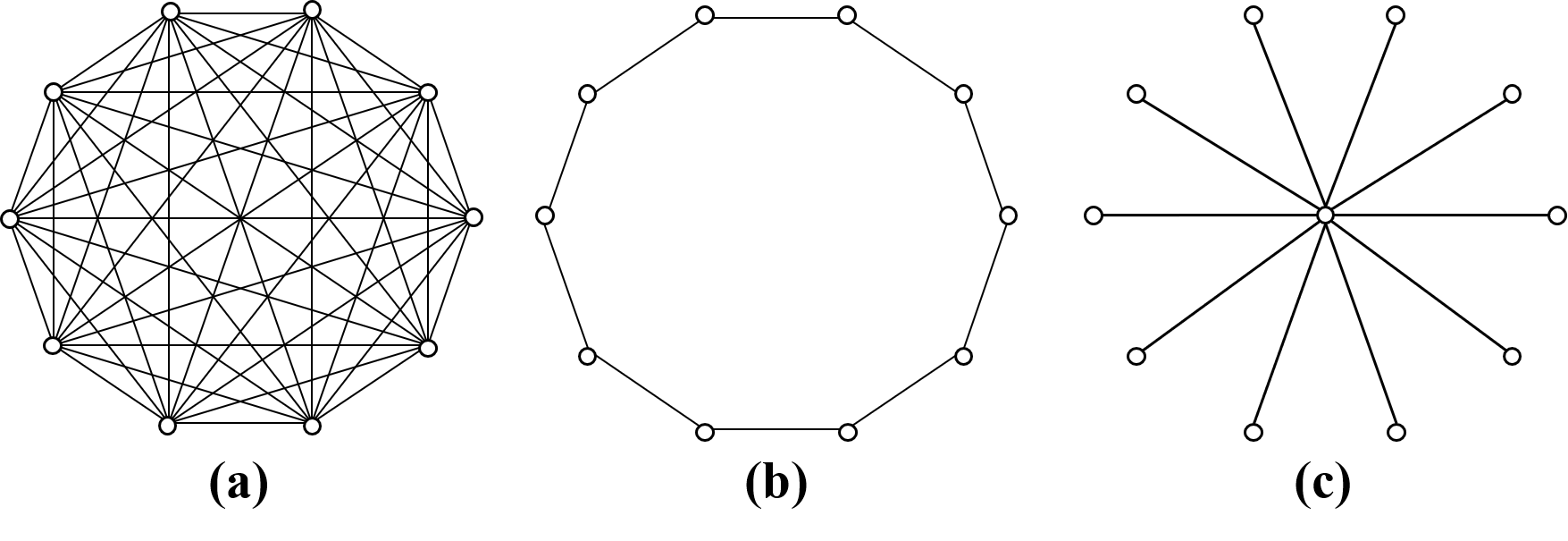}
\caption{Common population topology networks include: (a) Fully, all particles are fully connected; (b) Ring, each particle is connected to only two other particles; (c) Star, each particle interacts with others through one central particle.}
\label{fig1}
\end{figure}

As a representative swarm intelligence algorithm inspired by the self-organizing behaviors of social groups, PSO exhibits inherent topological characteristics, as the movement of each particle is influenced by others \cite{BDTPSO}. These interactions among particles enable the construction of a network structure that represents the swarm topology. 
Fig. \ref{fig1} illustrates common topologies in PSO.
As can be seen from Fig. \ref{fig1}, these topologies can be conceptualized as a graph, where a node corresponds to a particle and an edge directly connects two nodes. Unfortunately, such graph-based structures often fail to directly account for the higher-order social relationships that are essential for navigating complex search landscapes.

In PSO, effectively exploring complex search spaces and searching for the global optimum demands a higher degree of collaboration, in which multiple particles are expected to interact directly and jointly drive the search process. Different from graphs, a hypergraph generalizes the graph structure by allowing a single hyperedge to connect multiple nodes directly \cite{HypergraphLearningMethods2022}. Therefore, hypergraphs have the potential for modeling higher-order interactions among particles. However, there has not been a systematic investigation of how a hypergraph can model such higher-order interactions within a swarm.

This study proposes a novel PSO variant called Hypergraph-assisted PSO (HPSO). In HPSO, a hypergraph-based topology is established to capture higher-order relationships among particles, allowing each particle to form tight hyperedge-based connections with its neighbors. Each particle learns from other neighbors within the same hyperedge. Furthermore, since hypergraphs have been shown to fundamentally alter the dynamics of social contagion \cite{Iacopini2019, human_behaviour}, higher-order relationships can effectively accelerate information propagation, thereby facilitating faster convergence. In HPSO, the hypergraph structure is adaptively adjusted to maintain its consistency with the evolving distribution of solutions within a swarm across iterations. The main contributions of this study can be summarized as follows:

\begin{enumerate}

  \item We propose a novel hypergraph-based swarm topology for PSO, which captures higher-order relationships. This enables particles to directly learn from multiple exemplars at the dimensional level. Furthermore, an adaptive mechanism for hypergraph update is designed to effectively guide the search process while sustaining swarm diversity.

   \item We design a hypergraph-based learning strategy, where the social learning paradigm is converted from a vectorized format to a matrix-based approach. The proposed learning strategy enables particles to simultaneously draw information from multiple exemplars.  

  \item We present HPSO, a novel PSO variant that demonstrates superiority on the CEC'17 benchmark suite. Moreover, it exhibits a strong ability to escape local optima in the late evolutionary stages.
\end{enumerate}

The remainder of this article is organized as follows. Section II introduces hypergraph theory, PSO topologies, and related work. Section III describes the proposed HPSO method. Section IV presents the baseline methods, parameter settings, and benchmark suites in the experiments. Section V reports and discusses the experimental results. Finally, Section VI concludes the paper and suggests potential future research directions. 

\section{Background}
\subsection{Hypergraphs}

\begin{figure}
\centering
\includegraphics[scale=0.6]{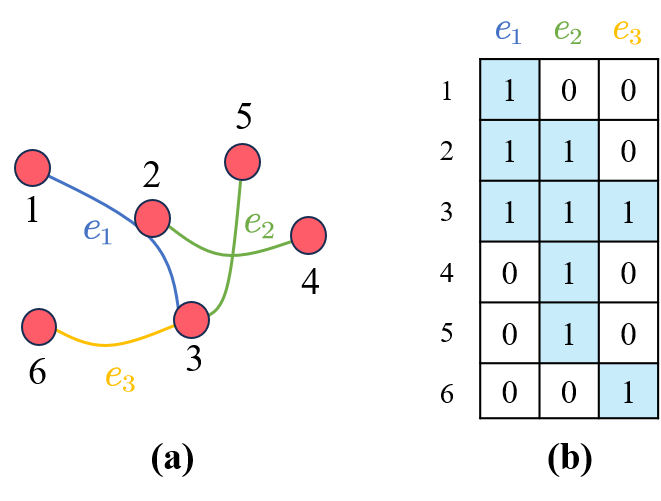}
\caption{An example of a hypergraph. (a) A hypergraph with 6 nodes and 3 hyperedges, where each node (indicated by the red dot) can be regarded as a particle, and the hyperedges represent the higher-order relationships among these nodes. (b) The corresponding incidence matrix of (a).}
\label{fig_hg}
\end{figure}

A hypergraph is a generalization of a graph, where a hyperedge can join multiple vertices simultaneously. This structure enables the modeling of complex, multi-way interactions that extend beyond the pairwise relationships typically captured by standard graphs \cite{l1Hypergraph, HypergraphComputation}. A hypergraph can be denoted as $ G\left( \mathbb{V},\ \mathbb{E},\ W \right) $, where $\mathbb{V}=\left\{ p_1,\ p_2,\ \cdots \ ,p_n \right\}$ is the set of vertices, $\mathbb{E}=\left\{ e_1,\ e_2,\ \cdots \ ,e_n \right\}$ is the set of hyperedges, and $W$ denotes the weight of each hyperedge.

A hypergraph $G$ can be represented by a vertex–edge incidence matrix $\mathbf{H}\in R^{|\mathbb{V}|\times |\mathbb{E}|}$. $\mathbf{H}$ is a binary matrix, where each row represents whether a vertex $p_i$ is incident to a hyperedge $e_j$. In particular, if $\mathbf{H}\left( p_i,\ e_j \right)=1$, it indicates $p_i\in e_j$, otherwise $p_i\notin e_j$. The matrix $\mathbf{H}$ can be defined as follows:
\begin{align}
\mathbf{H}\left( p_j,\ e_i \right) =\left\{ \begin{array}{l}
	1,\ \mathrm{if}\ p_j\in e_i\\
	0,\ \mathrm{otherwise},\\
\end{array} \right. 
\label{eql1H}
\end{align}
To understand the hypergraph intuitively, Fig. \ref{fig_hg} presents a simple hypergraph and the corresponding vertex–edge incidence matrix $\mathbf{H}$. In Fig. \ref{fig_hg}, node 3 belongs to hyperedges $e_1$, $e_2$, and $e_3$. Hence, the values of $\mathbf{H}(3, 1)$, $\mathbf{H}(3, 2)$, and $\mathbf{H}(3, 3)$ are 1.
\subsection{Introduction to PSO}
PSO \cite{PSO} \cite{PSO_theory} is a population-based metaheuristic algorithm inspired by the collective social behavior of bird flocks and fish schools. In PSO, at iteration $t$, a swarm consists of $N$ particles, each representing a candidate solution characterized by a position vector (i.e., $\vec{X_{i}^{t}}=\left\{ x_{i,1}^{t},\ x_{i,2}^{t},\ \cdots \ ,x_{i,D}^{t} \right\}$) and a velocity vector (i.e., $\vec{V_{i}^{t}}=\left\{ v_{i,1}^{t},\ v_{i,2}^{t},\ \cdots \ ,v_{i,D}^{t} \right\}$). 
At each iteration, each particle $i$ maintains a personal best position $\vec{pbest_i}$ (it represents the best solution it has discovered so far) and global best position $\vec{gbest_i}$ (it represents the best solution found by the entire swarm or its topological neighborhood). 
Subsequently, $\vec{X_{i}^{t}}$ and $\vec{V_{i}^{t}}$ are updated according to the following rules:

\begin{align}
    \begin{split}
        v_{i,d}^{t} =& w\times v_{i,d}^{t-1}+c_1\times rand_{1,d}\times \left( pbest_{i,d}^{t-1}-x_{i,d}^{t-1} \right) \\
        &+c_2\times rand_{2,d}\times \left( gbest_{i,d}^{t-1}-x_{i,d}^{t-1} \right) 
    \end{split} \label{eq_v_pso} \\ 
    x_{i,d}^{t} =& x_{i,d}^{t-1}+v_{i,d}^{t} \label{eq_x_pso}
\end{align}
where $x_{i,d}^{t}$ and $v_{i,d}^{t}$ denote the $d$-th dimension values of $\vec{X_{i}^{t}}$ and $\vec{V_{i}^{t}}$, respectively; $w$ is the inertia weight; $c_1$ and $c_2$ are acceleration coefficients determining the weights of learning from $pbest$ and $gbest$, respectively. The random numbers $rand_{1,d}$ and $rand_{2,d}$ are uniformly distributed in the range of [0, 1].

\subsection{Related Work}

To date, numerous studies \cite{ImpactCommunicationTopology2019a, pop_topologies_for_pso, population_structure,FIPS} have investigated different topologies in PSO, which are mainly categorized into global topology and local topology. As indicated in Fig. \ref{fig1} (a), the Fully connected topology is a global topology, where each particle is connected with all the other particles within a swarm. In the global topology, each particle is influenced by the best position discovered by the entire swarm, facilitating fast convergence but increasing the risk of premature convergence to local optima. In contrast, the local topology restricts each particle’s social interactions to a neighborhood, which preserves swarm diversity and enhances exploration, while it often leads to slower convergence. The Ring topology, illustrated in Fig. \ref{fig1} (b), is a local topology, where each particle is connected only to its neighboring particles. These standard topologies can be conceptualized as a graph, where a node corresponds to a particle and an edge connects only two particles.

Many researchers have extensively investigated them using graph theory. Oliveira et al. \cite{AssessingParticleSwarm2013} developed a set of network science metrics for analyzing information flow in PSO, enabling a better understanding of swarm behavior and facilitating the study of PSO variants and performance evaluation. He et al. \cite{GraphTheoreticalAnalysis2020} studied the influence of population topology on PSO performance based on randomly generated topologies, with a particular focus on two aspects: sparsity of the topology and the clustering coefficient. By employing complex network theory, Deng et al. \cite{CollectiveDynamicsParticle2025} investigated the topological structure of PSO and demonstrated that its network structure exhibits small-world architecture and heavy-tailed degree distributions. The aforementioned studies have investigated the topological characteristics of PSO through graph theory, thereby providing valuable insights and methodological foundations for the development of PSO variants. However, these studies primarily focus on analyzing static topological structures, often overlooking the dynamic nature of information flow during the evolutionary process. This limits their ability to capture the adaptive characteristics of swarm behavior.

Achieving topological improvement by altering information flow among particles has become a prominent direction in PSO research. Mendes et al. \cite{FIPS} proposed the fully informed particle swarm (FIPS). Each particle is guided by information from all its neighbors, using the complete information of the entire neighborhood of the particle. Based on the concept of multiple subpopulations, Liang et al. \cite{DMS-PSO} proposed the dynamic multi-swarm PSO (DMS-PSO). In \cite{DMS-PSO}, the entire swarm is divided into multiple sub-swarms, which are regrouped and repartitioned by using various regrouping schedules, with information exchanged between these sub-swarms. A dynamic tournament topology was proposed in \cite{DTT-PSO}, called DTT-PSO. DTT-PSO uses the winners of the tournament to guide the particles, ensuring extensive information exchange among them. 

Learning exemplars has also been acknowledged as an important method for improving the population topology. Comprehensive learning PSO (CLPSO) \cite{CLPSO} is a classic example of this approach. It constructs learning exemplars from the $pbest$ of other particles, enabling each particle to learn from different dimensions. Zhan et al. \cite{OLPSO} employed an orthogonal learning strategy to construct an efficient learning exemplar for guiding particles, referred to as orthogonal learning PSO (OLPSO). Cheng et al. \cite{SLPSO} proposed the social learning PSO (SLPSO). Each particle learns from any better particle in the current swarm. more recently, Xia et al. \cite{XPSO} introduced an expanded PSO (XPSO). In XPSO, each particle expands its social learning exemplars from one to two, and a forgetting mechanism is incorporated to simulate individual forgetting behavior. Xia et al. \cite{TAPSO} also proposed the triple archives PSO (TAPSO), which guides particle flight using elites, profiteers, and outstanding exemplars. These studies have effectively alleviated the limitation of premature convergence in PSO by improving the population topology. 

In summary, through topological improvement, existing studies have made significant progress in mitigating premature convergence. To date, existing studies have investigated the topological characteristics of PSO through graph theory, while they primarily focus on static structures, overlooking the dynamic nature of information flow during evolution \cite{ZHANG2022109660}. This limits their ability to capture complicated swarm behaviors. Moreover, edges in these graph-based representations usually fail to directly capture the higher-order social relationships that are essential for navigating complex search landscapes. Therefore, in this study, we investigate how a hypergraph can model the communication topology of the particles in a swarm. Since hyperedges can connect multiple particles, enabling each particle to directly learn from multiple exemplars rather than a single peer.

\section{The Proposed Method: Hypergraph-assisted Particle Swarm Optimizer}
This section introduces the proposed HPSO method. HPSO has two particular focuses, the first of which is how the communication topology of the particles in a swarm can be modeled by a hypergraph and adaptively updated. The second is how the hypergraph topology can be used to effectively navigate particles to search for complex search landscapes.

\subsection{Swarm Topology in HPSO}
The swarm topology in PSO is conventionally represented using a graph structure, wherein information dissemination is constrained by pairwise connections. As a generalization of the graph, hypergraph-based topology that is capable of capturing higher-order interactions, presents itself as a compelling alternative for overcoming this inherent limitation. 

\begin{figure}
\centering
\includegraphics[scale=0.7]{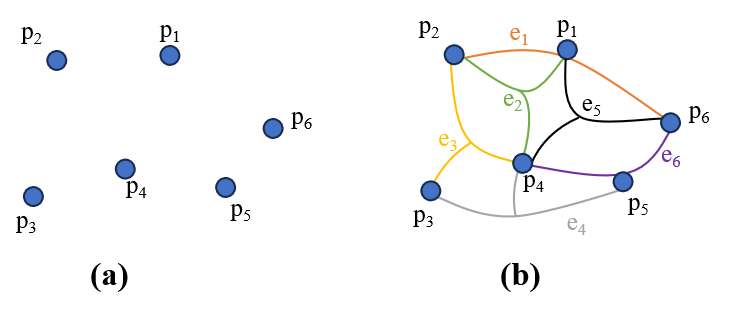}
\caption{(a) A swarm consisting of six particles; (b) The hypergraph topology constructed from the swarm.}
\label{fig2}
\end{figure}

In HPSO, we use k-nearest-neighbor (KNN) to construct a hypergraph based on the position vectors of the particles \cite{HypergraphLearningMethods2022}. In a swarm, each particle $p_i$ is treated as a vertex, for which a hyperedge $e_i$ is constructed by taking $p_i$ as the centroid and connecting it to its $k$ nearest neighbors, thereby capturing local neighborhood relationships.

Through the above steps, we construct the hypergraph topology of the particle swarm. 
In Fig. \ref{fig2}, the constructed swarm topology in HPSO is indicated intuitively. Particles are connected through hyperedges to form a unified social group. Each particle can learn from multiple other particles within a hyperedge. This is clearly different from the traditional graph topology, where learning is restricted to a single particle. Furthermore, since a particle belongs to multiple hyperedges, it can participate in learning across different hyperedges. For example, in Fig. \ref{fig2} (b), particle $p_1$ belongs to hyperedges $e_1$, $e_2$, and $e_5$, and thus serves as a learning exemplar for other particles within the three hyperedges. This diversification of interaction pathways facilitates efficient information dissemination, preventing premature convergence to local optima.

\subsection{Hypergraph Updating Strategy}

Because particles are constantly moving and their spatial distribution changes over time, a hypergraph constructed at the initial stage no longer reflects the true state of the swarm. Therefore, the hypergraph must be dynamically updated to capture the current distribution at any given stage of the optimization process. We propose an adaptive hypergraph updating strategy.

Specifically, the update is triggered by the cumulative average displacement of particles across generations. A large cumulative displacement of particles indicates a significant change in the swarm distribution, while a low displacement suggests that the current hypergraph structure is capable of adequately reflecting the swarm's distribution characteristics. Therefore, we define the average displacement of particles in the $a$-th cumulative as $dis^{a}$, calculated as follows:

\begin{align}
    dis^a=dis^{a-1}+\frac{1}{N}\sum_{i=1}^N{dis_{0,i}},
\label{eqd^a}
\end{align}
where $dis_{0,i}$ denotes the displacement distance $dis_0$ of the $i$-th particle and $N$ is the swarm size. 

Subsequently, we set a threshold $\delta$ to determine whether the hypergraph needs to be updated. When the cumulative average displacement $dis^a$ exceeds the predefined threshold $\delta$, the hypergraph is reconstructed to reflect the current swarm state, and $dis^a$ is thereupon reinitialized to zero. 
For generality, the search space can be regarded as a $D$-dimensional hypercube, with each dimension $d$ ranging from $[L_d, U_d]$.
To ensure that the threshold  $\delta$ remains consistent across different optimization problems, the diagonal distance of the hypercube is selected as the threshold for the $dis^a$.
Since the diagonal represents the maximum length in the space, if the displacement distance exceeds this threshold, it indicates that the swarm has moved a considerable distance and an updating strategy on the hypergraph is required. Therefore, the $\delta$ can be calculated as follows:
\begin{align}
    \delta =\sqrt{\sum_{d=1}^D{\left( U_d-L_d \right) ^2}}.
\label{eqd^delta}
\end{align}

The $\delta$ ensures that unnecessary computational overhead is avoided due to frequent updates, and also provides a unified and reasonable threshold reference across different dimensions and scales.

\subsection{Hypergraph-based Learning Strategy in HPSO}

In the hypergraph-based swarm topology, since each particle from the same swarm can belong to multiple hyperedges, hyperedge weights are assigned to determine which hyperedge a particle should learn from. The hyperedge weights ($w_e$) are calculated as follows:
\begin{align}
    w_e=\sqrt{\sum_{p_i\in e}{f_{p_i}^{2}}},
\label{eqw}
\end{align}
where $f_{p_i}$ represents the fitness value of the particle $p_i$. 

Throughout this study, without loss of generality, all problems are treated as minimization problems. Therefore, hyperedges with smaller weights are considered better. 
We calculate $w_e$ according to Eq. \eqref{eqw} to determine which hyperedge particle $i$ belongs to, and then particle $i$ will learn from the other particles (i.e., its neighbors) within this hyperedge. Given that $K$ is the number of the other particles (excluding $i$) in hyperedge $e$ to which $i$ belongs, the hyperedge is defined as $e= \left\{i, p_{i1},\cdots, p_{ik} \right\}$, where $ p_{i1}$, $\cdots$, $p_{ik}$ are the neighbors of particle $i$ in the hyperedge $e$. 
Afterwards, the velocity and position of particle $i$ at iteration $t$ in dimension $d$ are updated as follows:
\begin{align}
    v_{i,d}^{t} &= w \times v_{i,d}^{t-1} + c_1 \times rand_{1,d} \times (pbest_{i,d}^{t-1} - x_{i,d}^{t-1}) \notag \\
    &\quad + c_2 \times rand_{2,d} \times \frac{1}{K} \sum_{k=1}^K h^{t-1}_{k,d}, \label{eq_velocity} \\
    x_{i,d}^{t} &= x_{i,d}^{t-1} + v_{i,d}^{t}, \label{eq_position}
\end{align}
where the inertia weight $w$, acceleration coefficients $c_1$ and $c_2$, and $rand_{1,d}$ and $rand_{2,d}$ are the same as those in Eq. (\ref{eq_v_pso}); and $\textit{h}^{t-1}_{k,d}$ is the number in the $k$-th row and $d$-th column of matrix $\mathbf{S}$ (i.e., $\mathbf{S}  = [h_{k,d}]_{K \times D}$, where $D$ is the number of dimensions in a particle). $\mathbf{S}$ can be computed as follows:

\begin{align}
\mathbf{S} =\left( \begin{array}{c}
	pbest_{p_{i1}}^{t-1}\\
	\vdots\\
	pbest_{p_{ik}}^{t-1}\\
\end{array} \right) -\left( \begin{array}{c}
	x_{i}^{t-1}\\
	\vdots\\
	x_{i}^{t-1}\\
\end{array} \right). 
\label{eqh}
\end{align}

\begin{figure}
\centering
\includegraphics[scale=0.95]{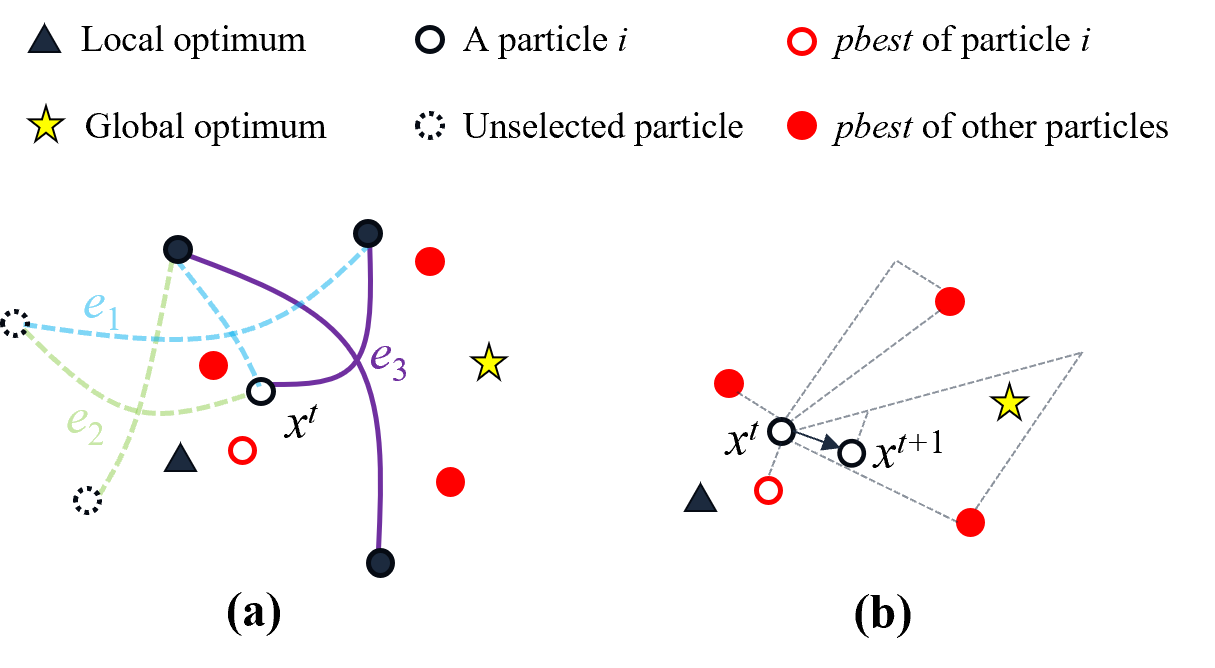}

\caption{Particle trajectory simulation diagram in HPSO: (a) The basic information of particle swarm and the corresponding hypergraph topology; (b) The trajectory diagram of the particle $i$ from $x^t$ to $x^{t+1}$.}
\label{fig3}
\end{figure}

The standard PSO is built upon a graph topology that facilitates only pairwise information flow, confining each particle to a single social learning source (i.e., the global best). This inherently limits the breadth and diversity of information accessible during the optimization process. By contrast, the proposed learning strategy allows particles to draw information from multiple individuals simultaneously. Fig. \ref{fig3} shows a typical flight trajectory formed by particles under this strategy. We first present a simple particle swarm along with the corresponding hypergraph topology in Fig. \ref{fig3} (a). In this swarm, particle $i$ is involved in three constructed hyperedges (i.e., $e_1$, $e_2$, and $e_3$). The weight of each hyperedge is calculated according to Eq. (\ref{eqw}). The hyperedge with the smallest weight, i.e., $e_3$, is selected, shown as a solid line, and the other particles it comprises are indicated by solid black dots. Subsequently, in Fig. \ref{fig3} (b), particle $i$ moves from position $x^t$ to $x^{t+1}$ according to Eq. \eqref{eq_position}. Specifically, $x^{t+1}$ can be determined from the $pbest$ of these particles using the parallelogram rule, with dashed lines serving as auxiliary guides. By incorporating information from multiple neighboring particles, $x^{t+1}$ is guided closer to the global optimum, thereby reducing the risk of premature convergence to local optima. 

\subsection{The Overall Design of HPSO}
\begin{figure}
\centering
\includegraphics[scale=0.5]{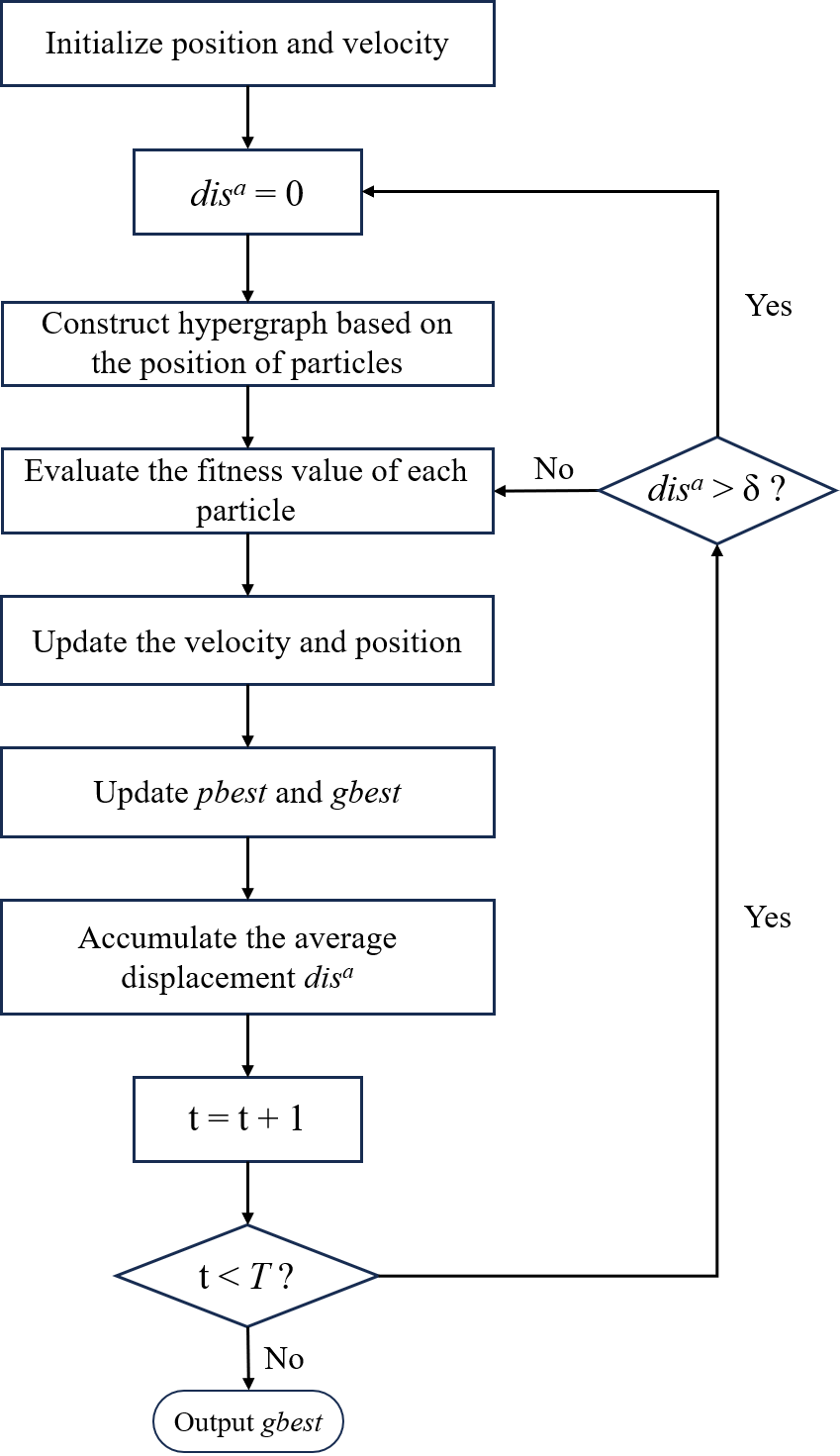}
\caption{Flowchart of the HPSO algorithm.}
\label{fig_flowchart}
\end{figure}

Based on the above hypergraph topological structure and learning strategy, the complete pseudocode of HPSO is presented in Algorithm \ref{HPSO}, and the flowchart of it is shown in Fig. \ref{fig_flowchart}. To be specific, HPSO initializes the particle swarm and $dis^a$ in step 1 and 2 of Algorithm \ref{HPSO}. In step 3, HPSO constructs a KNN hypergraph based on the position of the particles. If the $dis^a$ exceeds the threshold $\delta$, the hypergraph update strategy is executed to capture the current distribution of the swarm from step 5 to step 8. Next, HPSO updates the velocity based on the hypergraph topology and calculates the displacement of each particle from step 13 to step 18. Finally, the cumulative average displacement $dis^a$ of the particle swarm is calculated to determine whether the hypergraph update strategy (steps 19–20) should be executed.

\begin{algorithm}
    \caption{HPSO}
    \label{HPSO}
    
    \KwIn{Swarm size $N$, Max iteration $T$}
    \KwOut{Global best $gbest$}
    
    Randomly initialize Swarm\;
    Set the average cumulative displacement of the initial particles $dis^a$ to 0\;
    $\mathbf{H} \gets$ KNN hypergraph construction\;
    
    \For{$t=0$ \textnormal{to} $T$}{
        \If{$dis^a > \delta$}{
            $\mathbf{H} \gets$ KNN hypergraph construction\;
            $dis^a \gets 0$\;
        }
        
        \For{$i=0$ \textnormal{to} $N$}{
            Evaluate particle $i$\;
            Update $pbest_i$ and $gbest$\;
        }
        
        \For{$i=0$ \textnormal{to} $N$}{
            Calculate the hyperedge weights $w_e$ associated with particle $i$ according to the Eq. (\ref{eqw})\;
            Obtain the hyperedge $e$ with the smallest weight\;
            Update velocity and position of particle $i$ according to Eqs. \eqref{eq_velocity} and \eqref{eq_position}\;
            Calculate the displacement of particle $i$\;
        }
        
        Calculate the average particle displacement $dis^0$ of the current iteration $t$\;
        Obtain the cumulative average displacement $dis^a$ according to Eq. (\ref{eqd^a})\;
    }
    
    \Return $gbest$\;
\end{algorithm}

\section{Experiment Design}
\subsection{Benchmark Functions}

\begin{table}[htbp]
\centering
\small
\caption{Summary of the CEC’17 benchmark suite.}
\label{CEC17}
\setlength{\tabcolsep}{1pt}
\renewcommand\arraystretch{1}
\begin{tabular}{cclc}
\hline
                                                                                        & No. & \multicolumn{1}{c}{Functions}                                                                      & $F_i^*=F_i(x^*)$ \\ \hline
\multirow{2}{*}{\begin{tabular}[c]{@{}c@{}}Unimodal \\ Functions\end{tabular}}          & 1   & \begin{tabular}[c]{@{}l@{}}Shifted and Rotated Bent Cigar \\ Function\end{tabular}                 & 100              \\
                                                                                        & 3   & \begin{tabular}[c]{@{}l@{}}Shifted and Rotated Zakharov \\ Function\end{tabular}                   & 200              \\
\multirow{7}{*}{\begin{tabular}[c]{@{}c@{}}Simple \\Multimodal \\Functions\end{tabular}} & 4   & \begin{tabular}[c]{@{}l@{}}Shifted and Rotated Rosenbrock's \\ Function\end{tabular}               & 300              \\
                                                                                        & 5   & \begin{tabular}[c]{@{}l@{}}Shifted and Rotated Rastrigin's \\ Function\end{tabular}                & 400              \\
                                                                                        & 6   & \begin{tabular}[c]{@{}l@{}}Shifted and Rotated Expanded \\ Scaffer's F6 Function\end{tabular}      & 500              \\
                                                                                        & 7   & \begin{tabular}[c]{@{}l@{}}Shifted and Rotated Lunacek Bi \\ Rastrigin Function\end{tabular}       & 600              \\
                                                                                        & 8   & \begin{tabular}[c]{@{}l@{}}Shifted and Rotated Non-\\Continuous  Rastrigin's Function\end{tabular} & 700              \\
                                                                                        & 9   & \begin{tabular}[c]{@{}l@{}}Shifted and Rotated Levy \\ Function\end{tabular}                       & 800              \\
                                                                                        & 10  & \begin{tabular}[c]{@{}l@{}}Shifted and Rotated Schwefel's \\ Function\end{tabular}                 & 900              \\
\multirow{10}{*}{\begin{tabular}[c]{@{}c@{}}Hybrid \\ Functions\end{tabular}}           & 11  & Hybrid Function 1 (N=3)                                                                            & 1000             \\
                                                                                        & 12  & Hybrid Function 2 (N=3)                                                                            & 1100             \\
                                                                                        & 13  & Hybrid Function 3 (N=3)                                                                            & 1200             \\
                                                                                        & 14  & Hybrid Function 4 (N=4)                                                                            & 1300             \\
                                                                                        & 15  & Hybrid Function 5 (N=4)                                                                            & 1400             \\
                                                                                        & 16  & Hybrid Function 6 (N=4)                                                                            & 1500             \\
                                                                                        & 17  & Hybrid Function 6 (N=5)                                                                            & 1600             \\
                                                                                        & 18  & Hybrid Function 6 (N=5)                                                                            & 1700             \\
                                                                                        & 19  & Hybrid Function 6 (N=5)                                                                            & 1800             \\
                                                                                        & 20  & Hybrid Function 6 (N=6)                                                                            & 1900             \\
\multirow{10}{*}{\begin{tabular}[c]{@{}c@{}}Composition \\ Functions\end{tabular}}      & 21  & Composition Function 1 (N=3)                                                                       & 2000             \\
                                                                                        & 22  & Composition Function 2 (N=3)                                                                       & 2100             \\
                                                                                        & 23  & Composition Function 3 (N=4)                                                                       & 2200             \\
                                                                                        & 24  & Composition Function 4 (N=4)                                                                       & 2300             \\
                                                                                        & 25  & Composition Function 5 (N=5)                                                                       & 2400             \\
                                                                                        & 26  & Composition Function 6 (N=5)                                                                       & 2500             \\
                                                                                        & 27  & Composition Function 7 (N=6)                                                                       & 2600             \\
                                                                                        & 28  & Composition Function 8 (N=6)                                                                       & 2700             \\
                                                                                        & 29  & Composition Function 9 (N=3)                                                                       & 2800             \\
                                                                                        & 30  & Composition Function 10 (N=3)                                                                      & 2900             \\ \hline
\multicolumn{4}{c}{Search Range: $[-100,100]^D$}                                                                                                                                                                      \\ \hline
\end{tabular}
    \begin{flushleft}
      \footnotesize         
      Note that $F_2$ has been deprecated from the CEC'17 benchmark suite.
    \end{flushleft}
\end{table}

To verify the effectiveness of HPSO, the IEEE-CEC'17 benchmark suite \cite{CEC17Benchmark} is adopted as the set of test functions in our experiments. Table \ref{CEC17} presents the summary of the CEC'17 benchmark suite. The following is a brief introduction to this suit. Two unimodal functions ($F_1$ and $F_3$) exhibit a narrow ridge and are non-separable, used to evaluate the exploitation capability of optimization algorithms. The simple multimodal functions ($F_4$--$F_{10}$) contain numerous local optimum and are used to assess the algorithm’s ability to escape from local optima.  
For ten hybrid functions ($F_{11}$--$F_{20}$), the global optimum corresponds to the local optimum with the smallest bias. These functions thus present a significant challenge, requiring algorithms to effectively navigate complex, multi-structured fitness landscapes.
In addition, composition functions ($F_{21}$--$F_{30}$) combine multiple basic or hybrid functions in a weighted manner to create highly complex, irregular, and non-separable landscapes. Therefore, the IEEE-CEC'17 benchmark suite provides a comprehensive and reliable basis for evaluating the performance of optimization algorithms.

\subsection{Baseline Methods and Parameter Settings}

\begin{table}[]
\centering
\caption{Parameter settings of the eight algorithms.}
\setlength{\tabcolsep}{2.5pt}
\renewcommand\arraystretch{1}
\begin{tabular}{ccc}
\hline
Algorithms & Parameters                                      & References        \\ \hline
PSO       & $w$: 0.9-0.4, $c_1$=$c_2$=1.49445                       & \cite{SPSO}       \\ 
CLPSO      & $w$: 0.9-0.4, $m$=7, $c$=1.49445                      & \cite{CLPSO}      \\ 
OLPSO      & $w$: 0.9-0.4, $c$=1.49445                           & \cite{OLPSO}      \\ 
PPSO       & None                                            & \cite{PhaPSO}     \\ 
XPSO       & \begin{tabular}[c]{@{}c@{}}$w$: 0.9-0.4, $n$=0.2, $Stage_{max}$=5, \\$p$=0.5\end{tabular}            & \cite{XPSO}       \\ 
TAPSO      & \begin{tabular}[c]{@{}c@{}}$w$=0.7298, $p_c$=0.5, $p_m$=0.02, \\ $M$=$n$/4\end{tabular}                & \cite{TAPSO}      \\ 
AWPSO      & \begin{tabular}[c]{@{}c@{}}$w$: 0.9-0.4, $a$=0.000035*(U-L), \\ $b$=0.5, $c$=0, $d$=1.5\end{tabular} & \cite{AWPSO}      \\ 
HPSO       & $w$: 0.9-0.4, $c_1$=$c_2$=1.49445, $k$=5                  & \textbf{proposed} \\ \hline
\end{tabular}
\label{param}
\end{table}

To verify the effectiveness of our proposed method, several representative and state-of-the-art PSO variants were selected for comparisons, including CLPSO \cite{CLPSO}, OLPSO \cite{OLPSO}, PPSO \cite{PhaPSO}, XPSO \cite{XPSO}, TAPSO \cite{TAPSO}, and AWPSO \cite{AWPSO}. We also included the standard PSO \cite{SPSO} as an explicit baseline to facilitate performance comparison with the variants. The corresponding parameter settings are summarized in Table \ref{param}. The parameter settings and detailed descriptions of these methods can be found in their respective references. In addition, the experiments are conducted in dimensions $D$ = 30 (30D), 50 (50D), and 100 (100D). For all these methods, the swarm size is set to 50, and the number of fitness evaluations (FEs) is set to $10000\times D$. The maximum velocity $V_{max}$ is limited to $0.5 \times (U-L)$. 

To guarantee the reliability of our experiments, all methods are executed 30 times independently on each benchmark function. The implementations are conducted in Python, and hypergraph construction is performed using the THU-HyperG library \cite{HypergraphLearningMethods2022}. All of them are executed on the same personal computer with Intel Core i5-12600KF (3.70 GHz) and 16 GB of memory.

\begin{table*}[]
\centering
\caption{Results for the \textbf{five unimodal and five simple multimodal functions} of the CEC'17 benchmark suite (50D).}
\setlength{\tabcolsep}{6pt}
\renewcommand\arraystretch{1}
\label{50D-1}
\begin{minipage}{1\textwidth}  
    \tiny
    \centering  
  \resizebox{\textwidth}{!}{
\begin{tabular}{ccllllllll}
\hline
\textbf{}                          & \textbf{}     & \multicolumn{1}{c}{\textbf{PSO}} & \multicolumn{1}{c}{\textbf{CLPSO}} & \multicolumn{1}{c}{\textbf{OLPSO}} & \multicolumn{1}{c}{\textbf{PPSO}} & \multicolumn{1}{c}{\textbf{XPSO}} & \multicolumn{1}{c}{\textbf{TAPSO}} & \multicolumn{1}{c}{\textbf{AWPSO}} & \multicolumn{1}{c}{\textbf{HPSO}} \\ \hline
\multirow{3}{*}{\textbf{$F_1$}}    & \textbf{Mean} & 4.56E+11 +                       & \textbf{1.76E+03 =}                & 1.51E+04 +                         & 8.09E+09 +                        & 5.03E+04 +                        & 3.89E+03 =                         & 9.88E+10 +                         & 3.73E+03                          \\
                                   & \textbf{Std}  & 1.70E+11                         & \textbf{1.47E+03}                  & 1.12E+04                           & 1.34E+10                          & 2.20E+05                          & 5.15E+03                           & 4.88E+10                           & 3.57E+03                          \\
                                   & \textbf{Ranking} & 8                                & \textbf{1}                         & 4                                  & 6                                 & 5                                 & 3                                  & 7                                  & 2                                 \\
\multirow{3}{*}{\textbf{$F_3$}}    & \textbf{Mean} & 6.35E+04 +                       & 3.24E+04 =                         & 4.44E+05 +                         & 1.19E+05 +                        & 1.67E+03 -                        & \textbf{1.36E+03 -}                & 3.17E+04 =                         & 3.24E+04                          \\
                                   & \textbf{Std}  & 3.35E+04                         & 3.28E+03                           & 5.63E+04                           & 3.79E+04                          & 9.24E+02                          & \textbf{3.11E+03}                  & 2.41E+04                           & 5.38E+03                          \\
                                   & \textbf{Ranking} & 6                                & 5                                  & 8                                  & 7                                 & 2                                 & \textbf{1}                         & 3                                  & 4                                 \\
\multirow{3}{*}{\textbf{$F_4$}}    & \textbf{Mean} & 6.82E+03 +                       & 4.95E+02 -                         & 6.11E+02 +                         & 7.43E+02 +                        & 6.77E+02 +                        & \textbf{4.36E+02 -}                & 1.48E+03 +                         & 5.48E+02                          \\
                                   & \textbf{Std}  & 3.58E+03                         & 3.67E+01                           & 1.87E+01                           & 1.17E+02                          & 4.21E+01                          & \textbf{4.12E+01}                  & 6.01E+02                           & 4.55E+01                          \\
                                   & \textbf{Ranking} & 8                                & 2                                  & 4                                  & 6                                 & 5                                 & \textbf{1}                         & 7                                  & 3                                 \\
\multirow{3}{*}{\textbf{$F_5$}}    & \textbf{Mean} & 8.41E+02 +                       & 6.38E+02 +                         & 6.82E+02 +                         & 9.10E+02 +                        & 6.12E+02 +                        & 6.15E+02 +                         & 6.99E+02 +                         & \textbf{5.26E+02}                 \\
                                   & \textbf{Std}  & 4.47E+01                         & 1.50E+01                           & 2.47E+01                           & 5.79E+01                          & 2.60E+01                          & 1.93E+01                           & 3.88E+01                           & \textbf{6.53E+00}                 \\
                                   & \textbf{Ranking} & 7                                & 4                                  & 5                                  & 8                                 & 2                                 & 3                                  & 6                                  & \textbf{1}                        \\
\multirow{3}{*}{\textbf{$F_6$}}    & \textbf{Mean} & 6.64E+02 +                       & 6.26E+02 +                         & 6.30E+02 +                         & 7.00E+02 +                        & 6.14E+02 +                        & 6.17E+02 +                         & 6.34E+02 +                         & \textbf{6.03E+02}                 \\
                                   & \textbf{Std}  & 1.19E+01                         & 5.49E+00                           & 6.72E+00                           & 1.16E+01                          & 6.40E+00                          & 6.07E+00                           & 8.11E+00                           & \textbf{1.03E+00}                 \\
                                   & \textbf{Ranking} & 7                                & 4                                  & 5                                  & 8                                 & 2                                 & 3                                  & 6                                  & \textbf{1}                        \\
\multirow{3}{*}{\textbf{$F_7$}}    & \textbf{Mean} & 1.71E+03 +                       & 8.88E+02 +                         & 8.63E+02 +                         & 2.02E+03 +                        & 9.48E+02 +                        & 8.79E+02 +                         & 1.02E+03 +                         & \textbf{7.82E+02}                 \\
                                   & \textbf{Std}  & 3.78E+02                         & 1.52E+01                           & 2.27E+01                           & 1.70E+02                          & 5.45E+01                          & 2.52E+01                           & 6.49E+01                           & \textbf{8.06E+00}                 \\
                                   & \textbf{Ranking} & 7                                & 4                                  & 2                                  & 8                                 & 5                                 & 3                                  & 6                                  & \textbf{1}                        \\
\multirow{3}{*}{\textbf{$F_8$}}    & \textbf{Mean} & 1.16E+03 +                       & 9.41E+02 +                         & 9.92E+02 +                         & 1.22E+03 +                        & 9.17E+02 +                        & 9.08E+02 +                         & 1.01E+03 +                         & \textbf{8.27E+02}                 \\
                                   & \textbf{Std}  & 5.34E+01                         & 1.51E+01                           & 2.36E+01                           & 6.37E+01                          & 3.18E+01                          & 2.38E+01                           & 5.13E+01                           & \textbf{7.45E+00}                 \\
                                   & \textbf{Ranking} & 7                                & 4                                  & 5                                  & 8                                 & 3                                 & 2                                  & 6                                  & \textbf{1}                        \\
\multirow{3}{*}{\textbf{$F_9$}}    & \textbf{Mean} & 1.01E+04 +                       & 2.08E+03 +                         & 3.39E+03 +                         & 2.16E+04 +                        & 1.27E+03 +                        & 1.62E+03 +                         & 3.90E+03 +                         & \textbf{9.42E+02}                 \\
                                   & \textbf{Std}  & 3.61E+03                         & 5.00E+02                           & 9.73E+02                           & 3.94E+03                          & 2.94E+02                          & 5.54E+02                           & 1.44E+03                           & \textbf{5.27E+01}                 \\
                                   & \textbf{Ranking} & 7                                & 4                                  & 5                                  & 8                                 & 2                                 & 3                                  & 6                                  & \textbf{1}                        \\
\multirow{3}{*}{\textbf{$F_{10}$}} & \textbf{Mean} & 8.11E+03 +                       & 7.35E+03 +                         & 8.44E+03 +                         & 1.24E+04 +                        & 7.23E+03 +                        & \textbf{5.57E+03 =}                & 7.03E+03 +                         & 5.81E+03                          \\
                                   & \textbf{Std}  & 9.26E+02                         & 3.31E+02                           & 6.81E+02                           & 1.55E+03                          & 1.60E+03                          & \textbf{7.74E+02}                  & 7.98E+02                           & 9.37E+02                          \\
                                   & \textbf{Ranking} & 6                                & 5                                  & 7                                  & 8                                 & 4                                 & \textbf{1}                         & 3                                  & 2                                 \\ \hline
\end{tabular}
}
\end{minipage}
\end{table*}

\begin{table*}[]
\centering
\caption{Results for the \textbf{ten hybrid functions} of the CEC'17 benchmark suite (50D).}
\setlength{\tabcolsep}{6pt}
\renewcommand\arraystretch{1}
\label{50D-2}
\begin{minipage}{1\textwidth}  
    \tiny
    \centering  
  \resizebox{\textwidth}{!}{%

\begin{tabular}{ccllllllll}
\hline
\textbf{}                          & \textbf{}     & \textbf{PSO} & \textbf{CLPSO}      & \textbf{OLPSO} & \textbf{PPSO} & \textbf{XPSO}       & \textbf{TAPSO}      & \textbf{AWPSO} & \textbf{HPSO}     \\ \hline
\textbf{$F_{11}$}                  & \textbf{Mean} & 3.75E+03 +   & \textbf{1.18E+03 -} & 1.57E+03 +     & 1.95E+03 +    & 1.33E+03 +          & 1.21E+03 =          & 1.68E+03 +     & 1.21E+03          \\
\textbf{}                          & \textbf{Std}  & 3.20E+03     & \textbf{2.85E+01}   & 9.39E+02       & 5.73E+02      & 4.46E+01            & 4.46E+01            & 2.73E+02       & 3.82E+01          \\
\textbf{}                          & \textbf{Ranking} & 8            & \textbf{1}          & 5              & 7             & 4                   & 3                   & 6              & 2                 \\
\multirow{3}{*}{\textbf{$F_{12}$}} & \textbf{Mean} & 1.05E+11 +   & 1.28E+06 -          & 3.99E+08 +     & 3.37E+09 +    & 6.26E+06 =          & \textbf{1.12E+05 -} & 3.27E+10 +     & 2.44E+06          \\
                                   & \textbf{Std}  & 7.15E+10     & 4.70E+05            & 4.60E+08       & 8.38E+09      & 1.02E+07            & \textbf{8.92E+04}   & 2.45E+10       & 1.74E+06          \\
                                   & \textbf{Ranking} & 8            & 2                   & 5              & 6             & 4                   & \textbf{1}          & 7              & 3                 \\
\multirow{3}{*}{\textbf{$F_{13}$}} & \textbf{Mean} & 3.58E+10 +   & \textbf{3.48E+03 -} & 5.13E+07 +     & 1.31E+09 +    & 2.06E+05 =          & 1.42E+04 =          & 1.49E+10 +     & 1.11E+04          \\
                                   & \textbf{Std}  & 3.68E+10     & \textbf{1.85E+03}   & 9.20E+07       & 3.07E+09      & 8.66E+05            & 1.11E+04            & 2.36E+10       & 1.02E+04          \\
                                   & \textbf{Ranking} & 8            & \textbf{1}          & 5              & 6             & 4                   & 3                   & 7              & 2                 \\
\multirow{3}{*}{\textbf{$F_{14}$}} & \textbf{Mean} & 1.72E+06 +   & 1.97E+05 +          & 9.80E+05 +     & 2.70E+06 +    & 1.33E+05 =          & \textbf{2.38E+04 -} & 3.93E+05 =     & 1.08E+05          \\
                                   & \textbf{Std}  & 2.63E+06     & 1.15E+05            & 9.07E+05       & 8.79E+06      & 2.06E+05            & \textbf{3.19E+04}   & 5.59E+05       & 5.12E+04          \\
                                   & \textbf{Ranking} & 7            & 4                   & 6              & 8             & 3                   & \textbf{1}          & 5              & 2                 \\
\multirow{3}{*}{\textbf{$F_{15}$}} & \textbf{Mean} & 1.78E+09 +   & 9.95E+03 =          & 2.49E+04 +     & 1.80E+06 +    & 1.11E+04 =          & 9.07E+03 =          & 3.26E+08 +     & \textbf{8.66E+03} \\
                                   & \textbf{Std}  & 5.45E+09     & 3.74E+03            & 1.05E+04       & 3.31E+06      & 7.77E+03            & 6.91E+03            & 1.06E+09       & \textbf{5.20E+03} \\
                                   & \textbf{Ranking} & 8            & 3                   & 5              & 6             & 4                   & 2                   & 7              & \textbf{1}        \\
\multirow{3}{*}{\textbf{$F_{16}$}} & \textbf{Mean} & 4.10E+03 +   & 2.81E+03 +          & 3.94E+03 +     & 5.62E+03 +    & 2.94E+03 +          & 3.04E+03 +          & 3.34E+03 +     & \textbf{2.46E+03} \\
                                   & \textbf{Std}  & 5.22E+02     & 1.90E+02            & 3.40E+02       & 1.58E+03      & 4.45E+02            & 4.46E+02            & 4.35E+02       & \textbf{3.88E+02} \\
                                   & \textbf{Ranking} & 7            & 2                   & 6              & 8             & 3                   & 4                   & 5              & \textbf{1}        \\
\multirow{3}{*}{\textbf{$F_{17}$}} & \textbf{Mean} & 4.24E+03 +   & 2.49E+03 +          & 3.45E+03 +     & 5.26E+03 +    & 2.62E+03 +          & 2.68E+03 +          & 3.07E+03 +     & \textbf{2.28E+03} \\
                                   & \textbf{Std}  & 1.46E+03     & 1.02E+02            & 2.52E+02       & 1.02E+03      & 2.91E+02            & 2.93E+02            & 3.15E+02       & \textbf{2.08E+02} \\
                                   & \textbf{Ranking} & 7            & 2                   & 6              & 8             & 3                   & 4                   & 5              & \textbf{1}        \\
\multirow{3}{*}{\textbf{$F_{18}$}} & \textbf{Mean} & 3.98E+06 =   & 4.08E+05 -          & 3.79E+06 +     & 7.05E+06 +    & 1.64E+06 -          & \textbf{9.06E+04 -} & 3.18E+06 +     & 1.75E+06          \\
                                   & \textbf{Std}  & 6.35E+06     & 2.53E+05            & 2.78E+06       & 1.97E+07      & 2.16E+06            & \textbf{1.28E+05}   & 8.38E+06       & 1.17E+06          \\
                                   & \textbf{Ranking} & 7            & 2                   & 6              & 8             & 3                   & \textbf{1}          & 5              & 4                 \\
\multirow{3}{*}{\textbf{$F_{19}$}} & \textbf{Mean} & 7.80E+08 +   & 1.46E+04 =          & 8.55E+04 +     & 9.70E+06 +    & \textbf{1.30E+04 =} & 1.94E+04 =          & 1.54E+07 +     & 1.75E+04          \\
                                   & \textbf{Std}  & 2.62E+09     & 4.62E+03            & 1.83E+05       & 1.91E+07      & \textbf{9.96E+03}   & 1.38E+04            & 2.61E+07       & 1.21E+04          \\
                                   & \textbf{Ranking} & 8            & 2                   & 5              & 6             & \textbf{1}          & 4                   & 7              & 3                 \\
\multirow{3}{*}{\textbf{$F_{20}$}} & \textbf{Mean} & 3.08E+03 +   & 2.66E+03 +          & 3.26E+03 +     & 3.91E+03 +    & 2.65E+03 +          & 2.97E+03 +          & 2.85E+03 +     & \textbf{2.39E+03} \\
                                   & \textbf{Std}  & 3.00E+02     & 1.30E+02            & 2.70E+02       & 3.46E+02      & 3.05E+02            & 3.10E+02            & 2.54E+02       & \textbf{2.80E+02} \\
                                   & \textbf{Ranking} & 6            & 3                   & 7              & 8             & 2                   & 5                   & 4              & \textbf{1}        \\ \hline
\end{tabular}
}
\end{minipage}
\end{table*}

\begin{table*}[]
\centering
\caption{Results for the \textbf{ten composition functions} of the CEC'17 benchmark suite (50D).}
\setlength{\tabcolsep}{6pt}
\renewcommand\arraystretch{1}
\label{50D-3}
\begin{minipage}{1\textwidth}  
    \tiny
    \centering  
  \resizebox{\textwidth}{!}{%
\begin{tabular}{ccllllllll}
\hline
\textbf{}                          & \textbf{}     & \textbf{PSO} & \textbf{CLPSO}      & \textbf{OLPSO} & \textbf{PPSO} & \textbf{XPSO}       & \textbf{TAPSO}      & \textbf{AWPSO} & \textbf{HPSO}     \\ \hline
\multirow{3}{*}{\textbf{$F_{21}$}} & \textbf{Mean} & 2.70E+03 +   & 2.45E+03 +          & 2.52E+03 +     & 2.83E+03 +    & 2.40E+03 +          & 2.42E+03 +          & 2.53E+03 +     & \textbf{2.34E+03} \\
                                   & \textbf{Std}  & 5.51E+01     & 1.31E+01            & 3.04E+01       & 8.49E+01      & 2.91E+01            & 2.48E+01            & 4.98E+01       & \textbf{9.49E+00} \\
                                   & \textbf{Ranking} & 7            & 4                   & 5              & 8             & 2                   & 3                   & 6              & \textbf{1}        \\
\multirow{3}{*}{\textbf{$F_{22}$}} & \textbf{Mean} & 9.87E+03 +   & 7.63E+03 +          & 1.04E+04 +     & 1.42E+04 +    & 8.70E+03 +          & 7.77E+03 +          & 9.15E+03 +     & \textbf{7.01E+03} \\
                                   & \textbf{Std}  & 1.01E+03     & 2.63E+03            & 7.30E+02       & 1.65E+03      & 1.74E+03            & 8.38E+02            & 1.08E+03       & \textbf{1.04E+03} \\
                                   & \textbf{Ranking} & 6            & 2                   & 7              & 8             & 4                   & 3                   & 5              & \textbf{1}        \\
\multirow{3}{*}{\textbf{$F_{23}$}} & \textbf{Mean} & 3.46E+03 +   & 2.88E+03 +          & 3.01E+03 +     & 3.99E+03 +    & \textbf{2.86E+03 =} & 2.88E+03 =          & 3.25E+03 +     & 2.87E+03          \\
                                   & \textbf{Std}  & 1.05E+02     & 1.82E+01            & 2.34E+01       & 2.24E+02      & \textbf{3.63E+01}   & 3.41E+01            & 1.37E+02       & 5.70E+01          \\
                                   & \textbf{Ranking} & 7            & 4                   & 5              & 8             & \textbf{1}          & 3                   & 6              & 2                 \\
\multirow{3}{*}{\textbf{$F_{24}$}} & \textbf{Mean} & 3.67E+03 +   & 3.15E+03 +          & 3.31E+03 +     & 4.08E+03 +    & 3.13E+03 +          & 3.07E+03 =          & 3.51E+03 +     & \textbf{3.06E+03} \\
                                   & \textbf{Std}  & 1.43E+02     & 3.30E+01            & 2.76E+01       & 2.97E+02      & 9.77E+01            & 3.88E+01            & 1.66E+02       & \textbf{4.91E+01} \\
                                   & \textbf{Ranking} & 7            & 4                   & 5              & 8             & 3                   & 2                   & 6              & \textbf{1}        \\
\multirow{3}{*}{\textbf{$F_{25}$}} & \textbf{Mean} & 6.01E+03 +   & 3.07E+03 +          & 3.03E+03 =     & 3.19E+03 +    & 3.11E+03 +          & 3.04E+03 =          & 3.46E+03 +     & 3.04E+03          \\
                                   & \textbf{Std}  & 2.00E+03     & 2.17E+01            & 1.53E+01       & 3.99E+01      & 3.51E+01            & 4.24E+01            & 2.42E+02       & 2.86E+01          \\
                                   & \textbf{Ranking} & 8            & 4                   & 1              & 6             & 5                   & 3                   & 7              & 2                 \\
\multirow{3}{*}{\textbf{$F_{26}$}} & \textbf{Mean} & 1.16E+04 +   & 5.28E+03 +          & 6.69E+03 +     & 1.35E+04 +    & 5.21E+03 +          & 5.26E+03 +          & 7.54E+03 +     & \textbf{4.77E+03} \\
                                   & \textbf{Std}  & 1.77E+03     & 1.79E+02            & 2.77E+02       & 1.84E+03      & 2.18E+02            & 3.23E+02            & 1.36E+03       & \textbf{4.30E+02} \\
                                   & \textbf{Ranking} & 7            & 4                   & 5              & 8             & 2                   & 3                   & 6              & \textbf{1}        \\
\multirow{3}{*}{\textbf{$F_{27}$}} & \textbf{Mean} & 4.08E+03 +   & \textbf{3.35E+03 -} & 3.49E+03 +     & 4.75E+03 +    & 3.54E+03 +          & 3.38E+03 =          & 3.71E+03 +     & 3.41E+03          \\
                                   & \textbf{Std}  & 2.69E+02     & \textbf{2.93E+01}   & 4.98E+01       & 5.95E+02      & 1.30E+02            & 8.21E+01            & 2.07E+02       & 1.11E+02          \\
                                   & \textbf{Ranking} & 7            & \textbf{1}          & 4              & 8             & 5                   & 2                   & 6              & 3                 \\
\multirow{3}{*}{\textbf{$F_{28}$}} & \textbf{Mean} & 9.52E+03 +   & 3.32E+03 =          & 7.70E+03 +     & 3.59E+03 +    & 3.40E+03 +          & \textbf{3.30E+03 -} & 4.81E+03 +     & 3.31E+03          \\
                                   & \textbf{Std}  & 1.49E+03     & 1.42E+01            & 1.05E+03       & 2.77E+02      & 5.85E+01            & \textbf{3.83E+01}   & 1.12E+03       & 1.75E+01          \\
                                   & \textbf{Ranking} & 8            & 3                   & 7              & 5             & 4                   & \textbf{1}          & 6              & 2                 \\
\multirow{3}{*}{\textbf{$F_{29}$}} & \textbf{Mean} & 5.71E+03 +   & 3.72E+03 +          & 5.01E+03 +     & 9.77E+03 +    & 4.19E+03 +          & 3.91E+03 +          & 4.52E+03 +     & \textbf{3.61E+03} \\
                                   & \textbf{Std}  & 1.13E+03     & 1.17E+02            & 3.25E+02       & 2.82E+03      & 3.94E+02            & 2.95E+02            & 4.80E+02       & \textbf{1.92E+02} \\
                                   & \textbf{Ranking} & 7            & 2                   & 6              & 8             & 4                   & 3                   & 5              & \textbf{1}        \\
\multirow{3}{*}{\textbf{$F_{30}$}} & \textbf{Mean} & 1.65E+09 +   & 1.31E+06 +          & 4.18E+07 +     & 4.86E+08 +    & 3.50E+06 +          & \textbf{9.61E+05 -} & 4.25E+07 +     & 1.15E+06          \\
                                   & \textbf{Std}  & 2.08E+09     & 2.38E+05            & 4.29E+07       & 2.45E+09      & 1.68E+06            & \textbf{2.76E+05}   & 5.04E+07       & 3.47E+05          \\
                                   & \textbf{Ranking} & 8            & 3                   & 5              & 7             & 4                   & \textbf{1}          & 6              & 2                 \\ \hline
\end{tabular}
}
\end{minipage}
\end{table*}

\section{Results and Discussions}
\subsection{Results on CEC'17 Benchmark Suite}
In this subsection, we present the results of the eight PSOs on the CEC'17 benchmark suite. In addition, we conducted a Wilcoxon rank-sum test at a significance level of 0.05 \cite{PPSO}. The results are reported using the mean value (\textbf{Mean}), the standard deviation (\textbf{Std}), and the ranking (\textbf{Ranking}) achieved by these methods. The best \textbf{Mean} among the eight methods is highlighted in bold. In these tables, `+', `-', and `=' indicate cases where HPSO performs significantly better than, worse than, or similarly to a compared baseline method, respectively.
Due to page constraints, only the results for 50D are presented in Tables~\ref{50D-1}-\ref{50D-3}, while the 30D and 100D results are provided in \textbf{APPENDIX A}.
\subsubsection{Unimodal and simple multimodal functions}
According to Table \ref{50D-1}, Table S-1 and Table S-4 in \textbf{APPENDIX A}, HPSO achieves significantly superior performance on most simple multimodal functions. Specifically, HPSO achieves the best mean values on 5 out of the 7 simple multimodal functions across 30D, 50D, and 100D. For the unimodal functions, HPSO fails to achieve optimal performance. This is mainly because its hyperedge-based social learning mechanism prioritizes global exploration over local exploitation. 

Based on the results of the Wilcoxon rank-sum tests, HPSO outperforms the baseline methods in 51, 54, and 51 out of a total of 63 across the 30D, 50D, and 100D, respectively. As dimensionality increases, HPSO maintains a consistent advantage on the multimodal functions, demonstrating its robustness and satisfactory optimization capability in complex multimodal landscapes.

\subsubsection{Hybrid functions}
The hybrid function combines the characteristics of multiple sub-functions, featuring extremely high nonlinearity and a complex local landscape. This places higher demands on the algorithm's ability to balance global exploration and local exploitation. The results shown in Tables \ref{50D-2}, S-2, and S-5 in \textbf{APPENDIX A} indicate that HPSO consistently ranks first among the compared methods. 
Besides, HPSO demonstrates particularly strong performance on $F_{16}$, $F_{17}$, and $F_{20}$, while consistently ranking among the top three on most other functions. This indicates our proposed method has high optimization accuracy when dealing with complex search spaces. The main reason for this strong performance lies in the hypergraph topology. By capturing high-order relationships among particles, it effectively handles the complex interactions between functional variables.

Furthermore, the results in Tables \ref{50D-2}, S-2, and S-5 in \textbf{APPENDIX A} for the hybrid functions indicate that HPSO achieves the best performance in 3, 4, and 3 out of the total 10 functions across 30D, 50D, and 100D, respectively.

From a more comprehensive perspective, HPSO achieves a satisfactory performance, outperforming the compared algorithms in 47, 49, and 51 cases out of the total of 63 at 30D, 50D, and 100D, respectively. Thus, the hypergraph topology offers an inherent advantage in addressing hybrid functions with complex variable interactions, providing a satisfactory framework for optimization.
\subsubsection{Composition Functions}
Composition functions create rugged landscapes with numerous local optima by aggregating weighted sub-functions. They evaluate the algorithm's ability to escape local optima and maintain global exploration stability. The results on the composition functions shown in Tables \ref{50D-3}, S-3, and S-6 in \textbf{APPENDIX A} demonstrate that HPSO achieves better performance when dealing with such complex functions. 

Specifically, HPSO ranks first on functions $F_{21}$, $F_{26}$, and $F_{29}$ across all three tested dimensions. Furthermore, it ranks at least third for nearly all the other functions on the three dimensions, except for $F_{25}$ at 30D, $F_{23}$ and $F_{24}$ at 100D. This is mainly because of the hypergraph update mechanism, which maintains swarm diversity in the early stages to help particles escape local optima, while later focusing hyperedges on promising regions to facilitate effective local exploitation.

In Tables \ref{50D-3}, S-3 and S-6 in \textbf{APPENDIX A}, HPSO achieves the best performance in 4, 5, and 3 cases out of a total of 10 functions across 30D, 50D, and 100D, respectively. Overall, the Wilcoxon rank-sum test reveals that HPSO significantly outperforms the baseline methods in 59, 60, and 52 out of the 63 test cases across 30D, 50D, and 100D, respectively. These results indicate that HPSO achieves satisfactory performance on the composition functions, where the hypergraph topology and update mechanism effectively balance exploration and exploitation within these complex landscapes.

\subsection{Summaries on the Statistical Results}
\begin{table}[]
\centering
\caption{Summary of the results of the Wilcoxon rank-sum test for the seven comparison algorithms.}
\setlength{\tabcolsep}{5pt}
\renewcommand\arraystretch{1}
\begin{tabular}{ccccccccccc}
\hline
\multirow{2}{*}{\textbf{Algorithm}} & \multicolumn{3}{c}{\textbf{30D}} & \multicolumn{3}{c}{\textbf{50D}} & \multicolumn{3}{c}{\textbf{100D}} & \multirow{2}{*}{\textbf{CP}} \\ \cline{2-10}
                                    & +          & -        & =        & +         & -        & =         & +          & -          & =        &                              \\ \hline
PSO                        & 27         & 0        & 2        & 28        & 0        & 1         & 29         & 0          & 0        & 84                           \\
CLPSO                      & 16         & 7        & 6        & 18        & 6        & 5         & 14         & 9          & 6        & 26                           \\
OLPSO                      & 25         & 1        & 3        & 28        & 0        & 1         & 25         & 2          & 2        & 75                           \\
PPSO                       & 29         & 0        & 0        & 29        & 0        & 0         & 29         & 0          & 0        & 87                           \\
XPSO                       & 20         & 2        & 7        & 21        & 2        & 6         & 22         & 3          & 4        & 56                           \\
TAPSO                      & 14         & 7        & 8        & 12        & 7        & 10        & 10         & 15         & 4        & 7                            \\
AWPSO                      & 26         & 1        & 2        & 27        & 0        & 2         & 25         & 0          & 4        & 77                           \\ \hline
\end{tabular}
\label{rank-sum}
\end{table}

\begin{table}[htbp]
\centering
\footnotesize
\caption{Friedman test for the eight algorithms on the CEC'17 benchmark suite.}
\setlength{\tabcolsep}{0.8pt}
\renewcommand\arraystretch{1}
\begin{tabular}{cccccccc}
\hline
\multicolumn{2}{c}{\textbf{Overall Rank}} & \multicolumn{2}{c}{\textbf{30D}} & \multicolumn{2}{c}{\textbf{50D}} & \multicolumn{2}{c}{\textbf{100D}} \\ \hline
Algorithm             & Rank             & Algorithm        & Rank       & Algorithm        & Rank       & Algorithm        & Rank       \\ \hline
HPSO                  & 1.91                & HPSO             & 2.14          & HPSO             & 1.79          & TAPSO            & 2.03           \\
TAPSO                 & 2.24                & CLPSO            & 2.59          & TAPSO            & 2.48          & HPSO             & 2.21           \\
CLPSO                 & 2.95                & TAPSO            & 2.72          & CLPSO            & 2.97          & CLPSO            & 2.93           \\
XPSO                  & 3.22                & XPSO             & 3.24          & XPSO             & 3.28          & XPSO             & 3.76           \\
OLPSO                 & 4.90                & OLPSO            & 5.31          & OLPSO            & 5.28          & OLPSO            & 5.03           \\
AWPSO                 & 5.36                & AWPSO            & 5.52          & AWPSO            & 5.69          & AWPSO            & 5.79           \\
PSO                   & 6.85                & PSO              & 7.10          & PSO              & 7.21          & PPSO             & 7.00           \\
PPSO                  & 6.90                & PPSO             & 7.38          & PPSO             & 7.31          & PSO              & 7.24           \\ \hline
\end{tabular}
\label{Ftest}
\end{table}

Table \ref{rank-sum} presents the summary results of the Wilcoxon rank-sum test in the three dimensions. This table summarizes the number of `+', `-', and `=' outcomes for each method across all the benchmark functions. 
The metric \textbf{CP} \cite{TAPSO} denotes the comprehensive performance, which is calculated as the number of `+' minus the number of `-'. In Table \ref{rank-sum}, HPSO significantly outperforms the other 7 baseline methods in the majority of test cases. Furthermore, as indicated by the \textbf{CP} metric, HPSO achieves a satisfactory overall performance, with TAPSO and CLPSO ranking second and third, respectively.

To compare the overall performance of the 8 methods, we conducted a Friedman test \cite{GL-PSO} on their \textbf{Mean} results in the three dimensions. The results are shown in Table \ref{Ftest}, where \textbf{Overall Rank} represents the overall Friedman test results across the three dimensions and arranges them in descending order (the lower ranking, the better performance). In Table \ref{Ftest}, HPSO achieves the best overall ranking, followed by TAPSO and CLPSO. In addition, HPSO achieves the best results on both the 30D and 50D, and closely follows the TAPSO on the 100D.

\subsection{Convergence Analysis}

\begin{figure*}[htbp]
     \centering
     \begin{subfigure}[t]{0.24\textwidth}
         \centering
         \includegraphics[height=2.5cm, trim=0 15 0 13, clip]{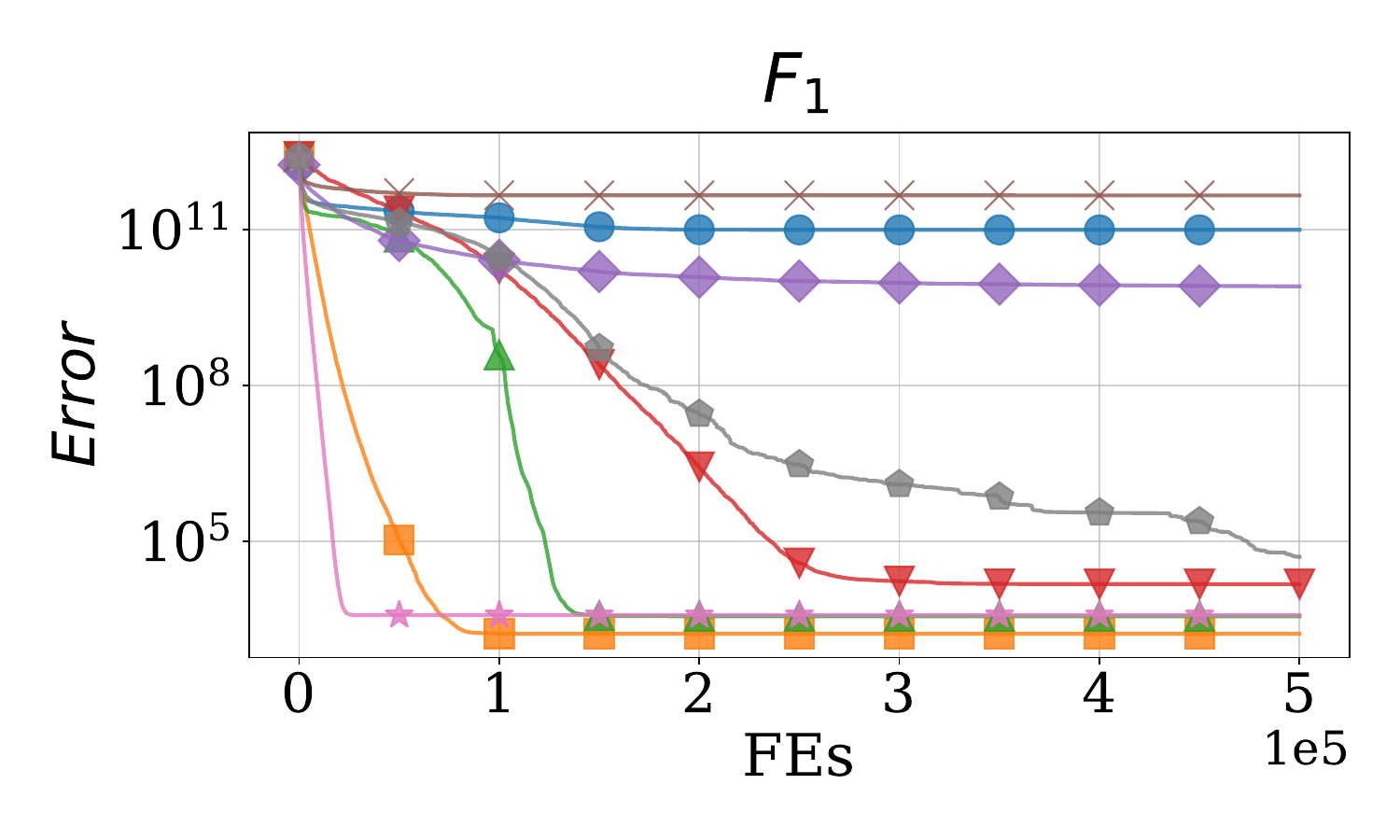}
     \end{subfigure}
     \hfill 
     \begin{subfigure}[t]{0.24\textwidth}
         \centering
         \includegraphics[height=2.5cm, trim=0 15 0 13, clip]{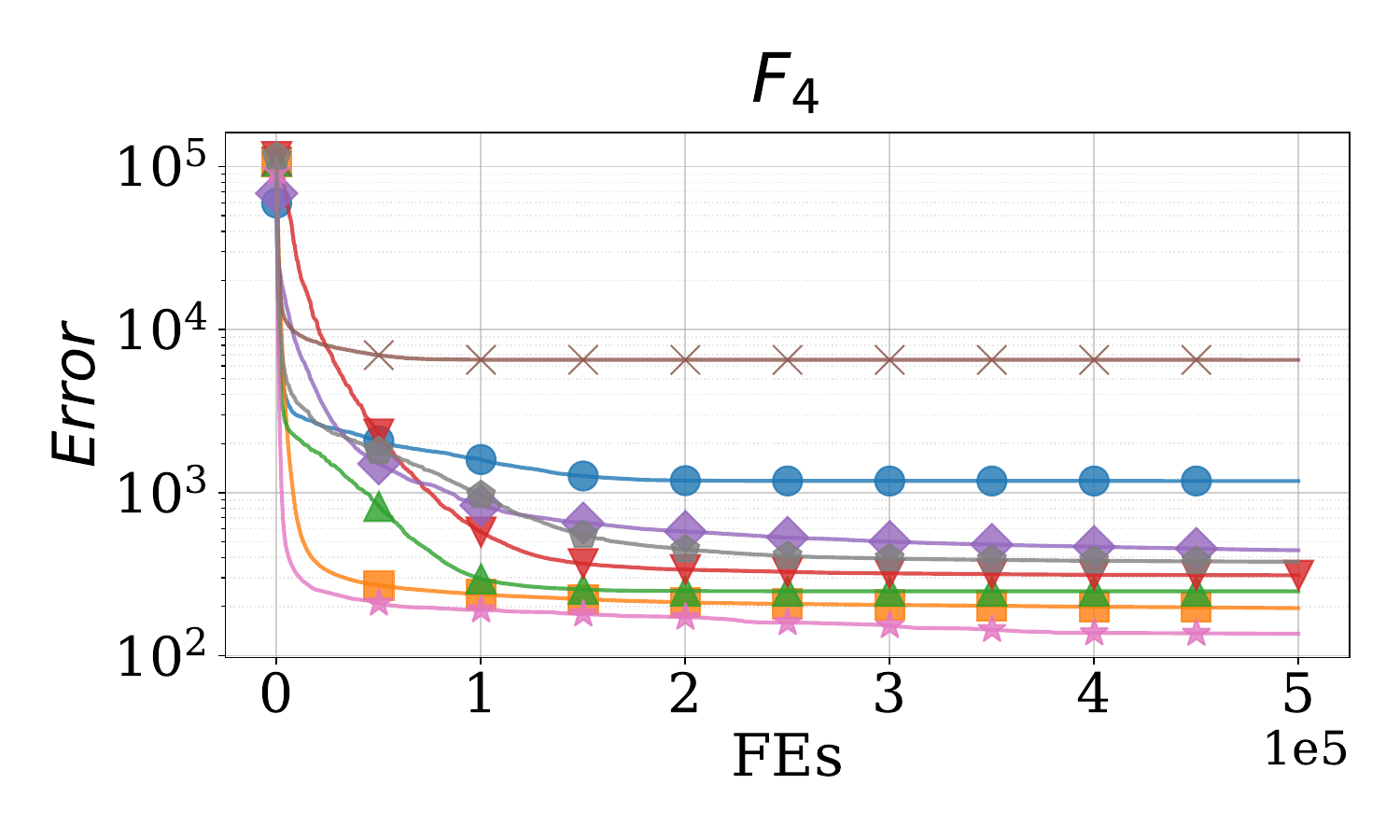}
     \end{subfigure}
     \hfill
     \begin{subfigure}[t]{0.24\textwidth}
         \centering
         \includegraphics[height=2.5cm, trim=0 15 0 13, clip]{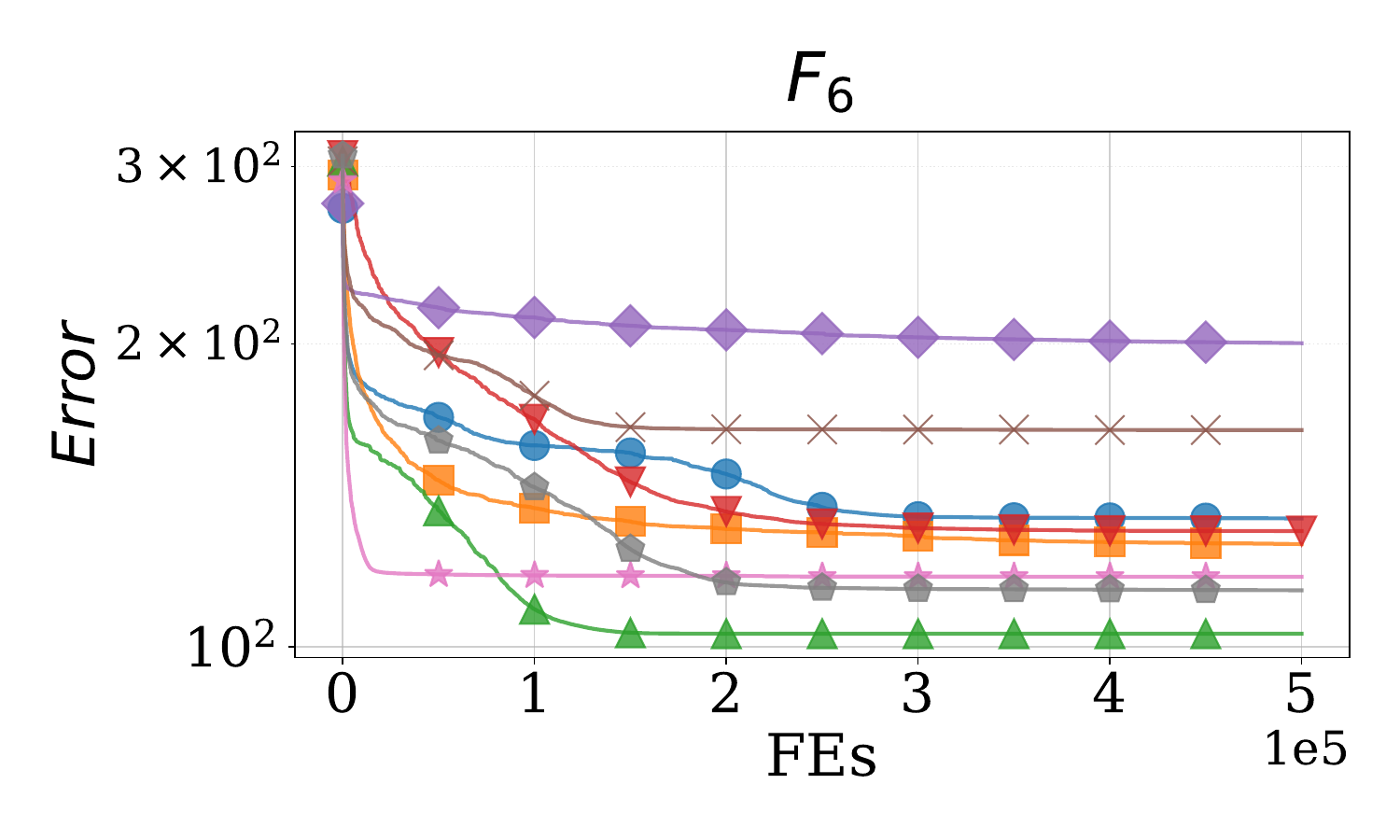}
     \end{subfigure}
     \hfill
     \begin{subfigure}[t]{0.24\textwidth}
         \centering
         \includegraphics[height=2.5cm, trim=0 15 0 13, clip]{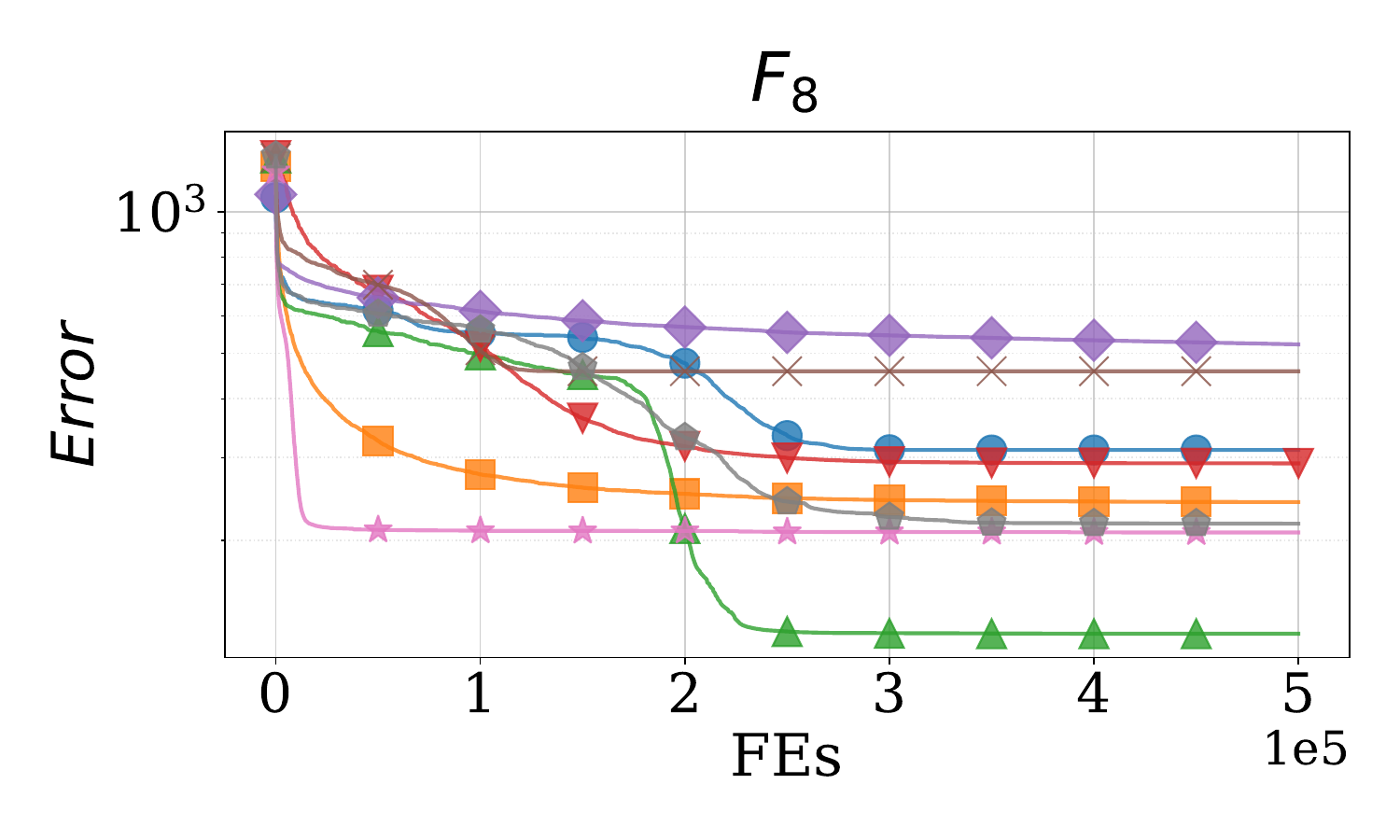}
     \end{subfigure}

      \vspace{2pt}
        
      \begin{subfigure}[t]{0.24\textwidth}
         \centering
         \includegraphics[height=2.5cm, trim=0 15 0 13, clip]{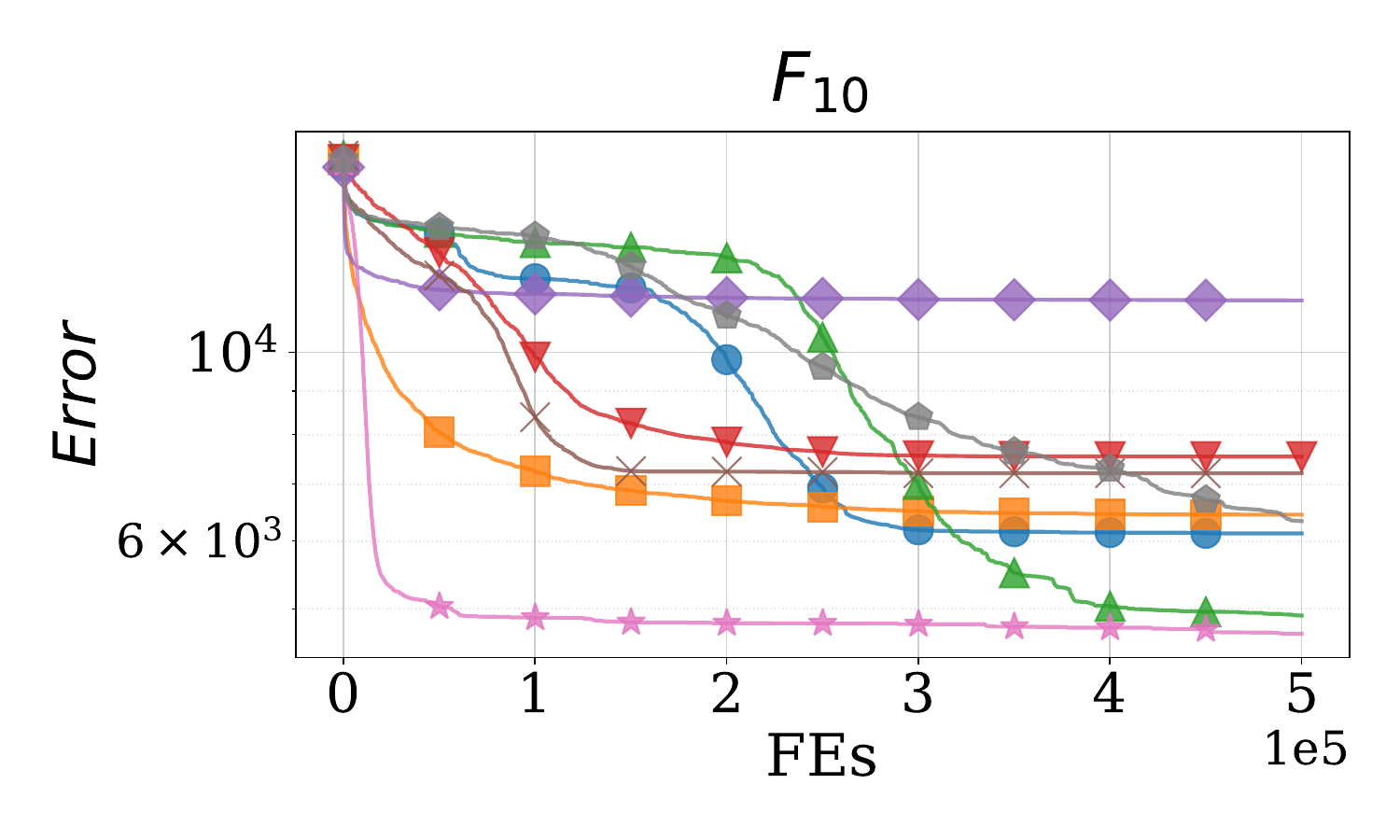}
     \end{subfigure}
     \hfill 
     \begin{subfigure}[t]{0.24\textwidth}
         \centering
         \includegraphics[height=2.5cm, trim=0 15 0 13, clip]{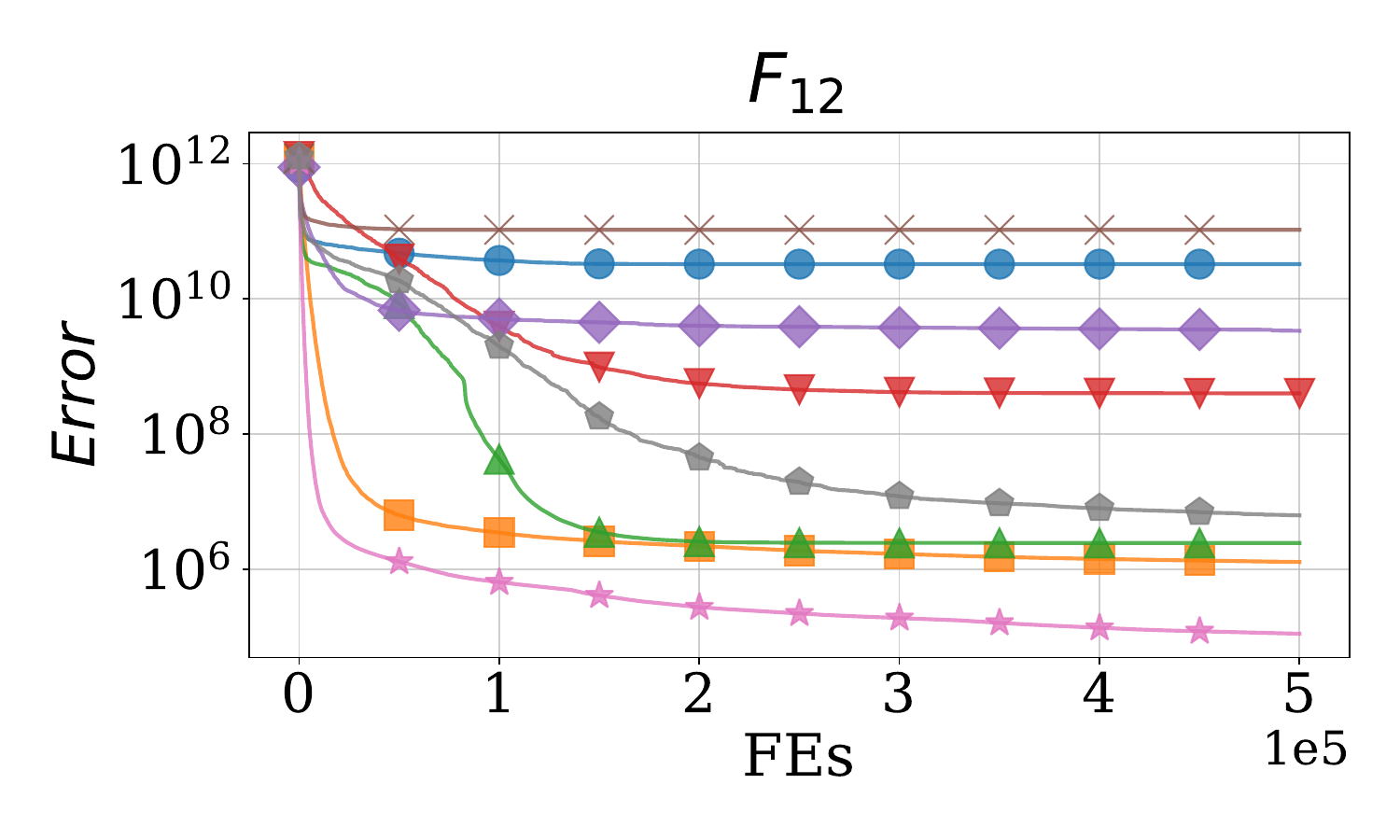}
     \end{subfigure}
     \hfill
     \begin{subfigure}[t]{0.24\textwidth}
         \centering
         \includegraphics[height=2.5cm, trim=0 15 0 13, clip]{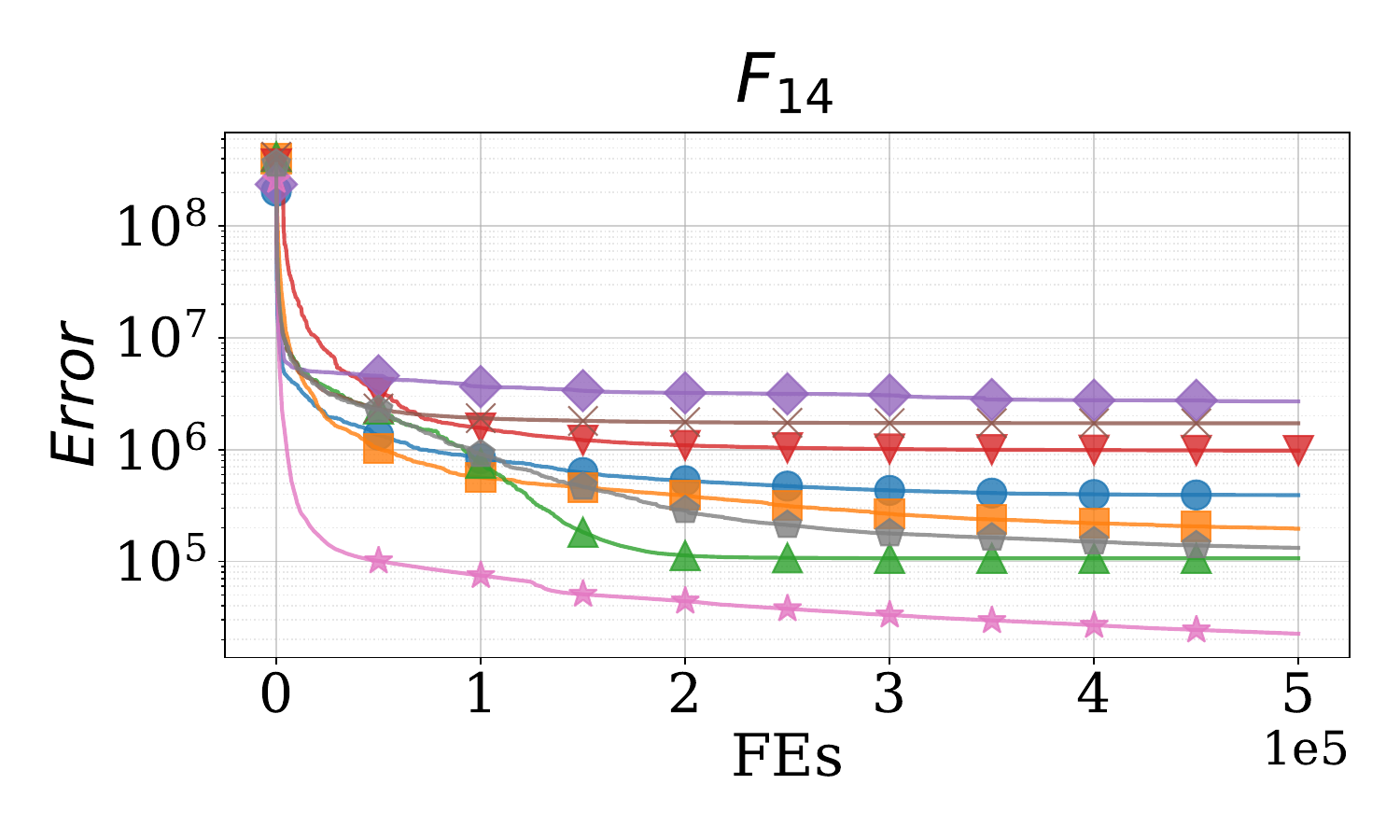}
     \end{subfigure}
     \hfill
     \begin{subfigure}[t]{0.24\textwidth}
         \centering
         \includegraphics[height=2.5cm, trim=0 15 0 13, clip]{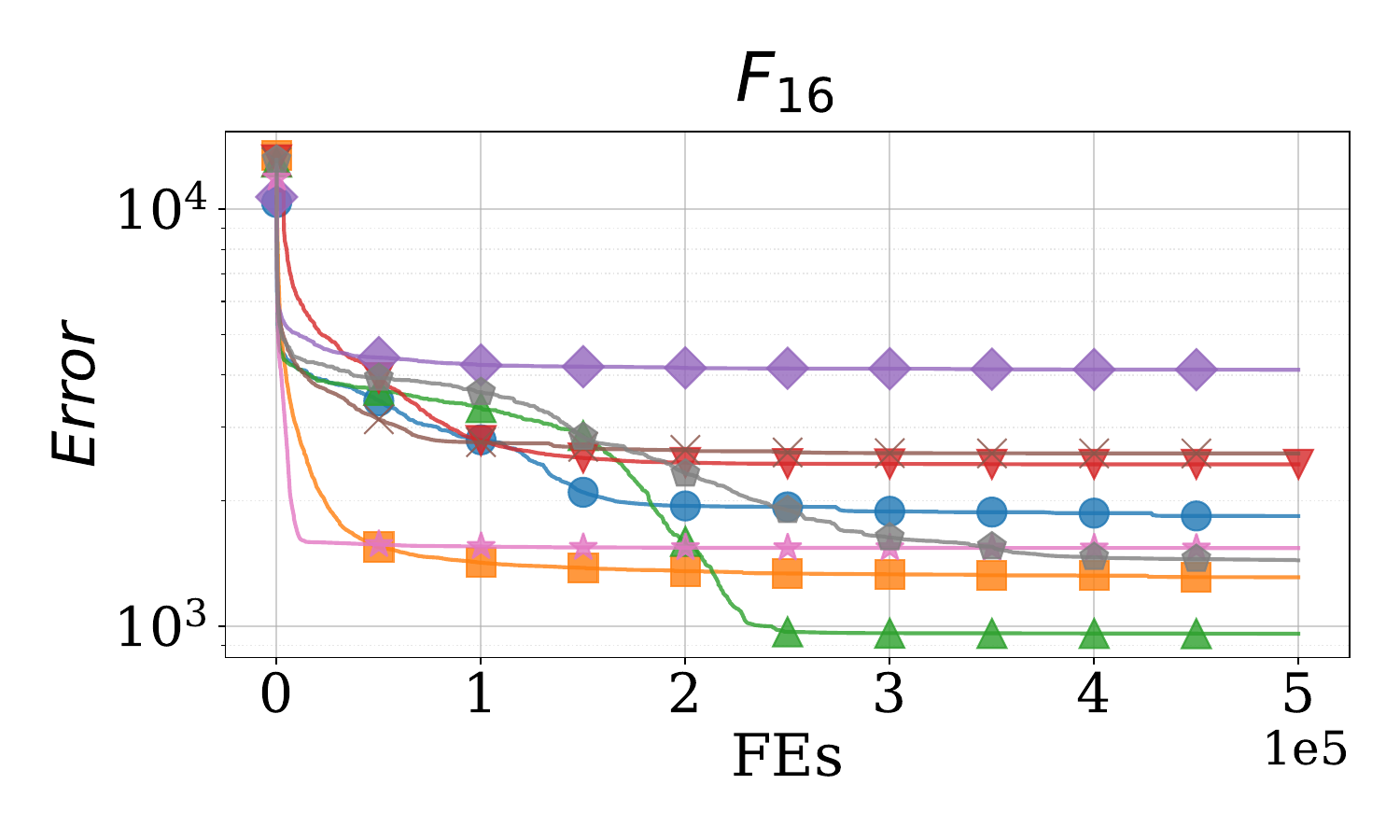}
     \end{subfigure}

      \vspace{2pt}
        
      \begin{subfigure}[t]{0.24\textwidth}
         \centering
         \includegraphics[height=2.5cm, trim=0 15 0 13, clip]{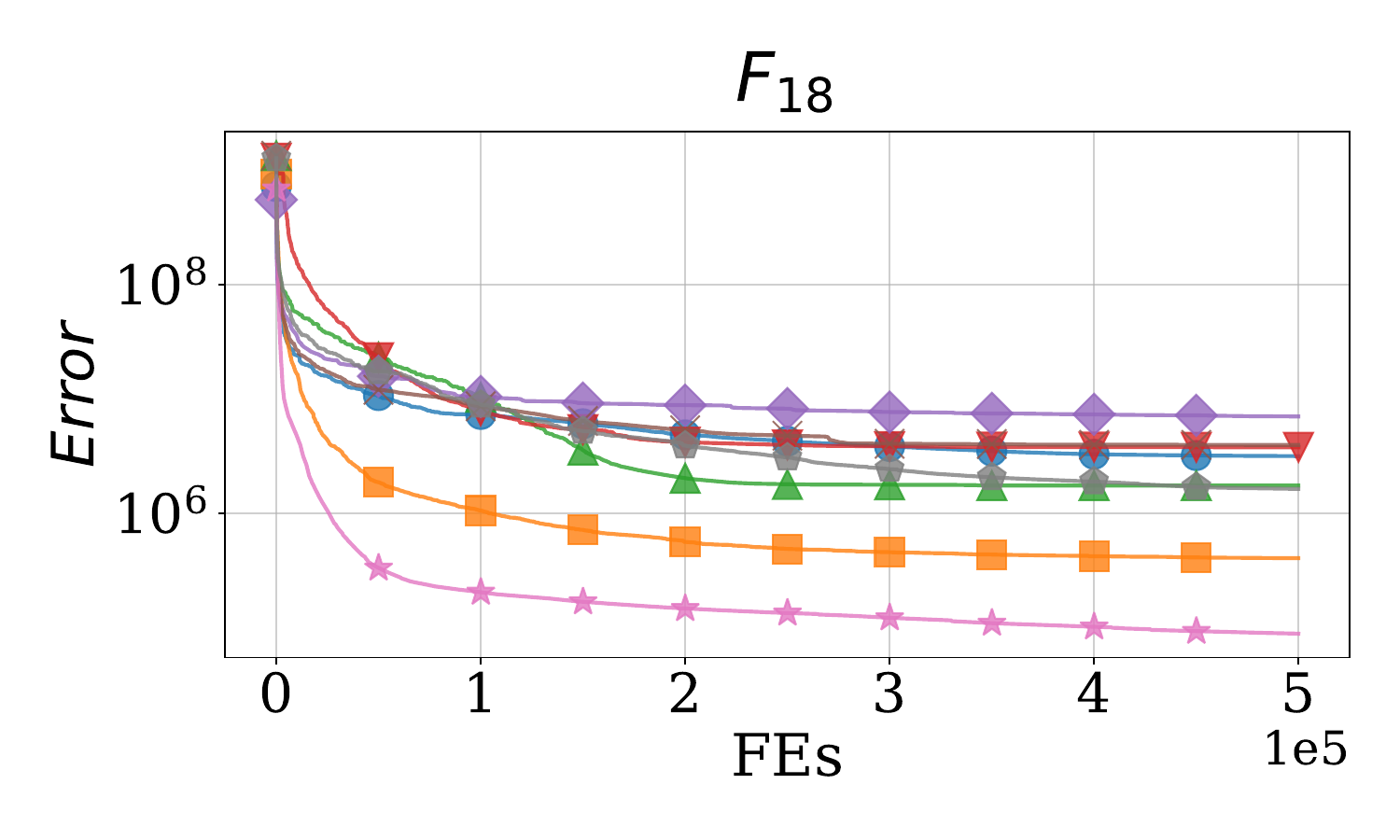}
     \end{subfigure}
     \hfill 
     \begin{subfigure}[t]{0.24\textwidth}
         \centering
         \includegraphics[height=2.5cm, trim=0 15 0 13, clip]{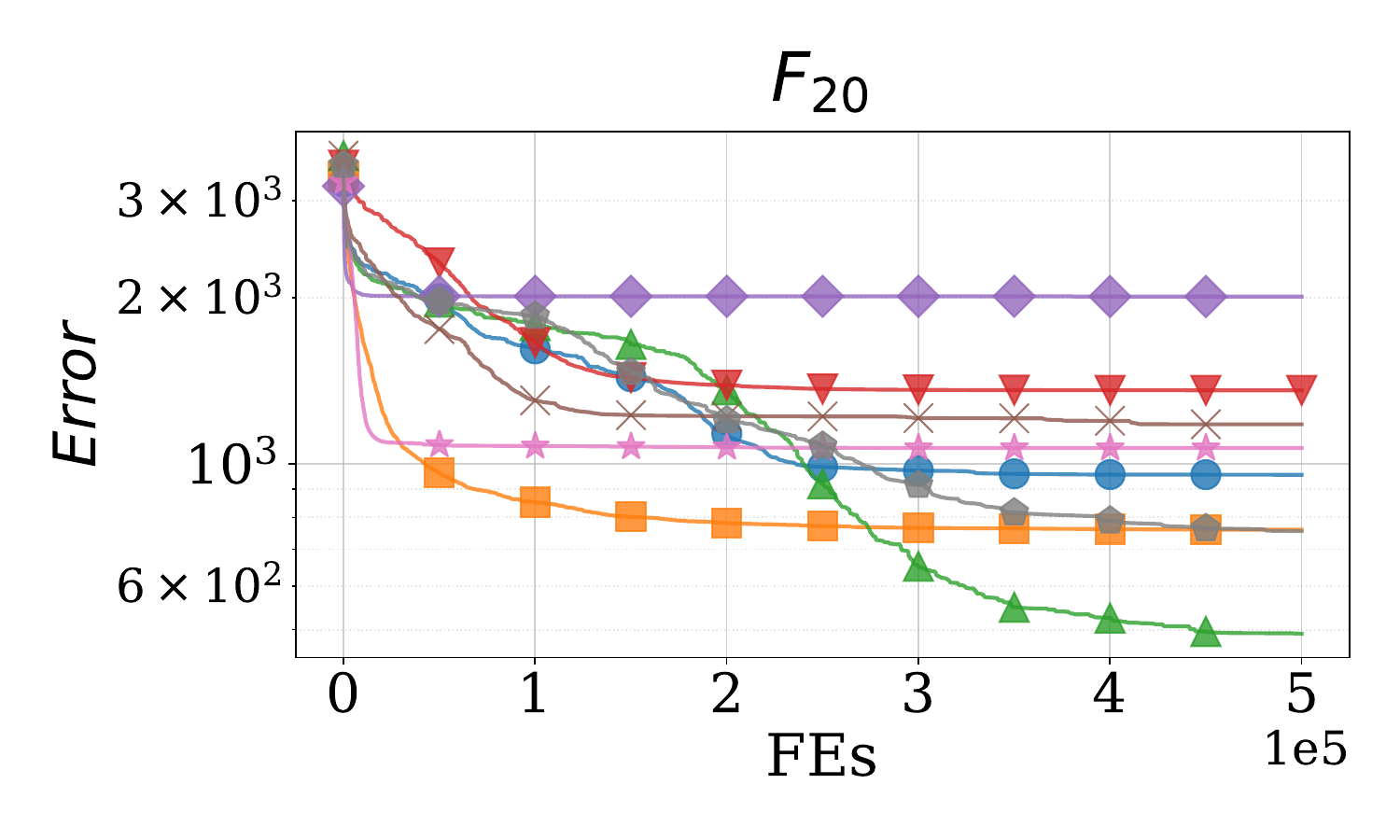}
     \end{subfigure}
     \hfill
     \begin{subfigure}[t]{0.24\textwidth}
         \centering
         \includegraphics[height=2.5cm, trim=0 15 0 13, clip]{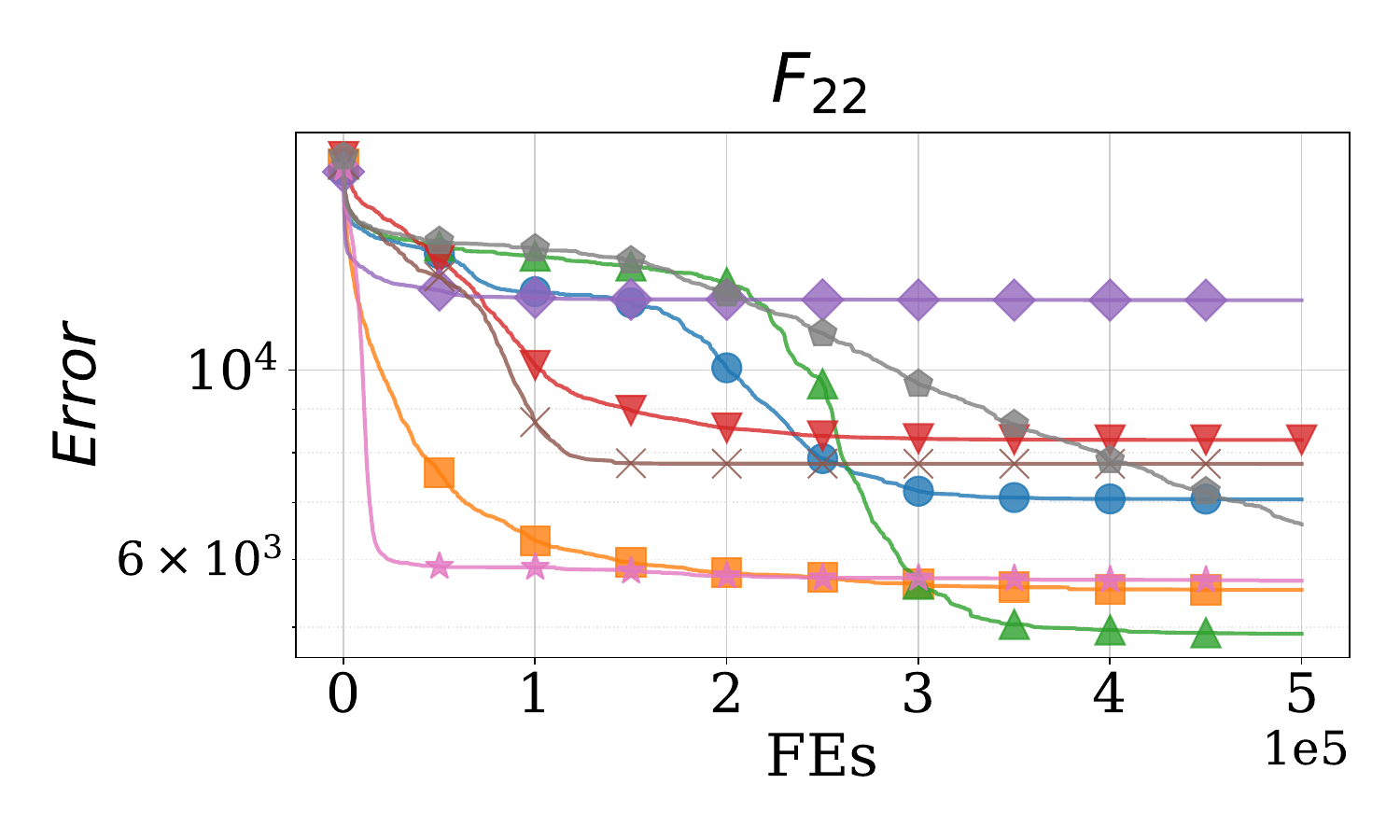}
     \end{subfigure}
     \hfill
     \begin{subfigure}[t]{0.24\textwidth}
         \centering
         \includegraphics[height=2.5cm, trim=0 15 0 13, clip]{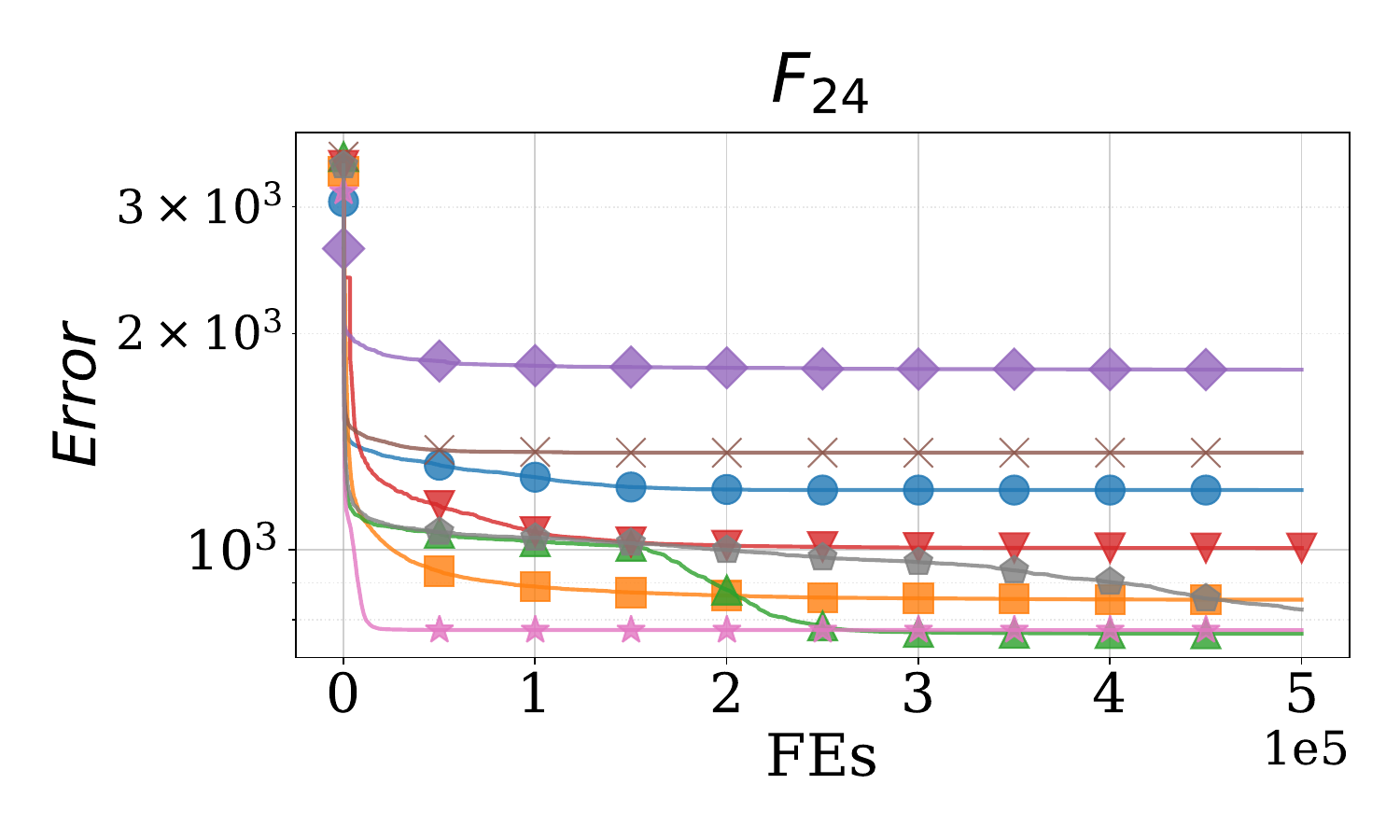}
     \end{subfigure}

      \vspace{2pt}
        
      \begin{subfigure}[t]{0.24\textwidth}
         \centering
         \includegraphics[height=2.5cm, trim=0 15 0 13, clip]{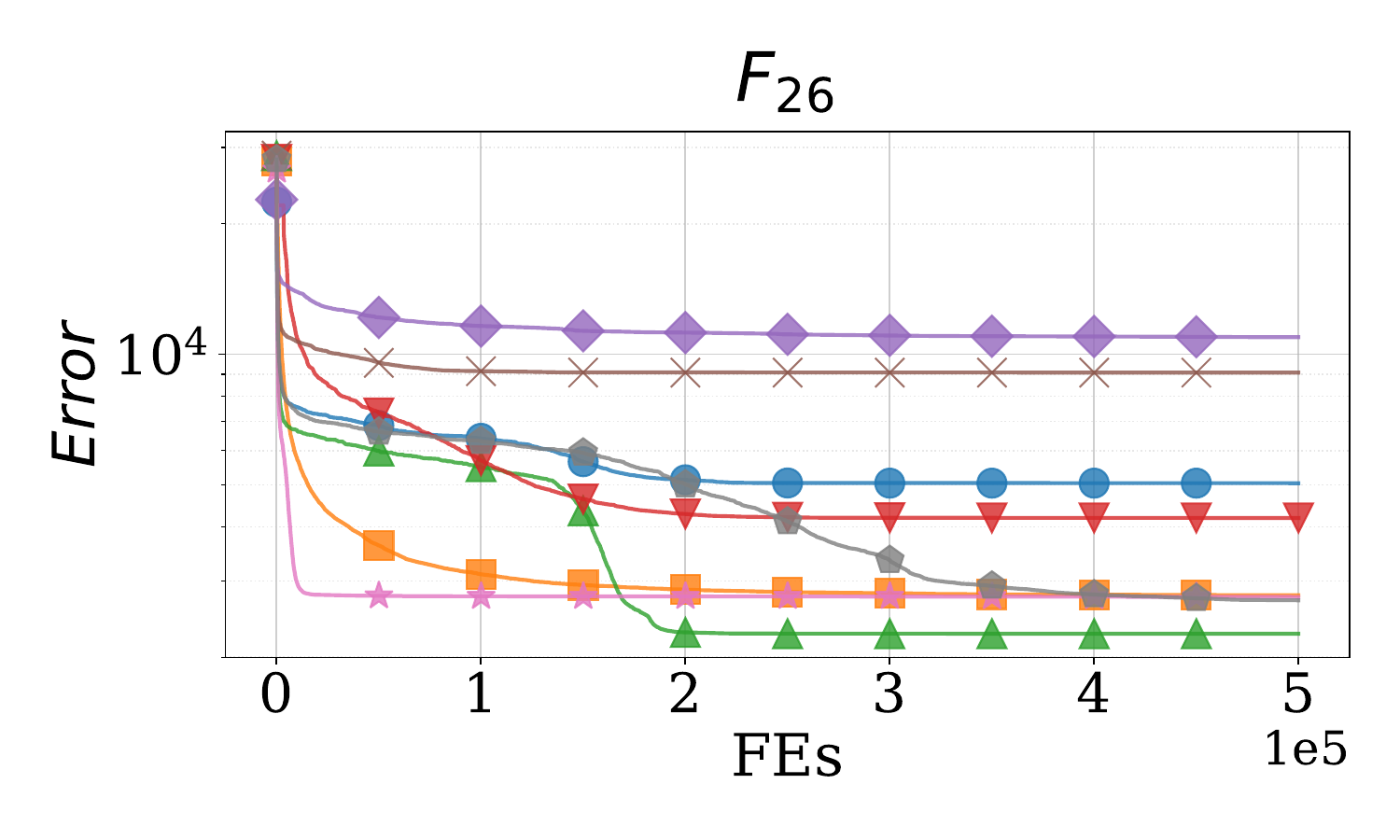}
     \end{subfigure}
     \hfill 
     \begin{subfigure}[t]{0.24\textwidth}
         \centering
         \includegraphics[height=2.5cm, trim=0 15 0 13, clip]{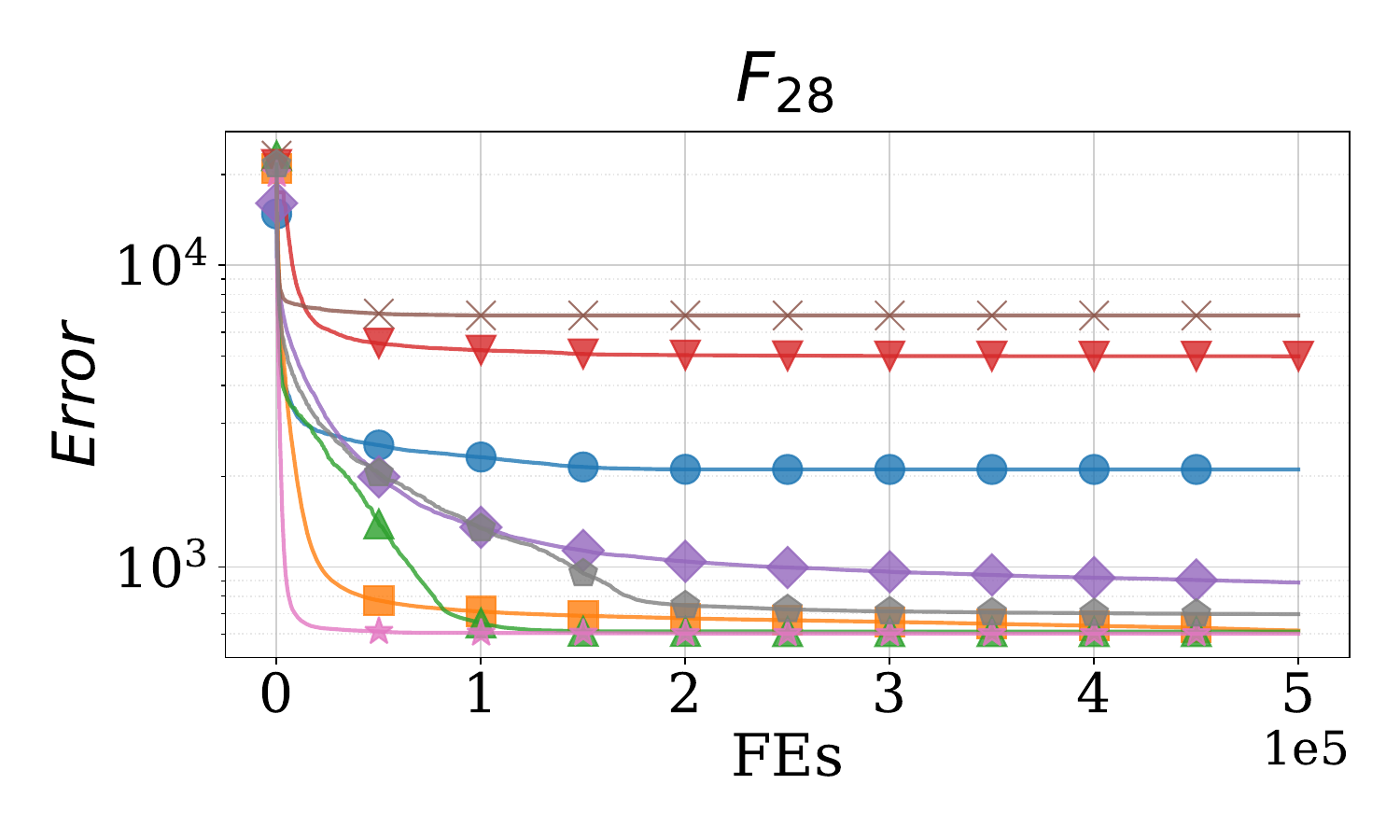}
     \end{subfigure}
     \hfill
     \begin{subfigure}[t]{0.24\textwidth}
         \centering
         \includegraphics[height=2.5cm, trim=0 15 0 13, clip]{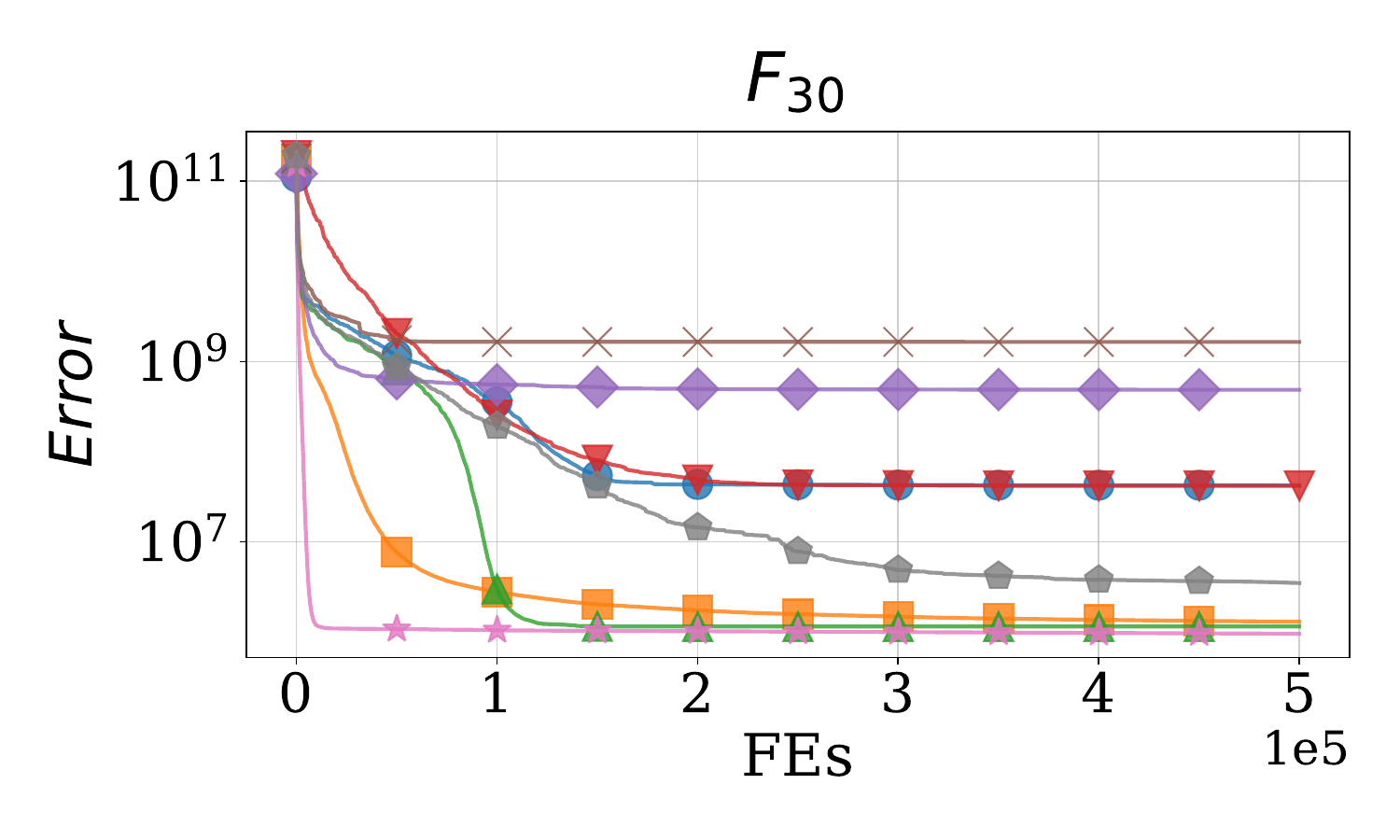}
     \end{subfigure}
     \hfill
     \begin{subfigure}[t]{0.24\textwidth}
         \centering
         \includegraphics[height=2.5cm, trim=0 15 0 13, clip]{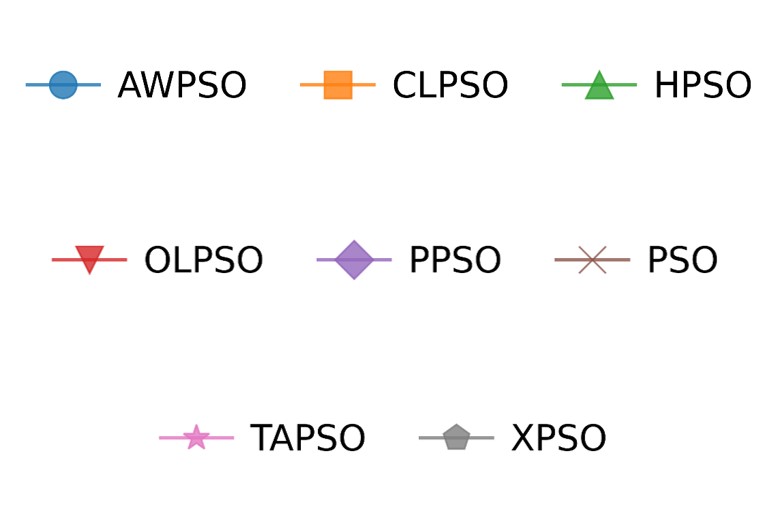}
     \end{subfigure}
     
     \caption{Mean convergence curves on 50D problems. $Error$ is defined as the logarithmic of the mean error to the best value.}
     \label{fig_curve}
\end{figure*}

\begin{figure*}[!htbp]
    \centering

    \captionsetup[subfigure]{
        labelformat=parens,
        labelfont=bf,
        singlelinecheck=off,
        justification=centering
    }

    \begin{subfigure}[t]{0.235\textwidth}
        \centering
        \includegraphics[width=\textwidth, keepaspectratio, trim=0 0 0 0, clip]{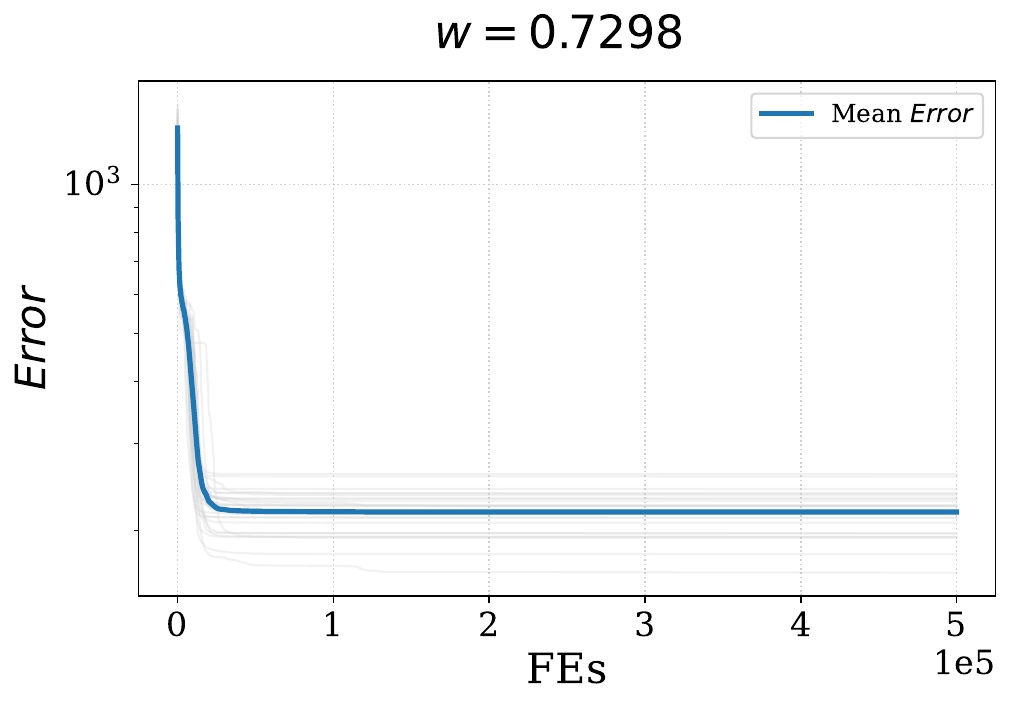}
        \caption{}
        \label{fig:sub-a}
    \end{subfigure}
    \hfill
    \begin{subfigure}[t]{0.235\textwidth}
        \centering
        \includegraphics[width=\textwidth, keepaspectratio, trim=0 0 0 0, clip]{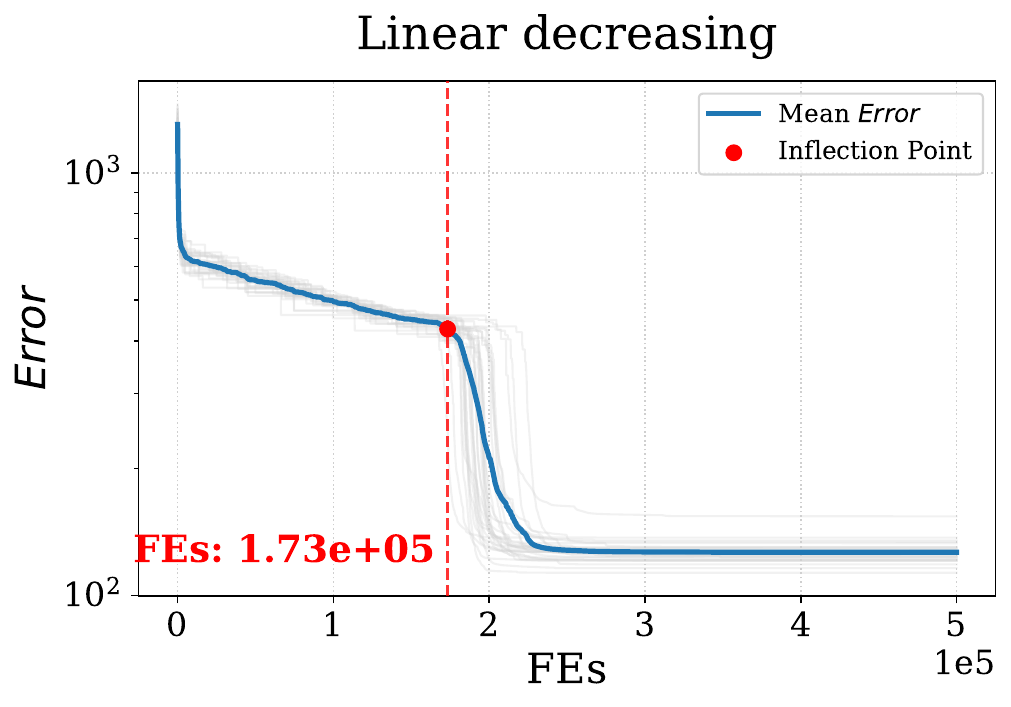}
        \caption{}
        \label{fig:sub-b}
    \end{subfigure}
    \hfill
    \begin{subfigure}[t]{0.235\textwidth}
        \centering
        \includegraphics[width=\textwidth, keepaspectratio, trim=0 0 0 0, clip]{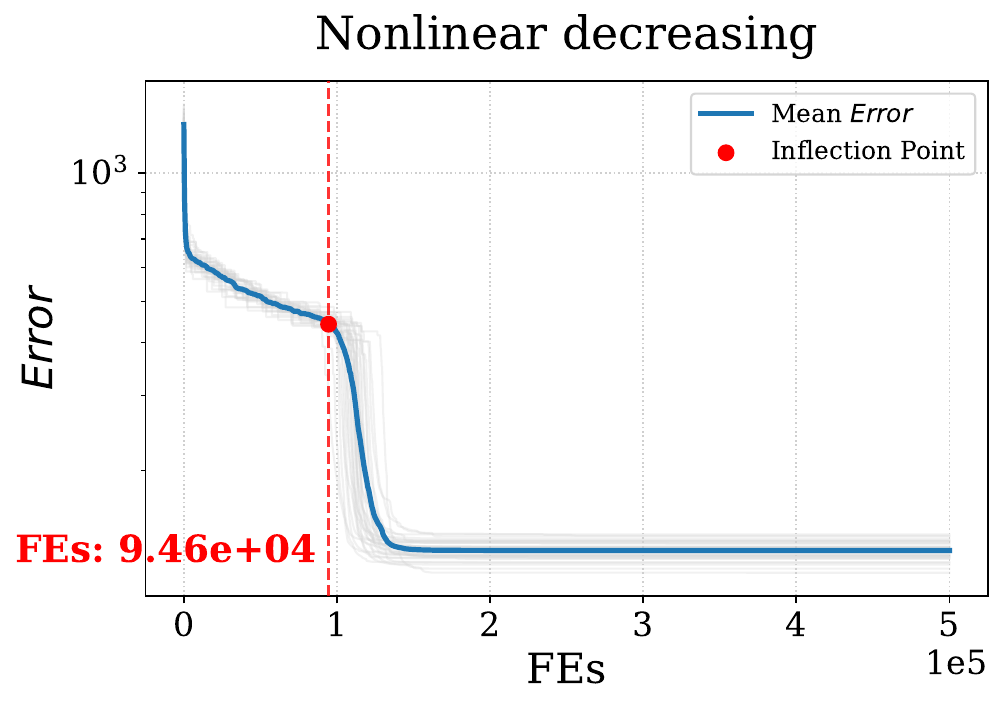}
        \caption{}
        \label{fig:sub-c}
    \end{subfigure}
    \hfill
    \begin{subfigure}[t]{0.235\textwidth}
        \centering
        \includegraphics[width=\textwidth, keepaspectratio, trim=0 0 0 0, clip]{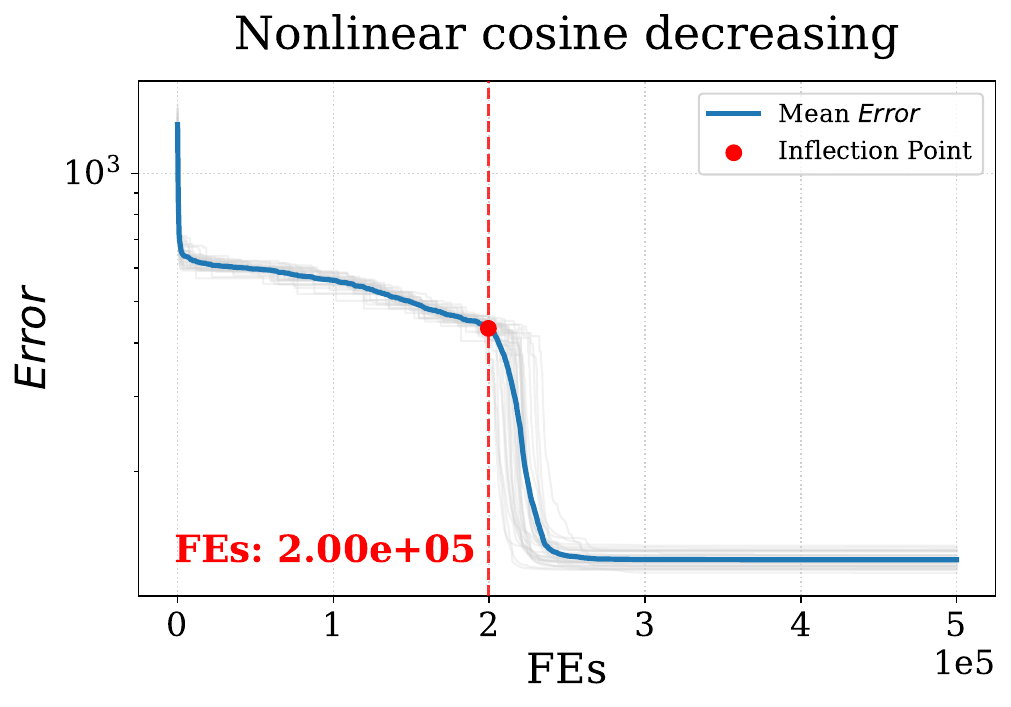}
        \caption{}
        \label{fig:sub-d}
    \end{subfigure}

    \caption{Convergence curves of HPSO with different $w$ strategies on $F_8$. 
    (a) $w$ fixed at 0.7298; (b) $w$ using linearly decreasing; 
    (c) $w$ using nonlinear decreasing; (d) $w$ using nonlinear cosine decreasing.}
    \label{fig:conAnalysis}
\end{figure*}

The convergence curves of HPSO and 7 baseline methods on 50D problems are illustrated in Fig. \ref{fig_curve}, where the $Error$ is defined as $f(x)-f (x*)$. Due to the page constraints, the convergence performance is illustrated for only a representative subset of functions. The full set of convergence curves is available in \textbf{Appendix B}.

It can be found from Fig. \ref{fig_curve} that HPSO achieves competitive performance on $F_1$ and $F_3$, indicating its strong exploitation capability. For the simple multimodal functions, HPSO achieves the best performance on five functions, including $F_5$, $F_6$, $F_7$, $F_8$, and $F_9$. This demonstrates a strong exploration capability. 
When tackling complex composition and hybrid functions, HPSO leverages its powerful global search capability to effectively handle complex variable interactions.
Even in cases where other methods fail, HPSO discovers solutions accurately. This can be attributed to its hypergraph structure, which allows each particle to acquire information from multiple peers simultaneously, thereby enhancing exploration capability and enabling escape from local optima.

Notably, Fig. \ref{fig_curve} illustrates that HPSO shows an abrupt decrease in $Error$ during the later evolutionary process for several functions. A possible reason for this decrease is the linearly decreasing strategy adopted by setting the inertia weight $w$. As $w$ changes with the evolutionary generations, it may influence the search strategy of the particle swarm. To investigate this further, we adopted 4 different $w$ settings (i.e., fixed at 0.7298, linearly decreasing, nonlinear decreasing, and nonlinear cosine decreasing) to conduct a convergence analysis of HPSO on $F_8$. The convergence curves of HPSO with different $w$ strategies on $F_8$ are shown in Fig. \ref{fig:conAnalysis}. In this figure, the red dotted line represents the FEs at the point where this phenomenon occurs (i.e., an abrupt decrease), and the blue line represents the average performance across the 30 independent runs, while the gray lines depict the individual results of each run.
As indicated in Fig. \ref{fig:conAnalysis} (a), the abrupt decrease disappears when $w$ is fixed. When adopting the decreasing strategy of $w$, as indicated in Figs. \ref{fig:conAnalysis} (b), (c), and (d), the convergence curves exhibit this decrease under the different FEs. Besides, based on the FEs at which the $Error$ significantly decreases, the corresponding value of $w$ can be estimated. In Figs. \ref{fig:conAnalysis} (b), (c), and (d), the FEs at which this phenomenon occurs are 1.73e+05, 9.46e+04, and 2.00e+05, with corresponding $w$ values of 0.727, 0.728, and 0.729, respectively. Although different decreasing strategies are adopted, the convergence curve of HPSO exhibits a significant decrease whenever $w$ reaches approximately the same value. 

Therefore, in HPSO, when $w$ decreases to a critical value, the inertial velocity no longer dominates, and the influence of both the individual and social cognitive terms on particle movement becomes stronger. This is because HPSO calculates the average position of adjacent particles within the hyperedge, therefore weakening the deviation of the particles due to learning a single exemplar in the direction of movement. Moreover, it learns the overall vector rather than a single dimension, so that a particle could move toward the target area more accurately and directly, contributing to a significant decrease in $Error$.

\subsection{Comparison of Training Time}

\begin{table*}
\footnotesize
\caption{Comparison of training time (seconds) on CEC'17 (50D). }
\setlength{\tabcolsep}{2pt}
\renewcommand\arraystretch{1.2}
\begin{tabular}{ccccccccc}
\hline
\textbf{}                                                    & \textbf{PSO} & \textbf{CLPSO} & \textbf{OLPSO} & \textbf{PPSO} & \textbf{XPSO} & \textbf{TAPSO} & \textbf{AWPSO} & \textbf{HPSO} \\ \hline
$F_{1}$                                                      & 48.235       & 110.934        & 16.978         & 29.617        & 41.879        & 111.092        & 29.603         & 123.647       \\
$F_{3}$                                                      & 52.653       & 110.484        & 25.546         & 34.514        & 46.988        & 142.105        & 42.227         & 124.158       \\
$F_{4}$                                                      & 52.395       & 108.404        & 25.976         & 32.875        & 45.174        & 170.003        & 37.406         & 121.326       \\
$F_{5}$                                                      & 48.026       & 107.832        & 21.069         & 31.705        & 44.010        & 173.326        & 35.223         & 119.197       \\
$F_{6}$                                                      & 51.028       & 109.923        & 26.374         & 34.763        & 44.876        & 186.411        & 41.856         & 121.515       \\
$F_{7}$                                                      & 58.029       & 120.903        & 44.835         & 41.231        & 53.047        & 201.326        & 62.727         & 129.272       \\
$F_{8}$                                                      & 50.661       & 112.051        & 32.367         & 33.758        & 46.033        & 207.983        & 43.461         & 122.720       \\
$F_{9}$                                                      & 54.018       & 114.789        & 36.321         & 37.239        & 49.399        & 239.598        & 52.526         & 125.755       \\
$F_{10}$                                                     & 57.420       & 120.159        & 41.386         & 37.745        & 53.998        & 254.830        & 64.081         & 135.352       \\
$F_{11}$                                                     & 78.353       & 147.106        & 86.333         & 63.685        & 74.478        & 302.559        & 124.603        & 154.574       \\
$F_{12}$                                                     & 84.859       & 150.565        & 100.580        & 70.011        & 80.994        & 313.460        & 140.319        & 159.739       \\
$F_{13}$                                                     & 86.499       & 151.416        & 88.774         & 71.705        & 82.088        & 307.853        & 145.185        & 161.634       \\
$F_{14}$                                                     & 87.959       & 152.349        & 88.261         & 72.632        & 82.878        & 294.473        & 150.426        & 163.051       \\
$F_{15}$                                                     & 82.979       & 148.266        & 79.070         & 68.462        & 78.570        & 335.209        & 141.433        & 162.307       \\
$F_{16}$                                                     & 95.749       & 168.136        & 96.764         & 78.847        & 91.660        & 320.413        & 178.575        & 216.237       \\
$F_{17}$                                                     & 122.371      & 250.312        & 136.220        & 101.744       & 116.831       & 293.912        & 243.410        & 337.938       \\
$F_{18}$                                                     & 94.852       & 273.425        & 91.335         & 76.881        & 92.078        & 247.583        & 163.071        & 303.215       \\
$F_{19}$                                                     & 108.863      & 321.068        & 119.624        & 92.140        & 106.960       & 276.552        & 215.406        & 387.554       \\
$F_{20}$                                                     & 121.242      & 402.227        & 139.519        & 106.677       & 120.242       & 297.081        & 337.253        & 465.693       \\
$F_{21}$                                                     & 100.302      & 397.030        & 104.633        & 86.964        & 99.295        & 259.089        & 332.401        & 404.422       \\
$F_{22}$                                                     & 113.634      & 428.754        & 125.988        & 96.934        & 115.353       & 280.627        & 436.423        & 422.533       \\
$F_{23}$                                                     & 130.191      & 445.280        & 153.063        & 112.599       & 131.194       & 302.746        & 636.511        & 448.310       \\
$F_{24}$                                                     & 125.463      & 414.081        & 139.938        & 107.837       & 123.590       & 289.776        & 573.986        & 458.471       \\
$F_{25}$                                                     & 156.586      & 491.399        & 160.992        & 122.763       & 146.568       & 308.419        & 624.591        & 357.023       \\
$F_{26}$                                                     & 242.580      & 374.289        & 186.674        & 135.562       & 201.073       & 335.108        & 704.469        & 379.268       \\
$F_{27}$                                                     & 283.214      & 395.666        & 206.180        & 152.943       & 282.535       & 335.085        & 627.724        & 398.901       \\
$F_{28}$                                                     & 288.170      & 391.477        & 247.624        & 180.689       & 283.645       & 368.872        & 608.729        & 389.994       \\
$F_{29}$                                                     & 587.559      & 561.555        & 555.465        & 426.657       & 552.906       & 537.304        & 1082.985       & 550.240       \\
$F_{30}$                                                     & 563.774      & 527.731        & 439.316        & 443.330       & 576.901       & 442.861        & 998.392        & 528.180       \\ \hline
\textbf{\begin{tabular}[c]{@{}c@{}}Avg.\\ Time\end{tabular}} & 138.885      & 262.331        & 124.731        & 102.845       & 133.284       & 280.540        & 306.035        & 274.904       \\ \hline
\end{tabular}
\label{time}
\end{table*}

\begin{figure*}[]
    \centering

    \begin{subfigure}[t]{0.22\textwidth}
        \centering
        \includegraphics[
            width=\linewidth,
            trim=0 0 40 20,
            clip
        ]{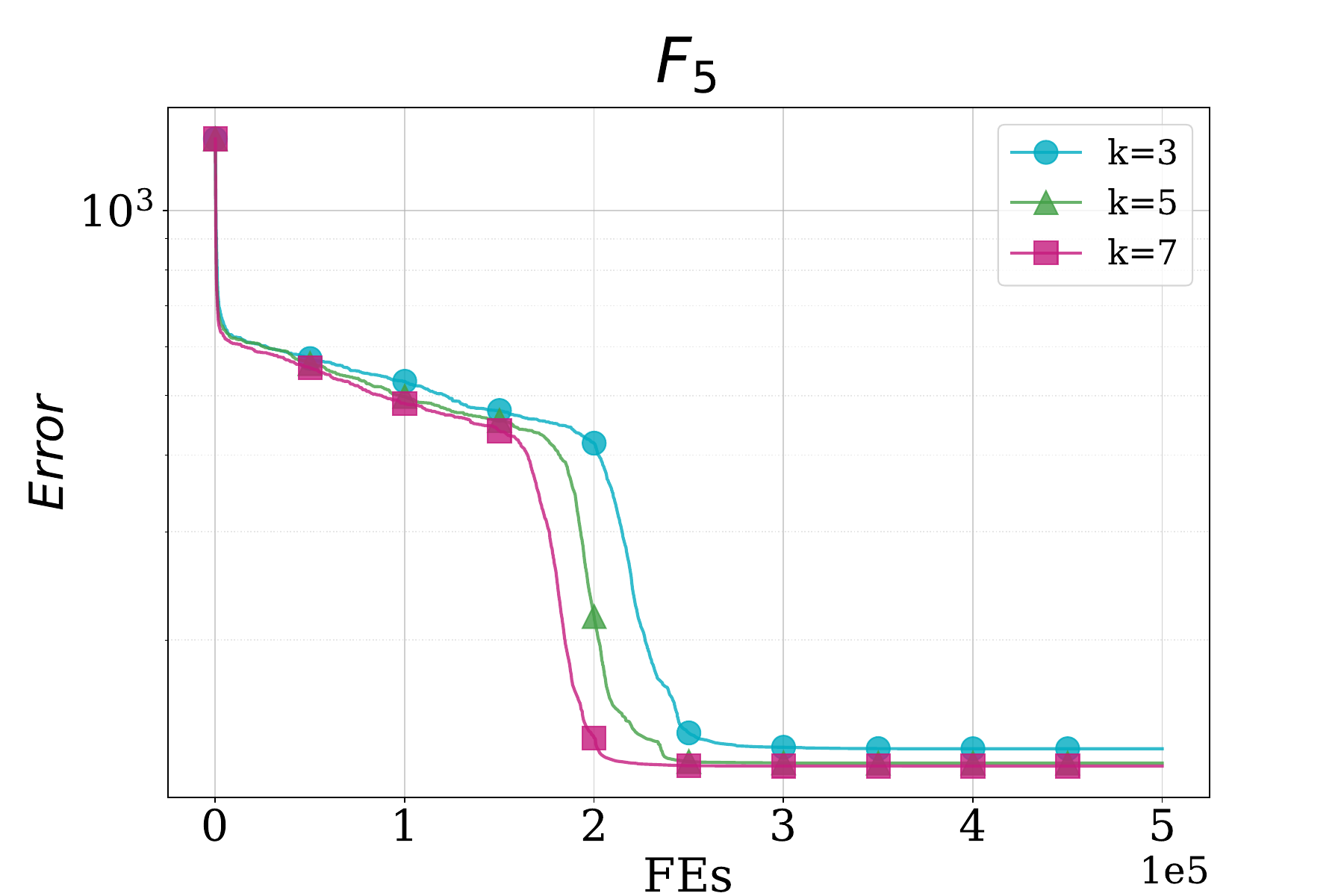}
    \end{subfigure}
    \hspace{0.001\textwidth}
    \begin{subfigure}[t]{0.22\textwidth}
        \centering
        \includegraphics[
            width=\linewidth,
            trim=0 0 40 20,
            clip
        ]{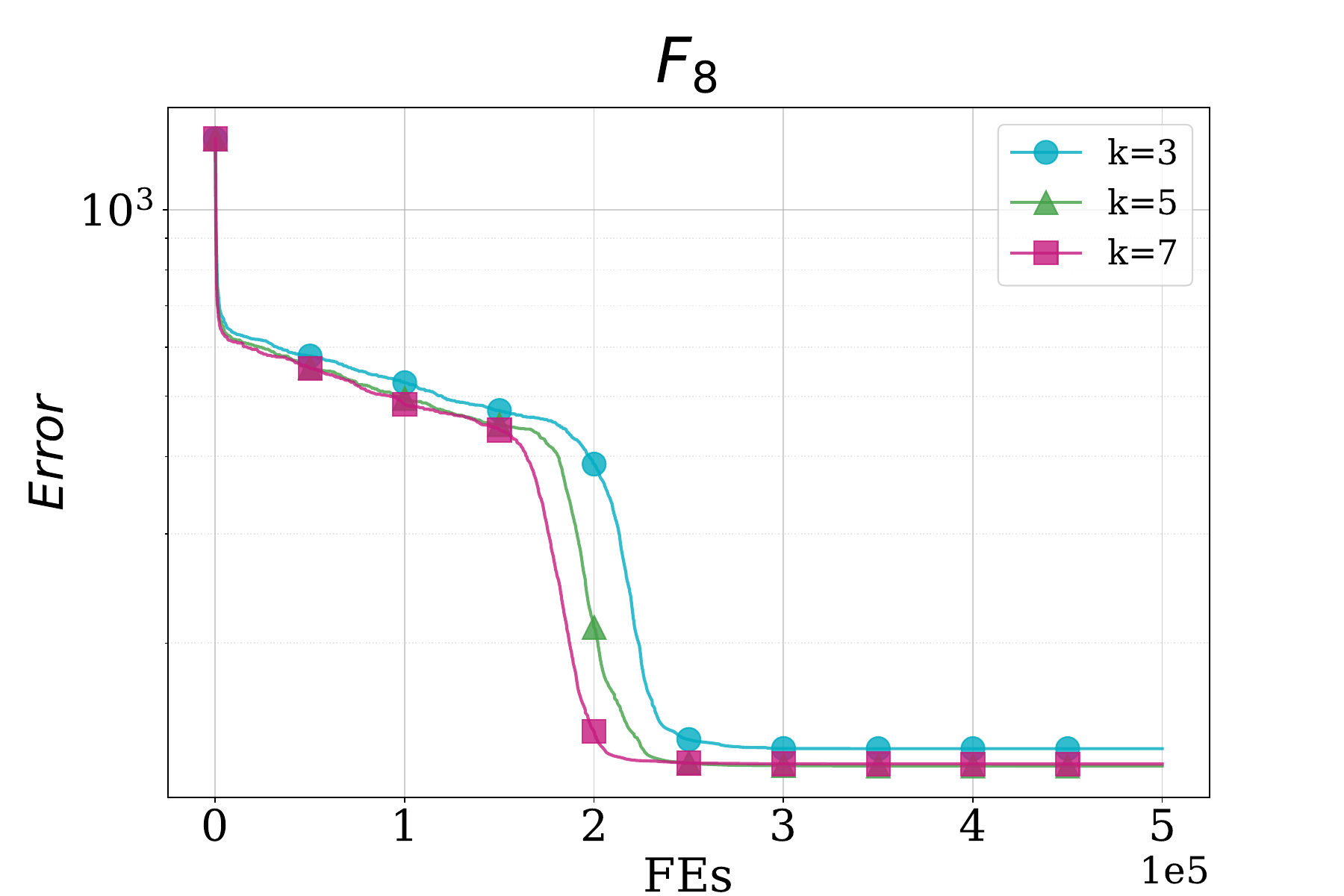}
    \end{subfigure}
    \hspace{0.001\textwidth}
    \begin{subfigure}[t]{0.22\textwidth}
        \centering
        \includegraphics[
            width=\linewidth,
            trim=0 0 40 20,
            clip
        ]{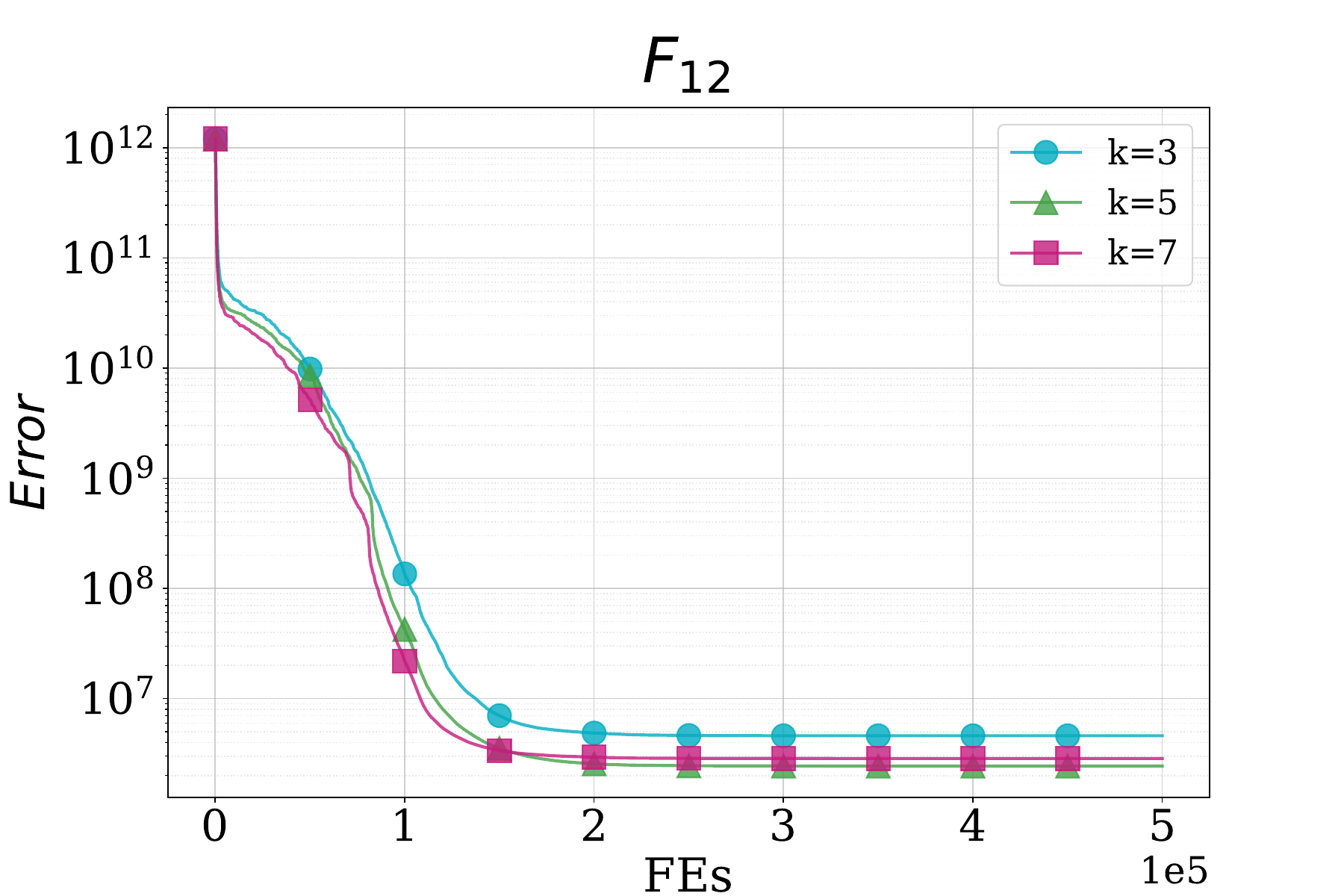}
    \end{subfigure}
    \hspace{0.001\textwidth}
    \begin{subfigure}[t]{0.23\textwidth}
        \centering
        \includegraphics[
            width=\linewidth,
            trim=0 0 40 20,
            clip
        ]{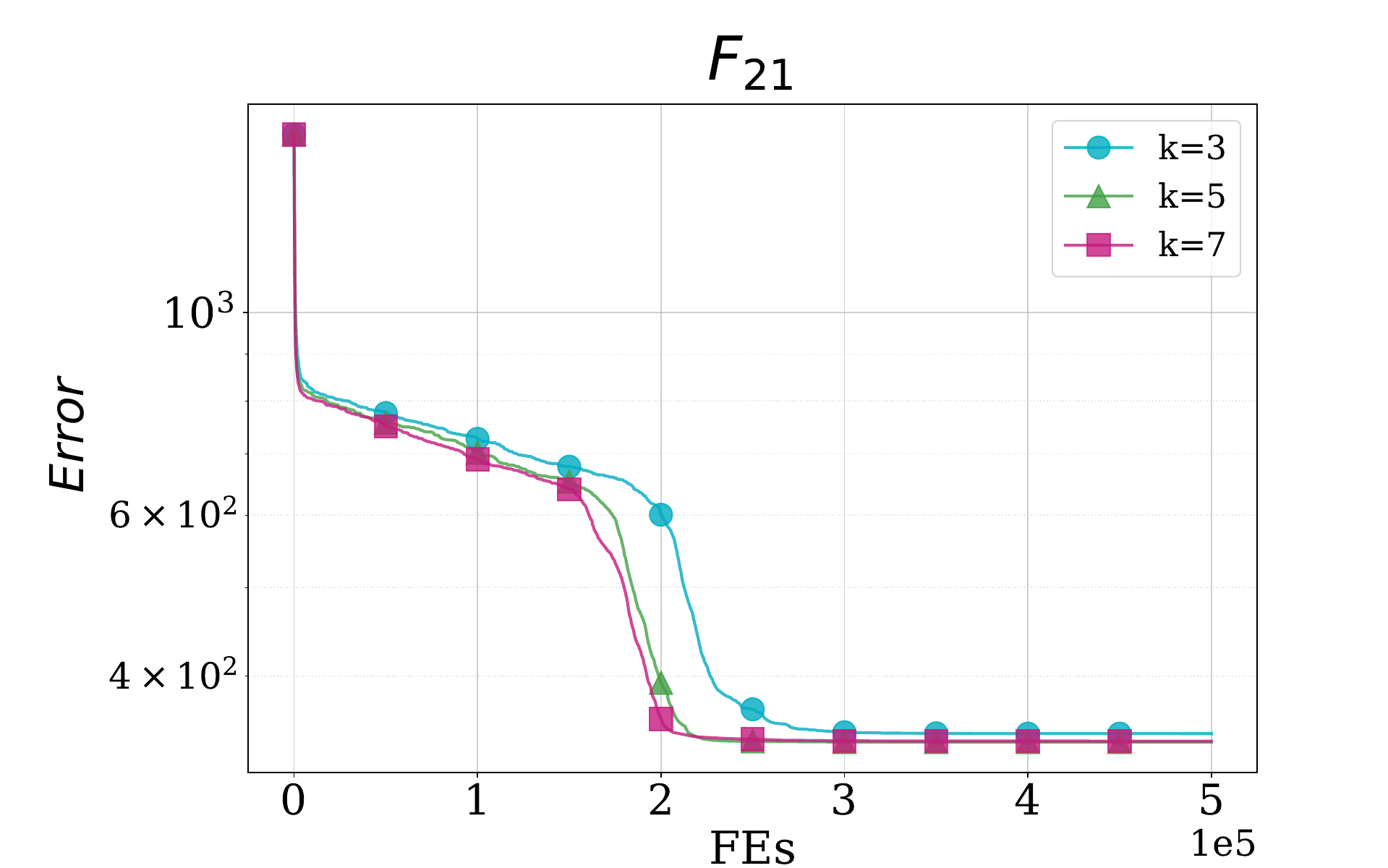}
    \end{subfigure}

    \caption{Mean convergence curves for different $k$ values on $F_5$, $F_8$, $F_{12}$ and $F_{21}$.}
    \label{param_c_curve}
\end{figure*}

In this section, we compare the time consumption of 8 methods on the 50D of CEC'17. The results are presented in Table \ref{time}. \textbf{Avg.Time} calculates the average time consumed across the 29 functions. According to Table \ref{time}, among the eight methods, PPSO achieves the fastest computational speed, while AWPSO is the slowest. HPSO ranks sixth in terms of computational cost. This is due mainly to the periodic reconstruction of the hypergraph topology during the evolutionary process, which introduces an additional overhead. However, compared to TAPSO, which also achieves high performance, HPSO exhibits a lower \textbf{Avg.Time}.

\subsection{Parameter Analysis}

In this section, the parameter $k$ in the KNN hypergraph construction stage of HPSO is further investigated. Specifically, we conduct experiments by varying the value of $k=[3,5,7]$. Table \ref{param_k} presents the \textbf{Mean} and \textbf{Std} values of HPSO for different $k$ at 50D. 
As shown in Table \ref{param_k}, HPSO achieves significantly better performance across most functions when $k=5$, particularly on simple multimodal functions. Furthermore, with 
$k=5$, HPSO achieves the best performance on 15 out of the 29 benchmark functions.

To further investigate the influence of $k$ on HPSO, we compared the convergence curves of four functions with different $k$ values at 50D, as shown in Fig. \ref{param_c_curve}.
It can be observed that the convergence rate varies with different $k$ values from Fig. \ref{param_c_curve}. Specifically, a larger $k$ accelerates convergence because the larger hyperedges increase the number of neighbors for each particle, which significantly reduces the average path length of the population topology. However, this also increases the risk of trapping in local optima due to the accelerated loss of swarm diversity. Therefore, $k=5$ achieves a good balance between exploration and exploitation, providing an optimal configuration for HPSO.

\begin{table}[]
\centering
\caption{Results for different $k$ values on the CEC'17 benchmark suite (50D).}
\setlength{\tabcolsep}{8pt}
\renewcommand\arraystretch{0.7}
\footnotesize
\label{param_k}
\begin{tabular}{cclll}
\hline
\textbf{}                          &               & \textbf{k=3}        & \textbf{k=7}        & \textbf{k=5}      \\ \hline
\multirow{2}{*}{\textbf{$F_1$}}    & \textbf{Mean} & 6.85E+03 =          & 2.28E+09 +          & \textbf{3.73E+03} \\
                                   & \textbf{Std}  & 8.86E+03            & 1.23E+10            & 3.51E+03          \\
\multirow{2}{*}{\textbf{$F_3$}}    & \textbf{Mean} & 7.69E+04 +          & 3.76E+04 +          & \textbf{3.24E+04} \\
                                   & \textbf{Std}  & 3.43E+04            & 6.86E+03            & 5.29E+03          \\
\multirow{2}{*}{\textbf{$F_4$}}    & \textbf{Mean} & 5.68E+02 =          & 5.77E+02 +          & \textbf{5.48E+02} \\
                                   & \textbf{Std}  & 3.20E+01            & 3.78E+01            & 4.48E+01          \\
\multirow{2}{*}{\textbf{$F_5$}}    & \textbf{Mean} & 5.33E+02 +          & 5.25E+02 =          & \textbf{5.26E+02} \\
                                   & \textbf{Std}  & 9.40E+00            & 6.13E+00            & 6.42E+00          \\
\multirow{2}{*}{\textbf{$F_6$}}    & \textbf{Mean} & \textbf{6.03E+02 =} & \textbf{6.03E+02 =} & \textbf{6.03E+02} \\
                                   & \textbf{Std}  & 1.23E+00            & 8.44E-01            & 1.02E+00          \\
\multirow{2}{*}{\textbf{$F_7$}}    & \textbf{Mean} & 7.89E+02 +          & 7.87E+02 +          & \textbf{7.82E+02} \\
                                   & \textbf{Std}  & 1.44E+01            & 7.95E+00            & 7.92E+00          \\
\multirow{2}{*}{\textbf{$F_8$}}    & \textbf{Mean} & 8.35E+02 +          & 8.28E+02 =          & \textbf{8.27E+02} \\
                                   & \textbf{Std}  & 1.13E+01            & 6.05E+00            & 7.32E+00          \\
\multirow{2}{*}{\textbf{$F_9$}}    & \textbf{Mean} & 9.65E+02 =          & \textbf{9.27E+02 =} & 9.42E+02          \\
                                   & \textbf{Std}  & 8.54E+01            & 2.92E+01            & 5.18E+01          \\
\multirow{2}{*}{\textbf{$F_{10}$}} & \textbf{Mean} & 9.03E+03 +          & 5.85E+03 =          & \textbf{5.81E+03} \\
                                   & \textbf{Std}  & 1.01E+03            & 1.11E+03            & 9.21E+02          \\
\multirow{2}{*}{\textbf{$F_{11}$}} & \textbf{Mean} & 1.23E+03 +          & \textbf{1.21E+03 =} & \textbf{1.21E+03} \\
                                   & \textbf{Std}  & 4.98E+01            & 4.18E+01            & 3.75E+01          \\
\multirow{2}{*}{\textbf{$F_{12}$}} & \textbf{Mean} & 4.60E+06 +          & 2.86E+06 =          & \textbf{2.44E+06} \\
                                   & \textbf{Std}  & 3.92E+06            & 1.72E+06            & 1.71E+06          \\
\multirow{2}{*}{\textbf{$F_{13}$}} & \textbf{Mean} & 1.78E+04 =          & \textbf{1.06E+04 =} & 1.11E+04          \\
                                   & \textbf{Std}  & 1.93E+04            & 9.57E+03            & 1.01E+04          \\
\multirow{2}{*}{\textbf{$F_{14}$}} & \textbf{Mean} & 1.52E+05 =          & 1.18E+05 =          & \textbf{1.08E+05} \\
                                   & \textbf{Std}  & 8.42E+04            & 5.53E+04            & 5.04E+04          \\
\multirow{2}{*}{\textbf{$F_{15}$}} & \textbf{Mean} & 8.57E+03 =          & \textbf{7.70E+03 =} & 8.66E+03          \\
                                   & \textbf{Std}  & 5.95E+03            & 4.48E+03            & 5.11E+03          \\
\multirow{2}{*}{\textbf{$F_{16}$}} & \textbf{Mean} & 2.48E+03 =          & \textbf{2.38E+03 =} & 2.46E+03          \\
                                   & \textbf{Std}  & 2.93E+02            & 3.33E+02            & 3.82E+02          \\
\multirow{2}{*}{\textbf{$F_{17}$}} & \textbf{Mean} & 2.36E+03 =          & 2.36E+03 =          & \textbf{2.28E+03} \\
                                   & \textbf{Std}  & 2.17E+02            & 2.08E+02            & 2.04E+02          \\
\multirow{2}{*}{\textbf{$F_{18}$}} & \textbf{Mean} & \textbf{1.61E+06 =} & 1.82E+06 =          & 1.75E+06          \\
                                   & \textbf{Std}  & 1.17E+06            & 8.71E+05            & 1.15E+06          \\
\multirow{2}{*}{\textbf{$F_{19}$}} & \textbf{Mean} & \textbf{1.15E+04 -} & 1.70E+04 =          & 1.75E+04          \\
                                   & \textbf{Std}  & 1.13E+04            & 9.47E+03            & 1.19E+04          \\
\multirow{2}{*}{\textbf{$F_{20}$}} & \textbf{Mean} & 2.57E+03 +          & 2.40E+03 =          & \textbf{2.39E+03} \\
                                   & \textbf{Std}  & 2.17E+02            & 1.95E+02            & 2.75E+02          \\
\multirow{2}{*}{\textbf{$F_{21}$}} & \textbf{Mean} & 2.35E+03 +          & \textbf{2.34E+03 =} & \textbf{2.34E+03} \\
                                   & \textbf{Std}  & 1.36E+01            & 1.22E+01            & 9.34E+00          \\
\multirow{2}{*}{\textbf{$F_{22}$}} & \textbf{Mean} & 9.92E+03 +          & \textbf{6.79E+03 =} & 7.01E+03          \\
                                   & \textbf{Std}  & 1.32E+03            & 9.57E+02            & 1.02E+03          \\
\multirow{2}{*}{\textbf{$F_{23}$}} & \textbf{Mean} & \textbf{2.84E+03 -} & 2.87E+03 =          & 2.87E+03          \\
                                   & \textbf{Std}  & 3.85E+01            & 4.23E+01            & 5.61E+01          \\
\multirow{2}{*}{\textbf{$F_{24}$}} & \textbf{Mean} & 3.06E+03 =          & \textbf{3.05E+03 =} & 3.06E+03          \\
                                   & \textbf{Std}  & 5.87E+01            & 4.46E+01            & 4.83E+01          \\
\multirow{2}{*}{\textbf{$F_{25}$}} & \textbf{Mean} & \textbf{3.02E+03 -} & 3.06E+03 +          & 3.04E+03          \\
                                   & \textbf{Std}  & 3.20E+01            & 2.26E+01            & 2.81E+01          \\
\multirow{2}{*}{\textbf{$F_{26}$}} & \textbf{Mean} & 4.85E+03 =          & \textbf{4.60E+03 -} & 4.77E+03          \\
                                   & \textbf{Std}  & 4.01E+02            & 4.71E+02            & 4.23E+02          \\
\multirow{2}{*}{\textbf{$F_{27}$}} & \textbf{Mean} & 3.39E+03 =          & \textbf{3.34E+03 -} & 3.41E+03          \\
                                   & \textbf{Std}  & 1.12E+02            & 7.88E+01            & 1.09E+02          \\
\multirow{2}{*}{\textbf{$F_{28}$}} & \textbf{Mean} & 3.45E+03 =          & 3.33E+03 +          & \textbf{3.31E+03} \\
                                   & \textbf{Std}  & 5.60E+02            & 3.12E+01            & 1.72E+01          \\
\multirow{2}{*}{\textbf{$F_{29}$}} & \textbf{Mean} & 3.77E+03 +          & \textbf{3.60E+03 =} & 3.61E+03          \\
                                   & \textbf{Std}  & 2.48E+02            & 1.59E+02            & 1.88E+02          \\
\multirow{2}{*}{\textbf{$F_{30}$}} & \textbf{Mean} & 1.89E+06 +          & \textbf{1.07E+06 =} & 1.15E+06          \\
                                   & \textbf{Std}  & 1.01E+06            & 2.53E+05            & 3.41E+05          \\ \hline
\multicolumn{5}{c}{\textbf{Overall: 18/5/35   (+/-/=)}}                                                            \\ \hline
\end{tabular}
\end{table}

\subsection{Ablation Study}
To verify the necessity of the hypergraph update mechanism, an ablation experiment is conducted. Specifically, the proposed HPSO method is compared with a variant of HPSO that employs a fixed hypergraph topology, called HPSO-F. The experiments are also conducted on the CEC'17 benchmark suite of 50D. The results of the Wilcoxon rank-sum test for the two methods are presented in Table \ref{ablation}. 
According to Table \ref{ablation}, HPSO achieves better mean values on 9 out of the 29 functions, while underperforming HPSO-F on only 2 functions. Furthermore, HPSO exhibits superior stability to HPSO-F across the majority of test functions. To further investigate the role of the hypergraph update mechanism, we calculate the swarm diversity for both methods.

Swarm diversity is highly useful for validating the exploration and exploitation capabilities of PSO methods \cite{diversity}. Therefore, we define the diversity measure according to \cite{ARPSO} \cite{DPSO-MLS} as follows:
\begin{align}
    Div = \frac{1}{N} \sum_{i=1}^{N} \sqrt{\sum_{j=1}^{D} (x_{i,j} - \bar{x}_j)^2},
\label{eqdiv}
\end{align}
where $D$ is the dimension of the search space, $N$ is the swarm size, and $\bar{x}_j$ is the centroid of the current swarm, defined as follows:
\begin{align}
    \bar{x}_j = \frac{1}{N} \sum_{i=1}^{N} x_{i,j},
\label{eqdivx_j}
\end{align}

\begin{table}[]
\setlength{\tabcolsep}{2pt}
\renewcommand\arraystretch{1}
\footnotesize
\caption{The ablation experiment result of HPSO and HPSO-F on the CEC'17 benchmark suite (50D).}
\label{ablation}
\begin{tabular}{ccllcll}
\hline
\textbf{}                          & \textbf{}     & \textbf{HPSO}     & \textbf{HPSO-F}     &                                    & \textbf{HPSO}     & \textbf{HPSO-F}     \\ \hline
\multirow{2}{*}{\textbf{$F_1$}}    & \textbf{Mean} & \textbf{3.73E+03} & 4.69E+03 =          & \multirow{2}{*}{\textbf{$F_{17}$}} & \textbf{2.28E+03} & 2.72E+03 +          \\
                                   & \textbf{Std}  & \textbf{3.51E+03} & 4.76E+03            &                                    & \textbf{2.04E+02} & 3.25E+02            \\
\multirow{2}{*}{\textbf{$F_3$}}    & \textbf{Mean} & \textbf{3.24E+04} & 3.99E+04 +          & \multirow{2}{*}{\textbf{$F_{18}$}} & \textbf{1.75E+06} & 2.10E+06 =          \\
                                   & \textbf{Std}  & \textbf{5.29E+03} & 8.49E+03            &                                    & \textbf{1.15E+06} & 1.14E+06            \\
\multirow{2}{*}{\textbf{$F_4$}}    & \textbf{Mean} & \textbf{5.48E+02} & 5.58E+02 =          & \multirow{2}{*}{\textbf{$F_{19}$}} & \textbf{1.75E+04} & 1.78E+04 =          \\
                                   & \textbf{Std}  & \textbf{4.48E+01} & 4.65E+01            &                                    & \textbf{1.19E+04} & 1.09E+04            \\
\multirow{2}{*}{\textbf{$F_5$}}    & \textbf{Mean} & \textbf{5.26E+02} & 5.40E+02 +          & \multirow{2}{*}{\textbf{$F_{20}$}} & \textbf{2.39E+03} & 2.70E+03 +          \\
                                   & \textbf{Std}  & \textbf{6.42E+00} & 1.23E+01            &                                    & \textbf{2.75E+02} & 2.48E+02            \\
\multirow{2}{*}{\textbf{$F_6$}}    & \textbf{Mean} & 6.03E+02          & \textbf{6.02E+02 -} & \multirow{2}{*}{\textbf{$F_{21}$}} & \textbf{2.34E+03} & 2.35E+03 +          \\
                                   & \textbf{Std}  & 1.02E+00          & \textbf{1.18E+00}   &                                    & \textbf{9.34E+00} & 1.21E+01            \\
\multirow{2}{*}{\textbf{$F_7$}}    & \textbf{Mean} & \textbf{7.82E+02} & 8.21E+02 +          & \multirow{2}{*}{\textbf{$F_{22}$}} & \textbf{7.01E+03} & 1.08E+04 +          \\
                                   & \textbf{Std}  & \textbf{7.92E+00} & 4.68E+01            &                                    & \textbf{1.02E+03} & 2.81E+03            \\
\multirow{2}{*}{\textbf{$F_8$}}    & \textbf{Mean} & \textbf{8.27E+02} & 8.36E+02 +          & \multirow{2}{*}{\textbf{$F_{23}$}} & \textbf{2.87E+03} & \textbf{2.87E+03 =} \\
                                   & \textbf{Std}  & \textbf{7.32E+00} & 8.38E+00            &                                    & \textbf{5.61E+01} & \textbf{6.50E+01}   \\
\multirow{2}{*}{\textbf{$F_9$}}    & \textbf{Mean} & 9.42E+02          & \textbf{9.15E+02 -} & \multirow{2}{*}{\textbf{$F_{24}$}} & \textbf{3.06E+03} & 3.09E+03 =          \\
                                   & \textbf{Std}  & 5.18E+01          & \textbf{1.44E+01}   &                                    & \textbf{4.83E+01} & 5.48E+01            \\
\multirow{2}{*}{\textbf{$F_{10}$}} & \textbf{Mean} & \textbf{5.81E+03} & 9.57E+03 +          & \multirow{2}{*}{\textbf{$F_{25}$}} & \textbf{3.04E+03} & 3.05E+03 =          \\
                                   & \textbf{Std}  & \textbf{9.21E+02} & 2.54E+03            &                                    & \textbf{2.81E+01} & 2.06E+01            \\
\multirow{2}{*}{\textbf{$F_{11}$}} & \textbf{Mean} & \textbf{1.21E+03} & 1.22E+03 =          & \multirow{2}{*}{\textbf{$F_{26}$}} & \textbf{4.77E+03} & 4.90E+03 =          \\
                                   & \textbf{Std}  & \textbf{3.75E+01} & 3.84E+01            &                                    & \textbf{4.23E+02} & 5.92E+02            \\
\multirow{2}{*}{\textbf{$F_{12}$}} & \textbf{Mean} & \textbf{2.44E+06} & 2.71E+06 =          & \multirow{2}{*}{\textbf{$F_{27}$}} & 3.41E+03          & \textbf{3.36E+03 =} \\
                                   & \textbf{Std}  & \textbf{1.71E+06} & 1.96E+06            &                                    & 1.09E+02          & \textbf{7.18E+01}   \\
\multirow{2}{*}{\textbf{$F_{13}$}} & \textbf{Mean} & 1.11E+04          & \textbf{1.08E+04 =} & \multirow{2}{*}{\textbf{$F_{28}$}} & \textbf{3.31E+03} & \textbf{3.31E+03 =} \\
                                   & \textbf{Std}  & 1.01E+04          & \textbf{8.40E+03}   &                                    & \textbf{1.72E+01} & \textbf{2.06E+01}   \\
\multirow{2}{*}{\textbf{$F_{14}$}} & \textbf{Mean} & \textbf{1.08E+05} & 1.18E+05 =          & \multirow{2}{*}{\textbf{$F_{29}$}} & 3.61E+03          & \textbf{3.60E+03 =} \\
                                   & \textbf{Std}  & \textbf{5.04E+04} & 6.40E+04            &                                    & 1.88E+02          & \textbf{1.75E+02}   \\
\multirow{2}{*}{\textbf{$F_{15}$}} & \textbf{Mean} & 8.66E+03          & \textbf{6.51E+03 =} & \multirow{2}{*}{\textbf{$F_{30}$}} & \textbf{1.15E+06} & 1.25E+06 =          \\
                                   & \textbf{Std}  & 5.11E+03          & \textbf{4.03E+03}   &                                    & \textbf{3.41E+05} & 4.34E+05            \\
\multirow{2}{*}{\textbf{$F_{16}$}} & \textbf{Mean} & \textbf{2.46E+03} & 2.65E+03 =          & \multicolumn{3}{c}{\multirow{2}{*}{\textbf{Overall: 9/2/18 (+/-/=)}}}        \\
                                   & \textbf{Std}  & \textbf{3.82E+02} & 4.48E+02            & \multicolumn{3}{c}{}                                                         \\ \hline
\end{tabular}
\end{table}

The swarm diversity curves of HPSO and HPSO-F for 4 functions (i.e., $F_8$, $F_{16}$, $F_{22}$, and $F_{29}$) on 50D are illustrated in Fig. \ref{fig_abl_div}, while the diversity curves for the remaining functions are provided in \textbf{Appendix C}. In Fig. \ref{fig_abl_div}, it is observed that HPSO maintains higher swarm diversity than HPSO-F during the early stages of evolution. This indicates that the hypergraph update mechanism effectively enhances swarm diversity, preventing the diversity loss that occurs when particles are restricted to learning from fixed hyperedges. In addition, maintaining high swarm diversity during the early evolutionary phase facilitates effective exploration of the search space. In the late stages of evolution, as the movement distance of particles decreases, the frequency of hypergraph updates naturally diminishes according to Eq. \eqref{eqd^a}. Therefore, the swarm diversity becomes lower than that of HPSO-F, which is more conducive to the exploitation of potentially optimal regions. 

\begin{figure}
     \centering
     \begin{subfigure}[t]{0.48\columnwidth}
         \centering
         \includegraphics[width=1.1\textwidth, keepaspectratio, trim=0 0 50 20, clip]{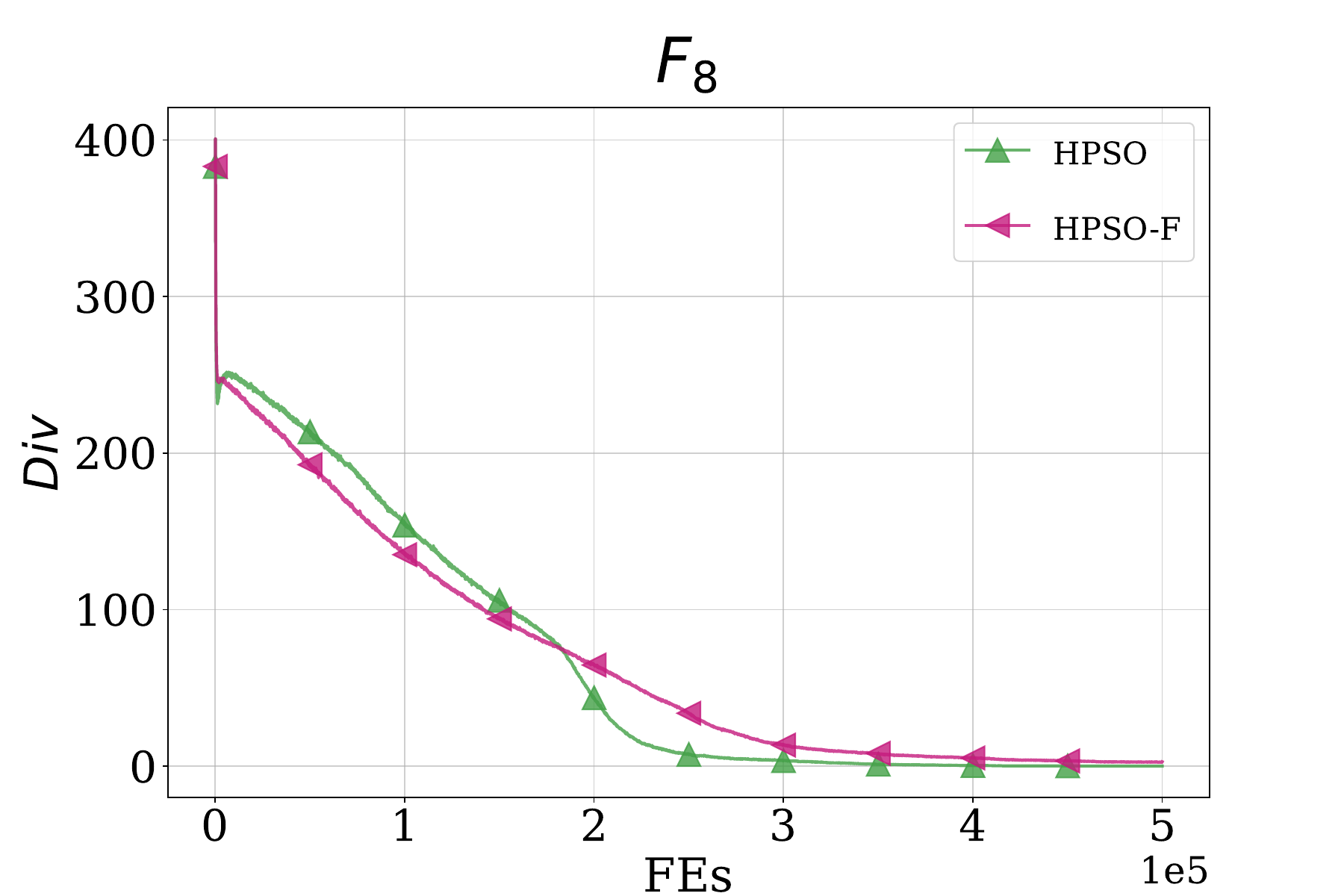}
     \end{subfigure}\hfill
     \begin{subfigure}[t]{0.48\columnwidth}
         \centering
         \includegraphics[width=1.1\textwidth, keepaspectratio, trim=0 0 50 20, clip]{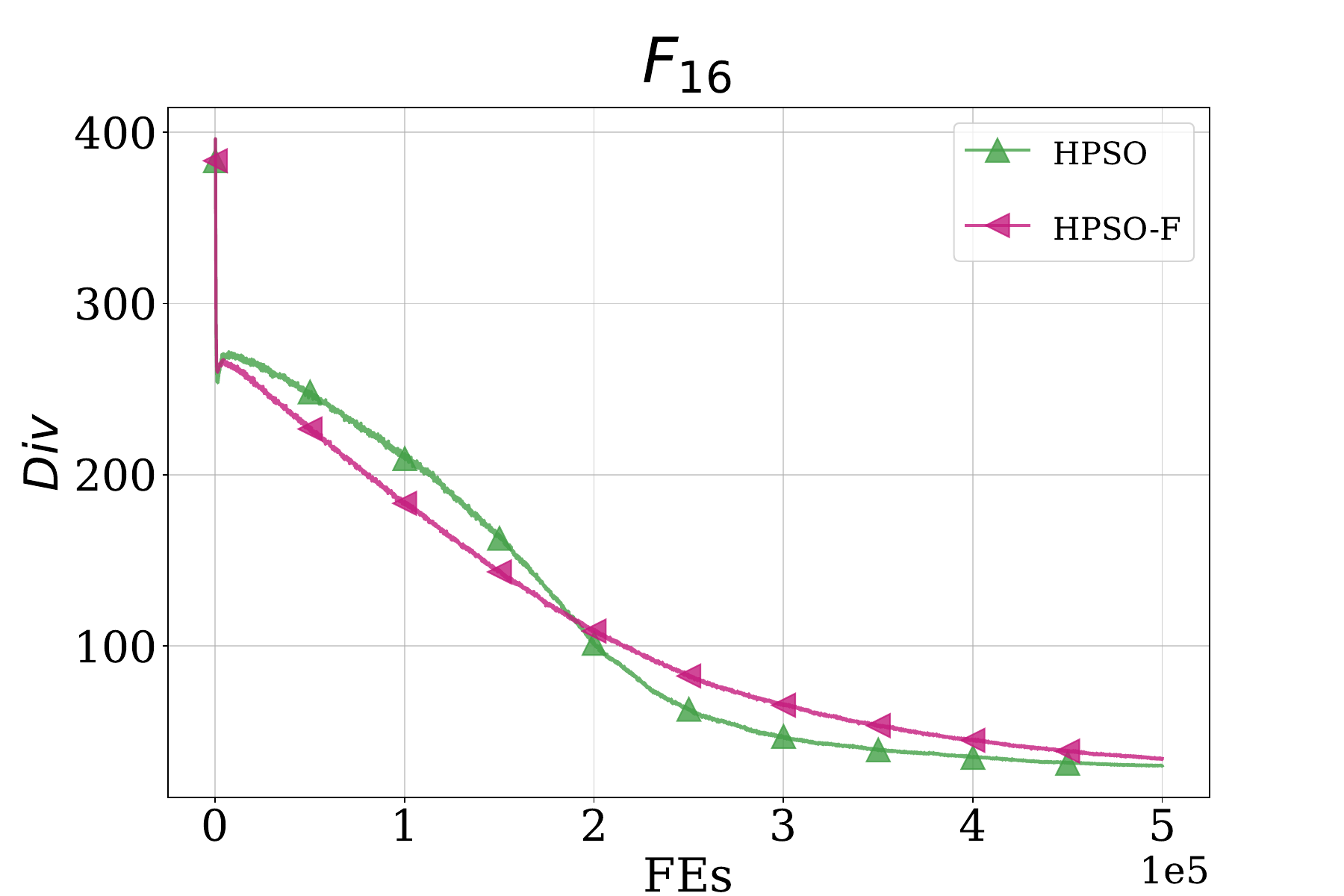}
     \end{subfigure}
      \vspace{2pt}
        
      \begin{subfigure}[t]{0.48\columnwidth}
         \centering
         \includegraphics[width=1.1\textwidth, keepaspectratio, trim=0 0 50 20, clip]{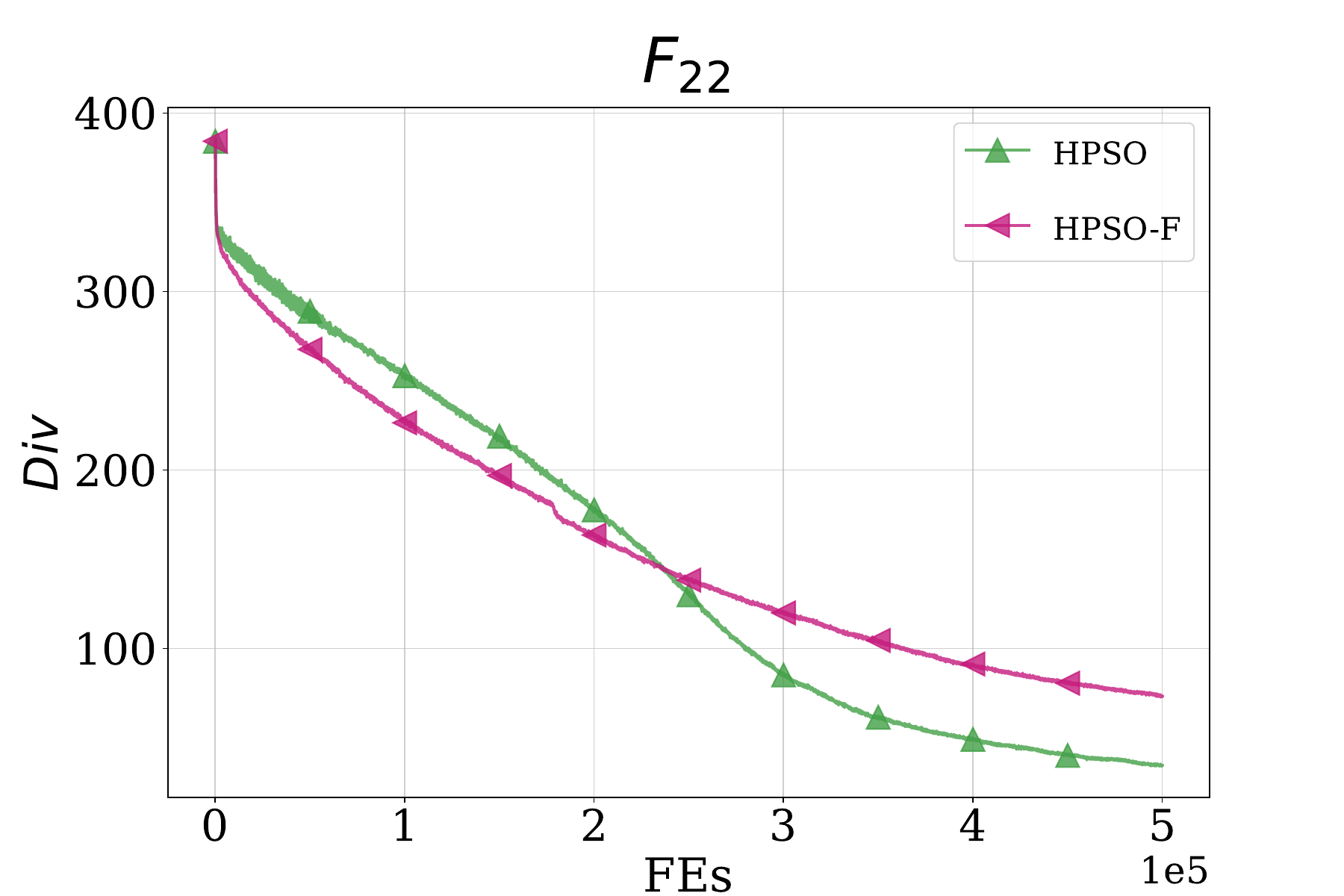}
     \end{subfigure}\hfill
     \begin{subfigure}[t]{0.48\columnwidth}
         \centering
         \includegraphics[width=1.1\textwidth, keepaspectratio, trim=0 0 50 20, clip]{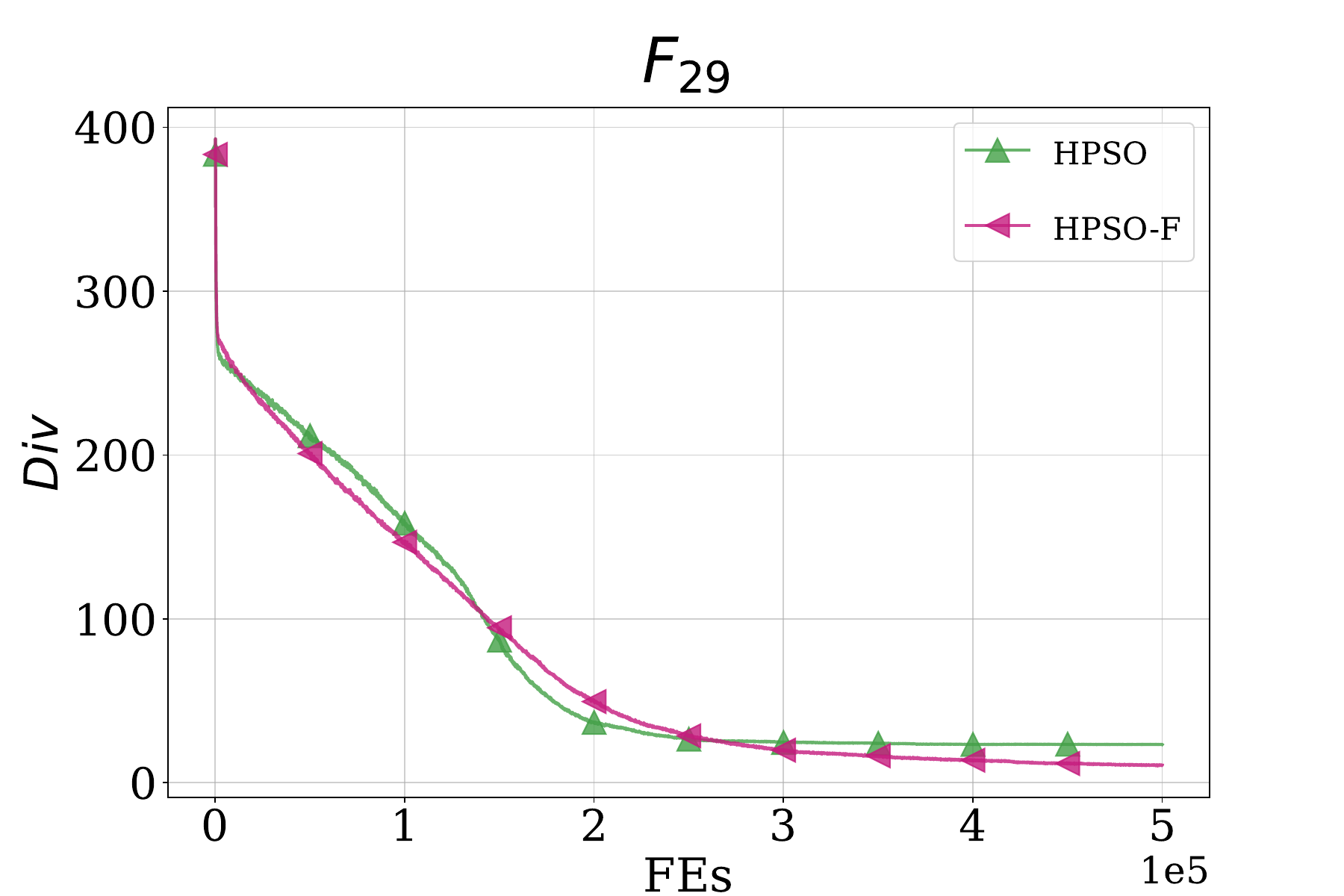}
     \end{subfigure}
     
     \caption{Swarm diversity curves of HPSO and HPSO-F on $F_8$, $F_{16}$, $F_{22}$, and $F_{29}$.}
\label{fig_abl_div}
\end{figure}

\section{Conclusions}

This study aims to investigate how higher-order social relationships among particles can be captured for navigating complex search landscapes. The goal has been achieved by designing a new particle swarm optimization method, HPSO, which is built upon hypergraph topology. In HPSO, a hypergraph is constructed to capture high-order relationships among particles, with information disseminated via hyperedges. Unlike standard graph topologies, this high-order connectivity broadens information pathways and accelerates knowledge dissemination, leading to more rapid convergence. Furthermore, a dynamic hypergraph update strategy is introduced to adapt to the evolving particle distribution while maintaining swarm diversity to prevent local optima. A matrix-based learning strategy is further designed, enabling each particle to learn from multiple exemplars within the same hyperedge. 

To verify the effectiveness of HPSO, extensive experiments were conducted on the CEC'17 benchmark suite across 30D, 50D, and 100D. Results demonstrate that HPSO significantly outperforms the compared PSO variants, particularly on complex optimization problems. Furthermore, HPSO exhibits a strong ability to escape local optima in the late evolutionary stages, mainly because of the hypergraph topology that guides particles away from suboptimal regions. However, the periodic reconstruction of the hypergraph topology during evolution introduces additional computational overhead. In future work, we will investigate strategies to reduce the training time.

\section*{Acknowledgments}
This work was supported in part by the National Key Research and Development Program of China under grant 2024YFA1012700, the National Natural Science Foundation of China under grants 62206041, 12371516 and U21A20491, and the NSFC-Liaoning Province United Foundation under grant U1908214, the 111 Project under grant D23006, the Liaoning Revitalization Talents Program under grant XLYC2008017, and China University Industry-University-Research Innovation Fund under grants 2022IT174, Natural Science Foundation of Liaoning Province under grant 2023-BSBA-030, and an Open Fund of National Engineering Laboratory for Big Data System Computing Technology under grant SZU-BDSC-OF2024-09.

\printcredits

\bibliographystyle{model1-num-names}

\bibliography{cas-refs}


\clearpage
\appendix

\renewcommand{\thetable}{S-\arabic{table}}
\renewcommand{\thefigure}{S-\arabic{figure}}
\setcounter{table}{0} 
\setcounter{figure}{0}
\captionsetup[subfigure]{    
    labelsep=none,           
    font={bf},         
    justification=centering, 
    singlelinecheck=false,    
    labelformat=empty
}
\captionsetup[figure]{labelsep=period}
\renewcommand{\figurename}{Fig.}
\pagestyle{plain}

\onecolumn

\section{Results of the CEC'17 benchmark suite on 30D and 100D}
\label{appendix_A}
This section presents the comparative results between HPSO and the other seven PSOs. Tables~\ref{30D-1},~\ref{30D-2} and \ref{30D-3} show the results for 30D, while Tables~\ref{100D-1}, ~\ref{100D-2}, \ref{100D-3} display the results for 100D.

\section{Mean convergence curves on the remaining 50D problems}
\label{appendix_B}
\begin{figure*}[htbp]
     \centering

     \begin{subfigure}[t]{0.24\textwidth}
         \centering
         \includegraphics[width=\linewidth, trim=0 15 0 13, clip]{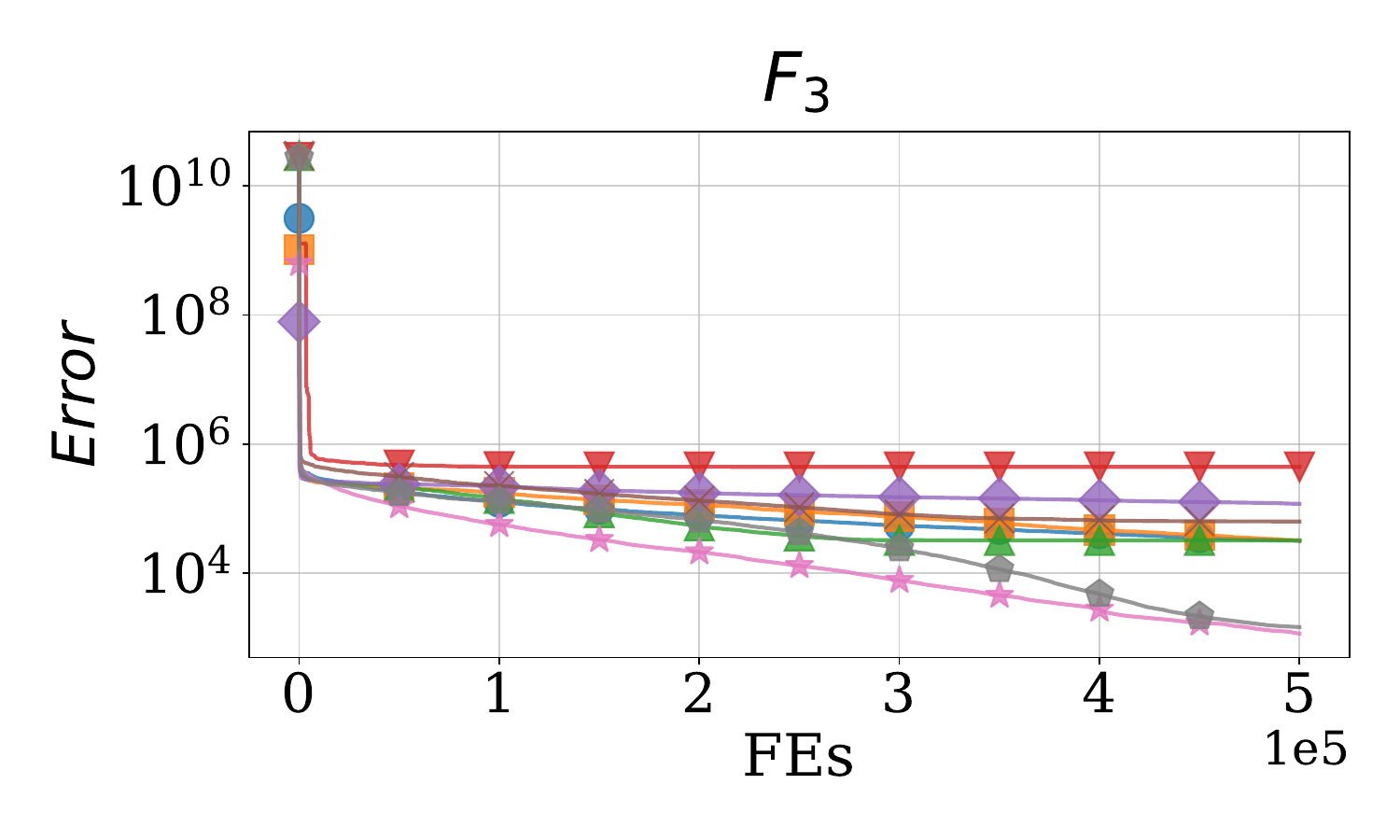}
     \end{subfigure}
     \hfill 
     \begin{subfigure}[t]{0.24\textwidth}
         \centering
         \includegraphics[width=\linewidth, trim=0 15 0 13, clip]{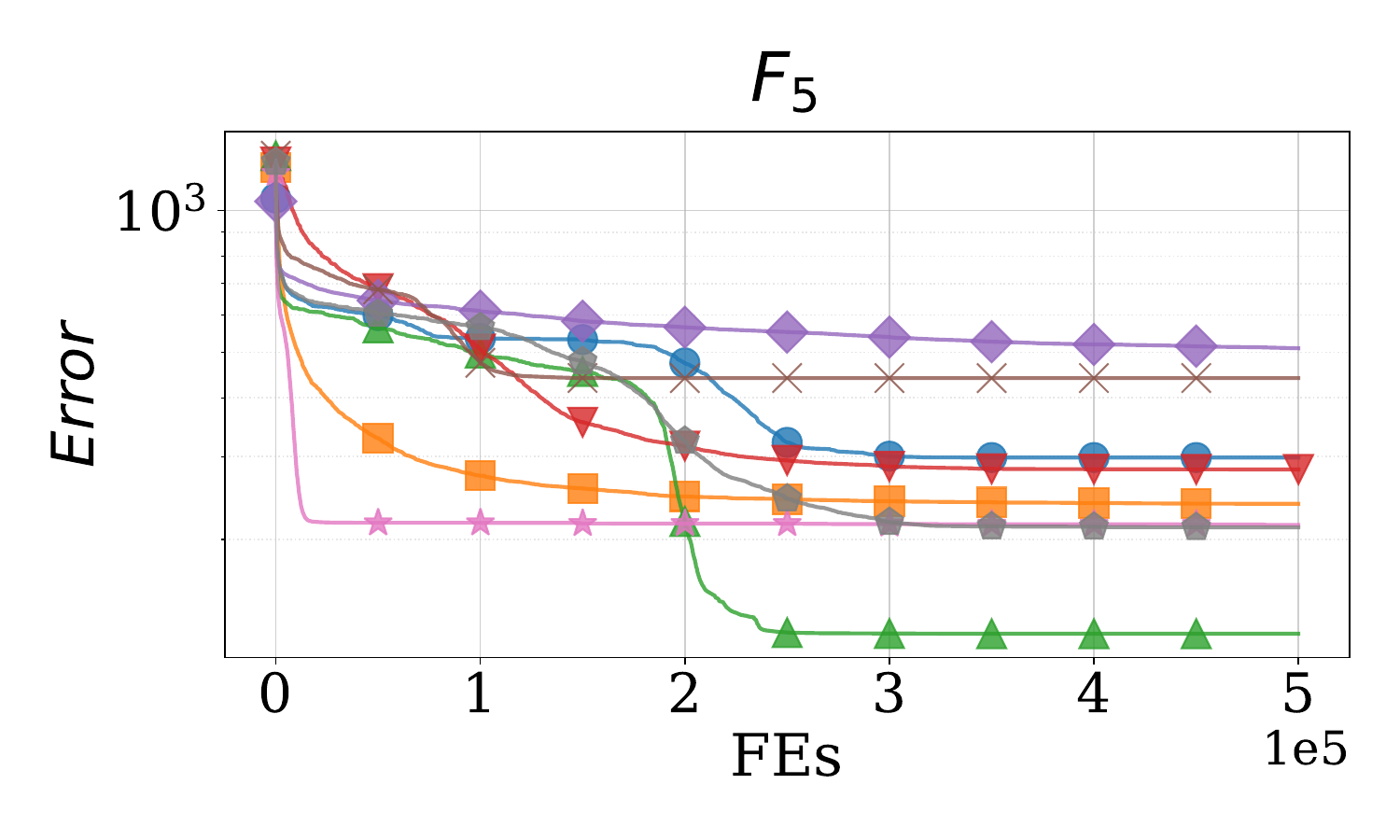}
     \end{subfigure}
     \hfill
     \begin{subfigure}[t]{0.24\textwidth}
         \centering
         \includegraphics[width=\linewidth, trim=0 15 0 13, clip]{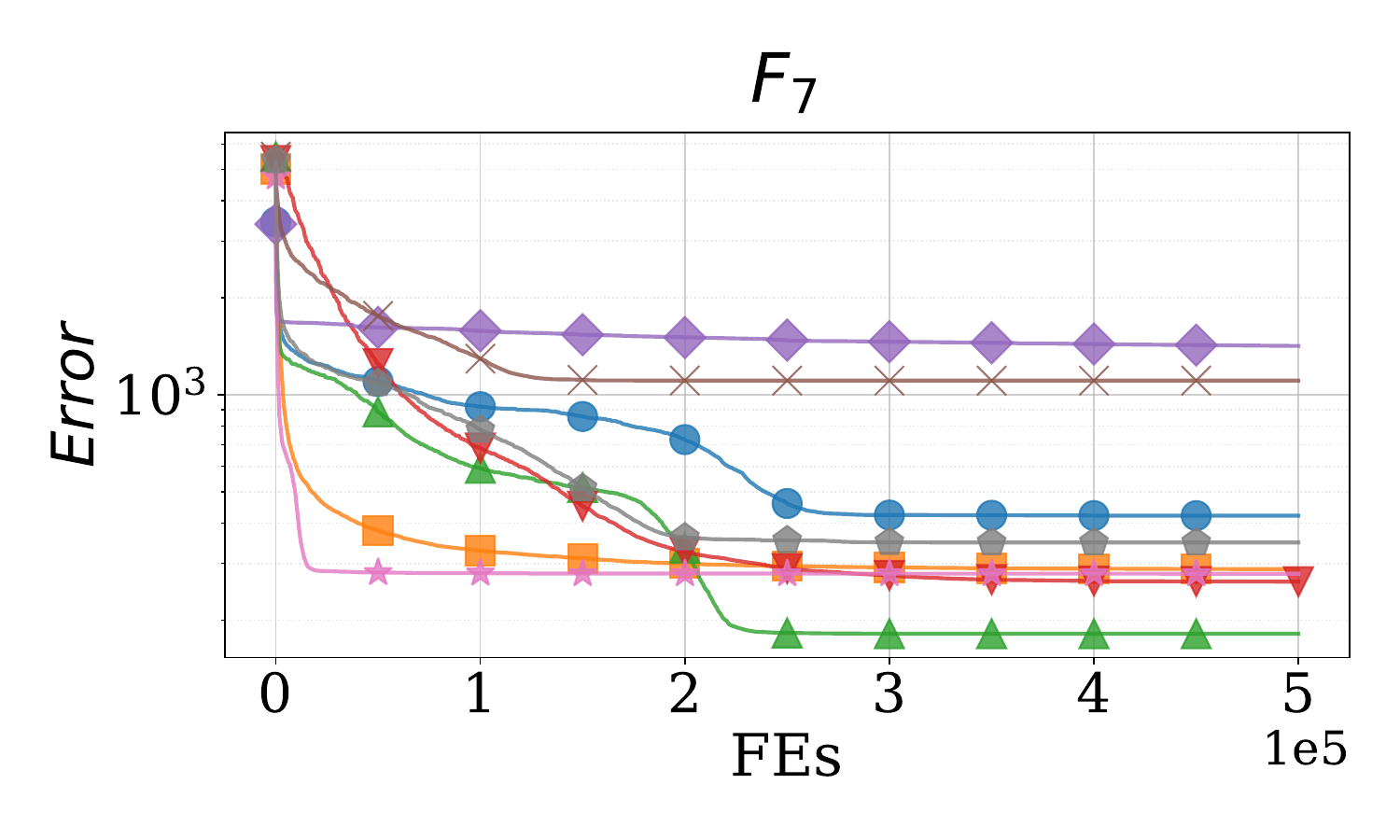}
     \end{subfigure}
     \hfill
     \begin{subfigure}[t]{0.24\textwidth}
         \centering
         \includegraphics[width=\linewidth, trim=0 15 0 13, clip]{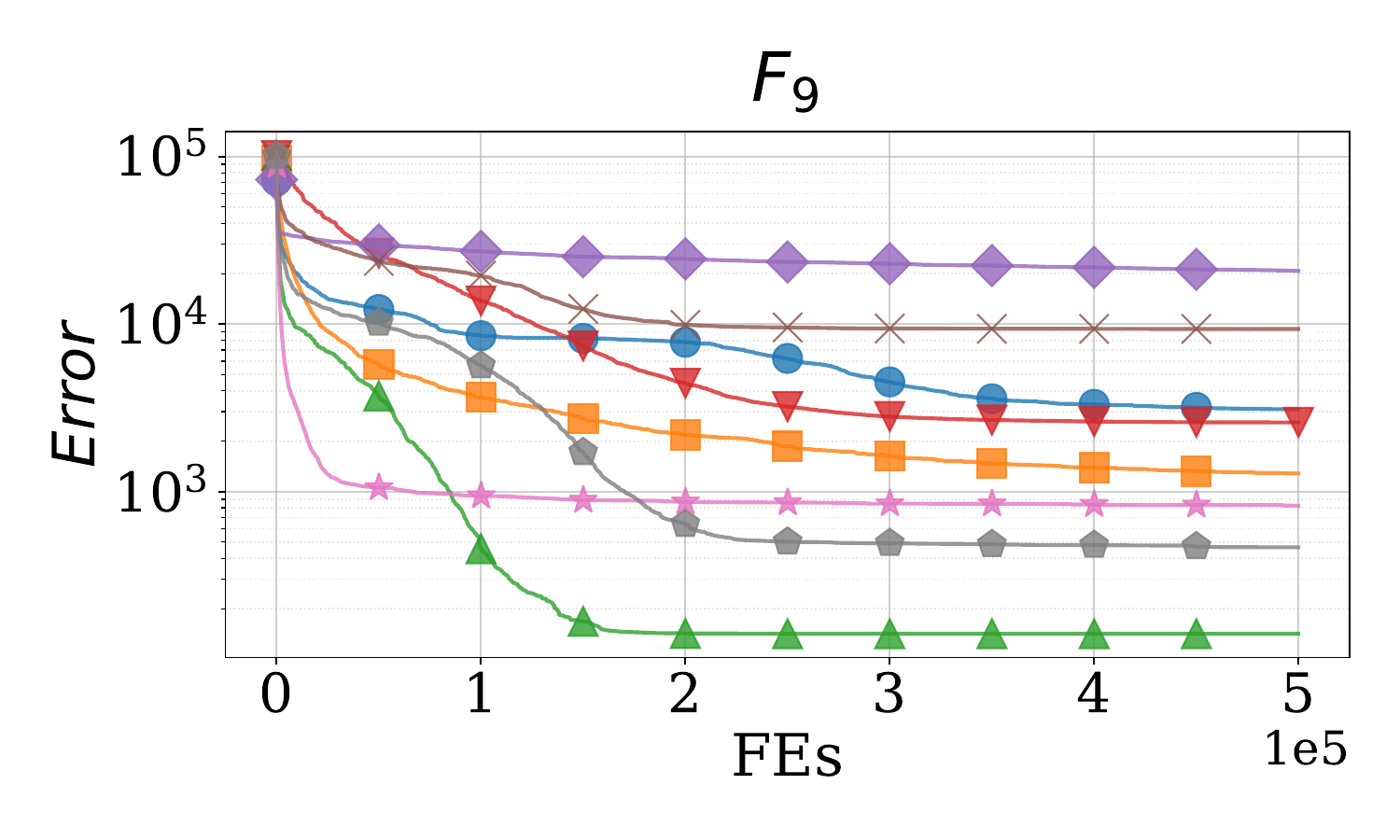}
     \end{subfigure}

      \vspace{2pt}
      \begin{subfigure}[t]{0.24\textwidth}
         \centering
         \includegraphics[width=\linewidth, trim=0 15 0 13, clip]{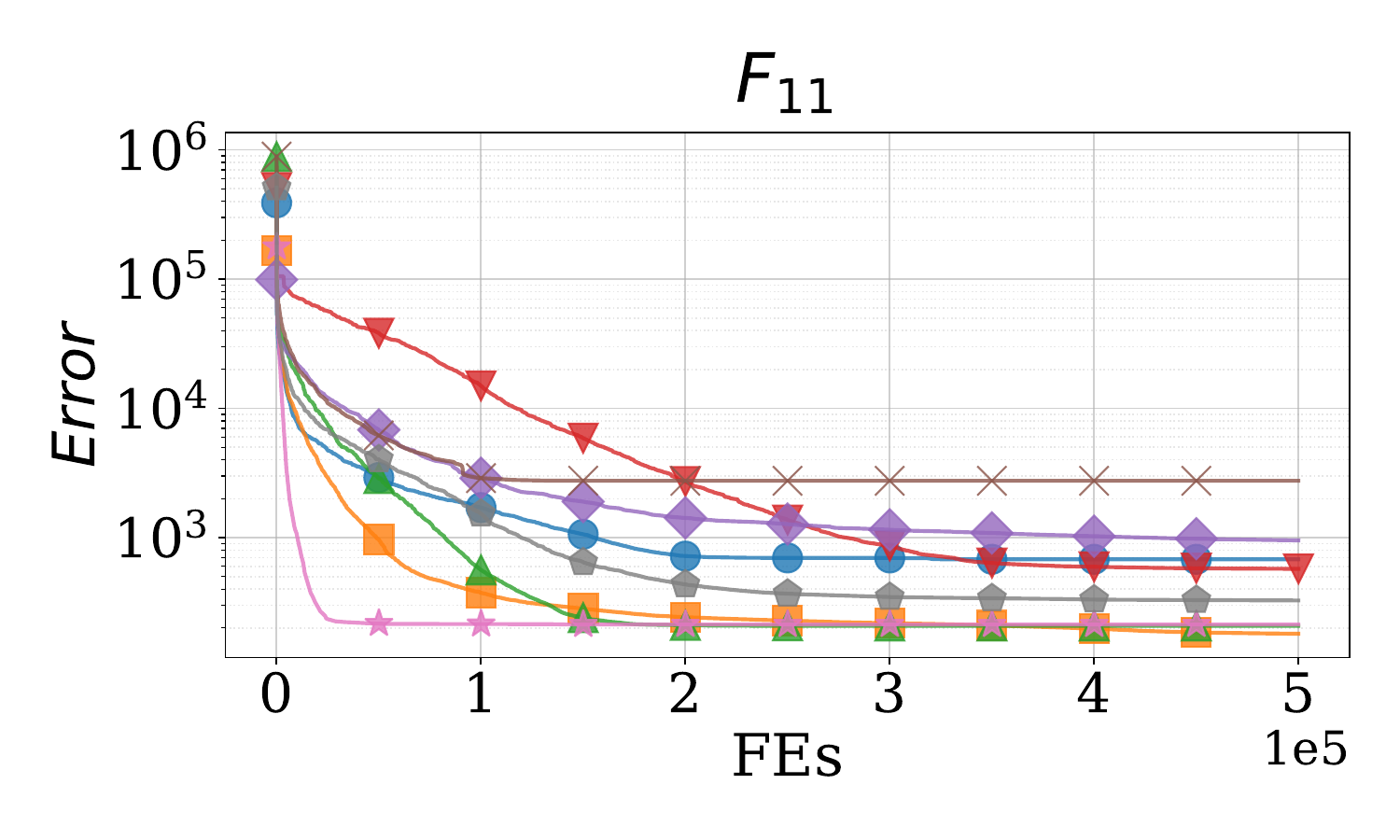}
     \end{subfigure}
     \hfill 
     \begin{subfigure}[t]{0.24\textwidth}
         \centering
         \includegraphics[width=\linewidth, trim=0 15 0 13, clip]{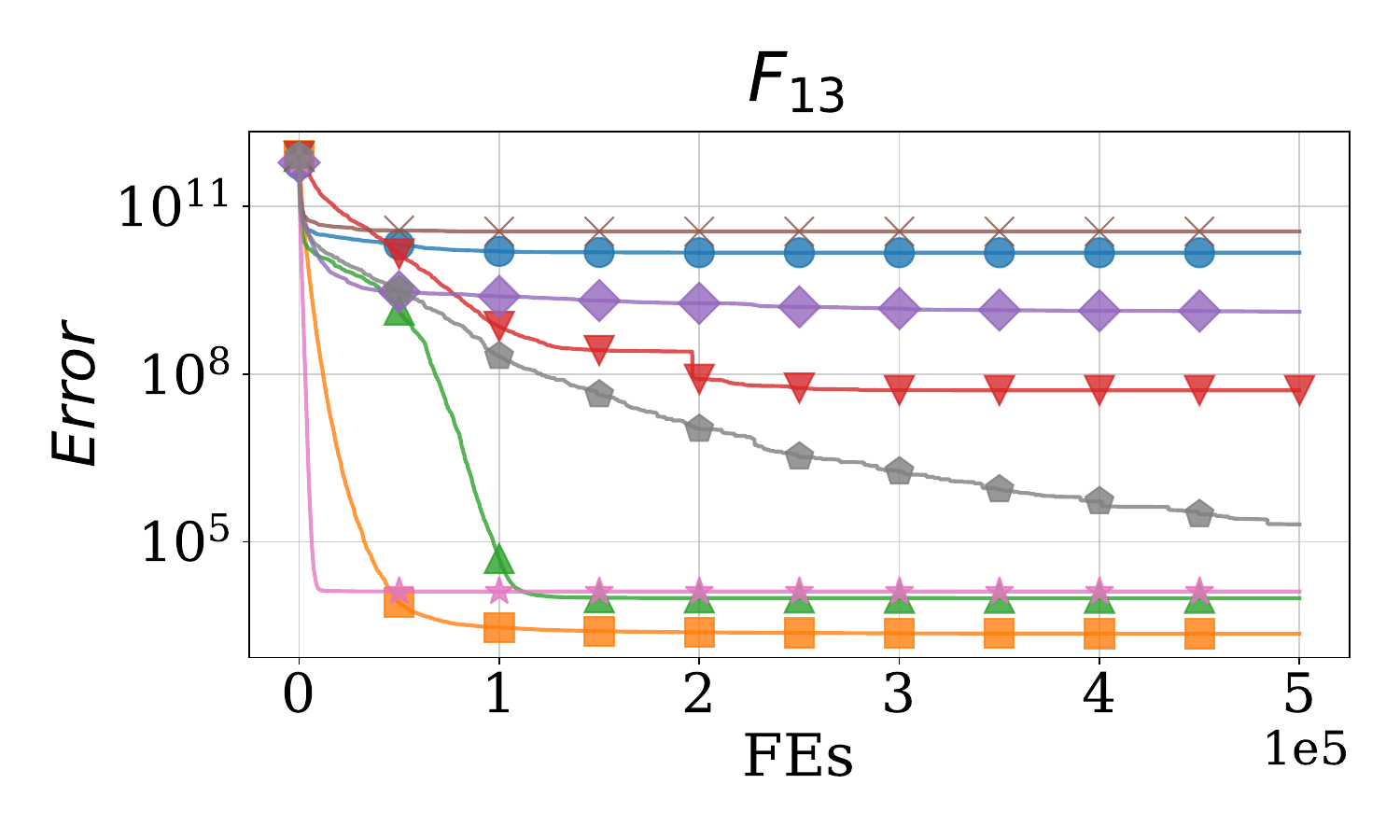}
     \end{subfigure}
     \hfill
     \begin{subfigure}[t]{0.24\textwidth}
         \centering
         \includegraphics[width=\linewidth, trim=0 15 0 13, clip]{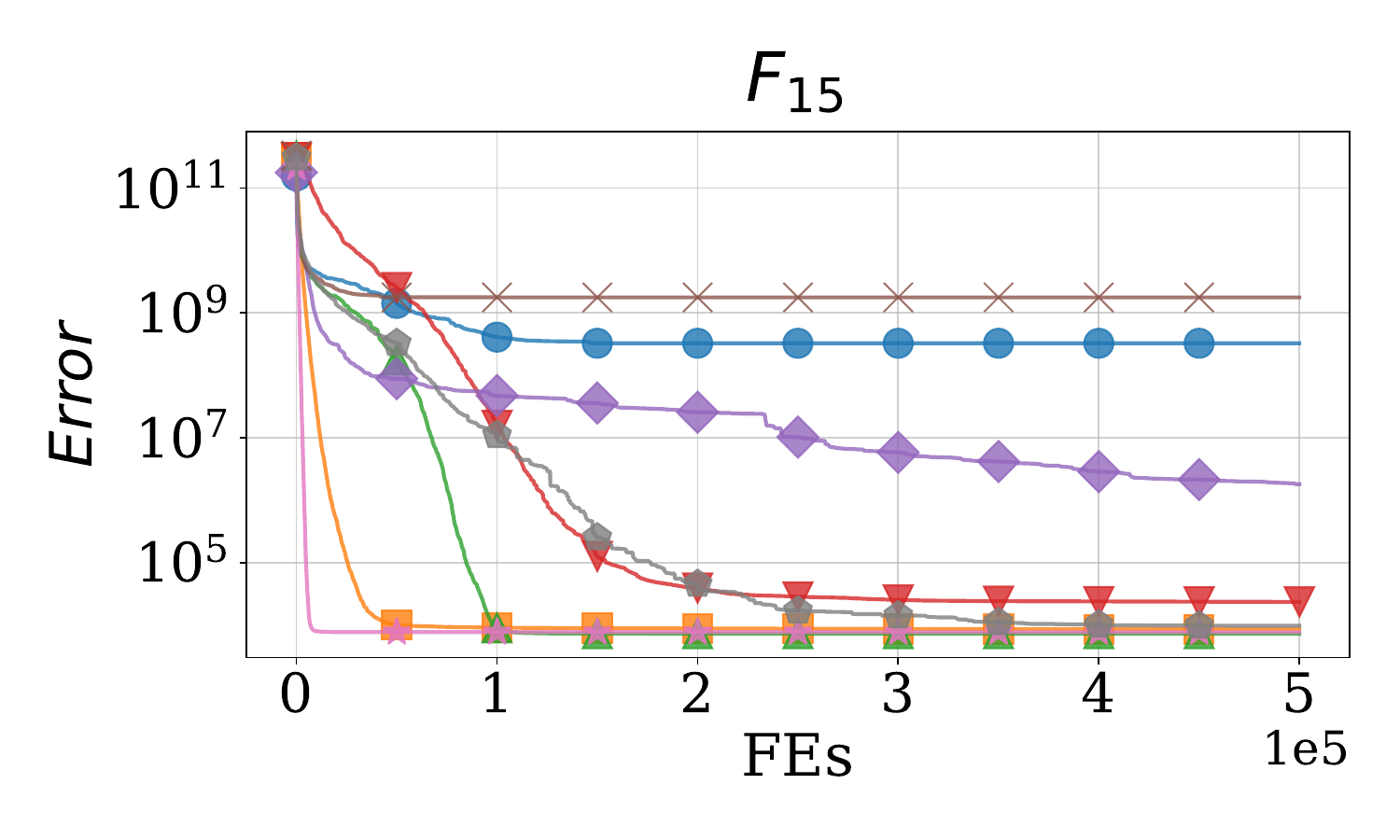}
     \end{subfigure}
     \hfill
     \begin{subfigure}[t]{0.24\textwidth}
         \centering
         \includegraphics[width=\linewidth, trim=0 15 0 13, clip]{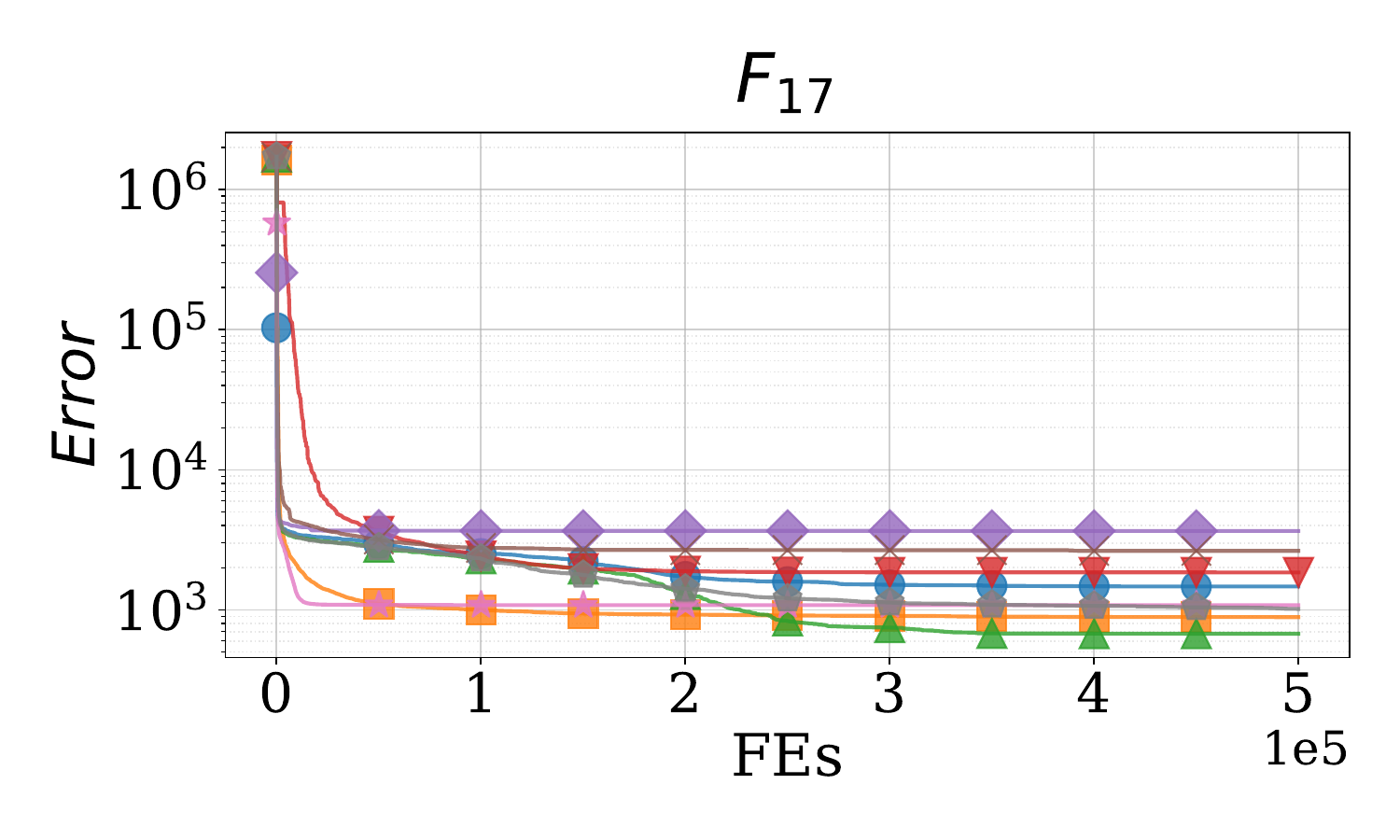}
     \end{subfigure}

      \vspace{2pt}
        
      \begin{subfigure}[t]{0.24\textwidth}
         \centering
         \includegraphics[width=\linewidth, trim=0 15 0 13, clip]{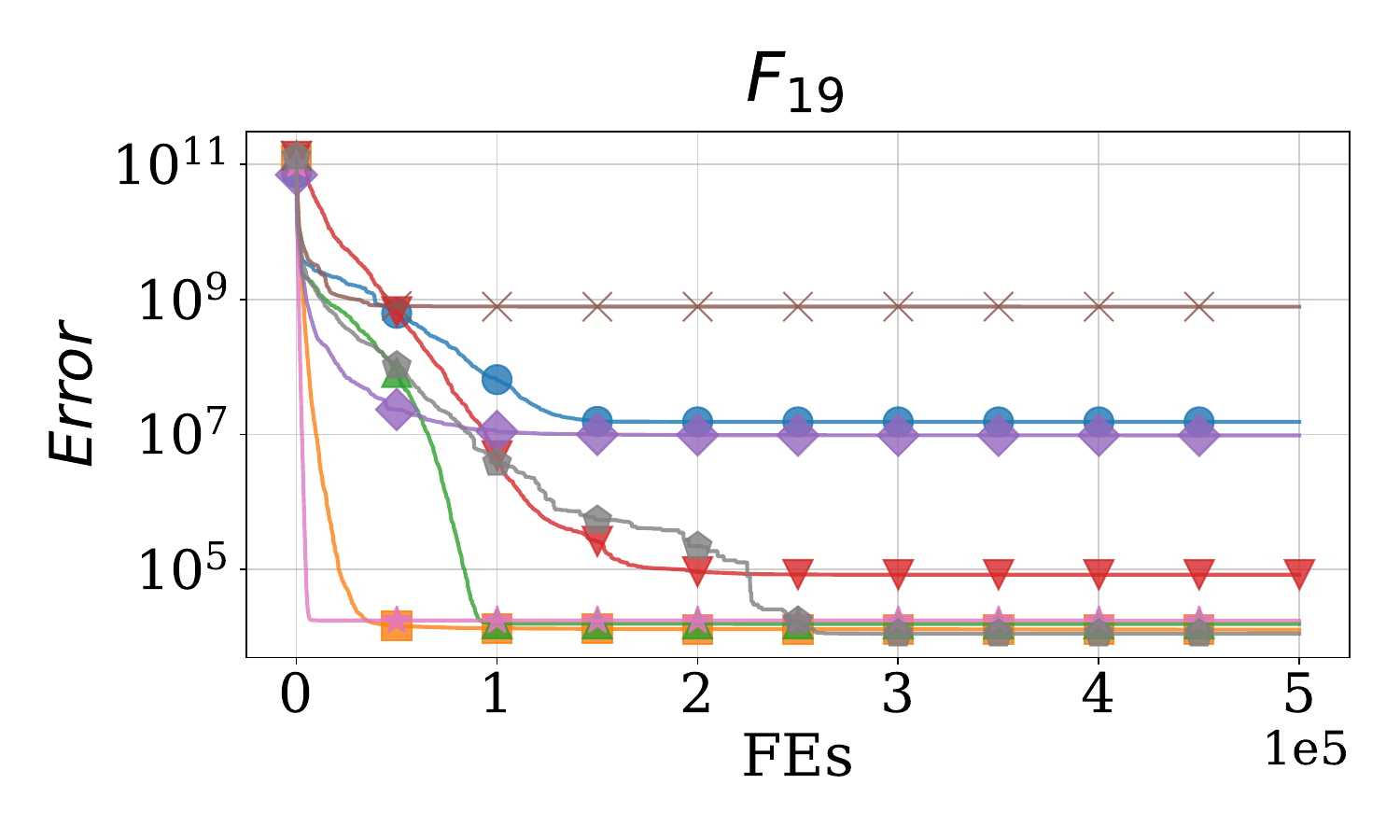}
     \end{subfigure}\hfill 
     \begin{subfigure}[t]{0.24\textwidth}
         \centering
         \includegraphics[width=\linewidth, trim=0 15 0 13, clip]{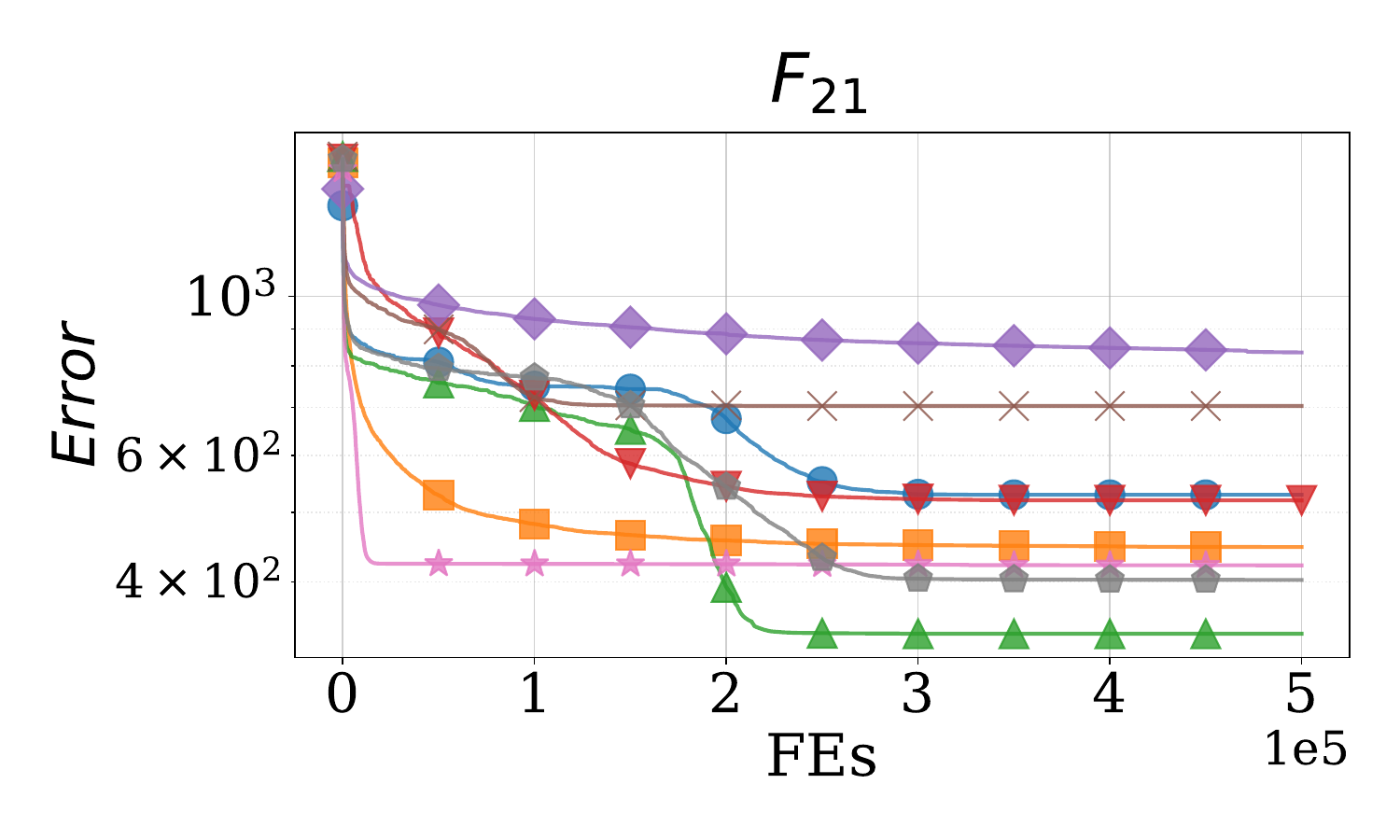}
     \end{subfigure}\hfill
     \begin{subfigure}[t]{0.24\textwidth}
         \centering
         \includegraphics[width=\linewidth, trim=0 15 0 13, clip]{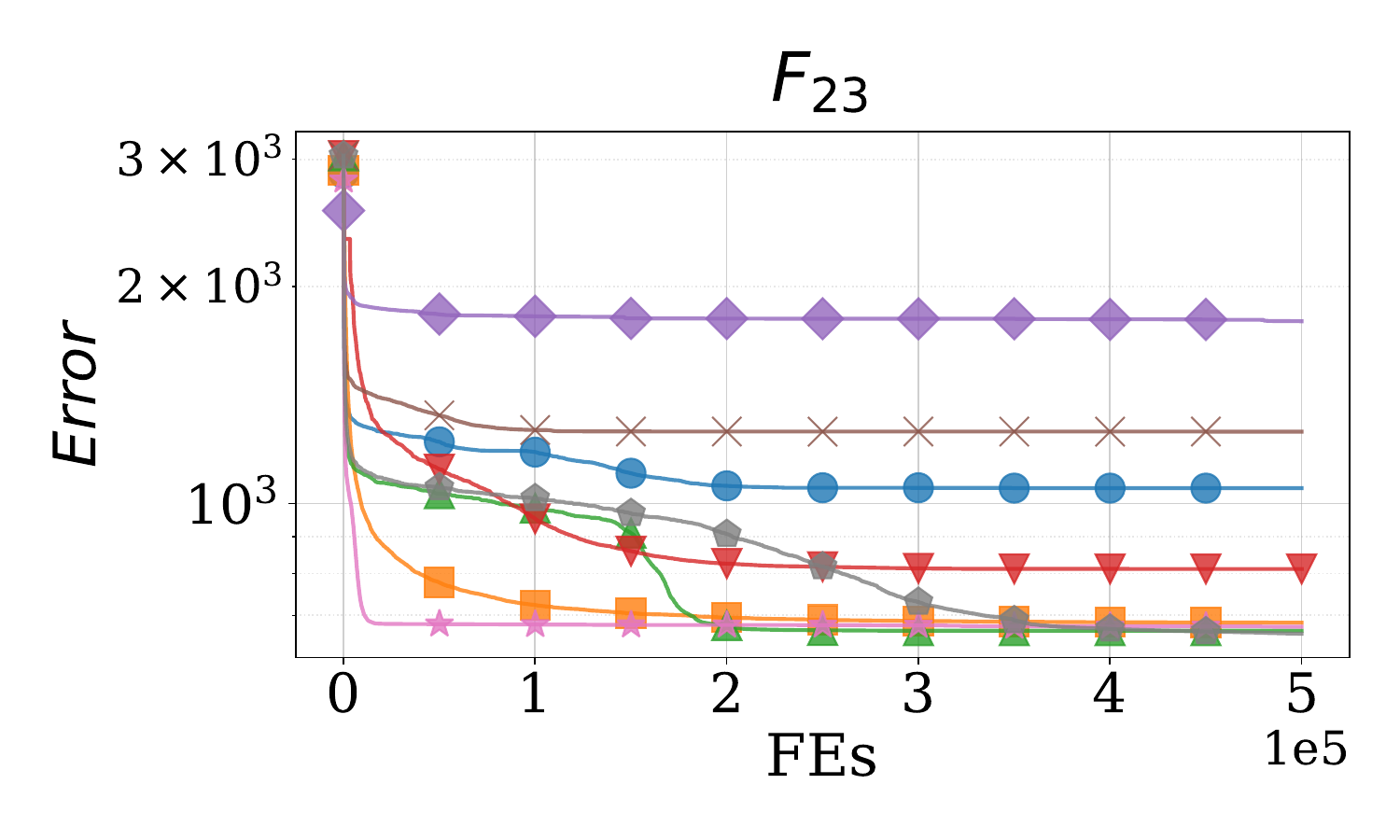}
     \end{subfigure}\hfill
     \begin{subfigure}[t]{0.24\textwidth}
         \centering
         \includegraphics[width=\linewidth, trim=0 15 0 13, clip]{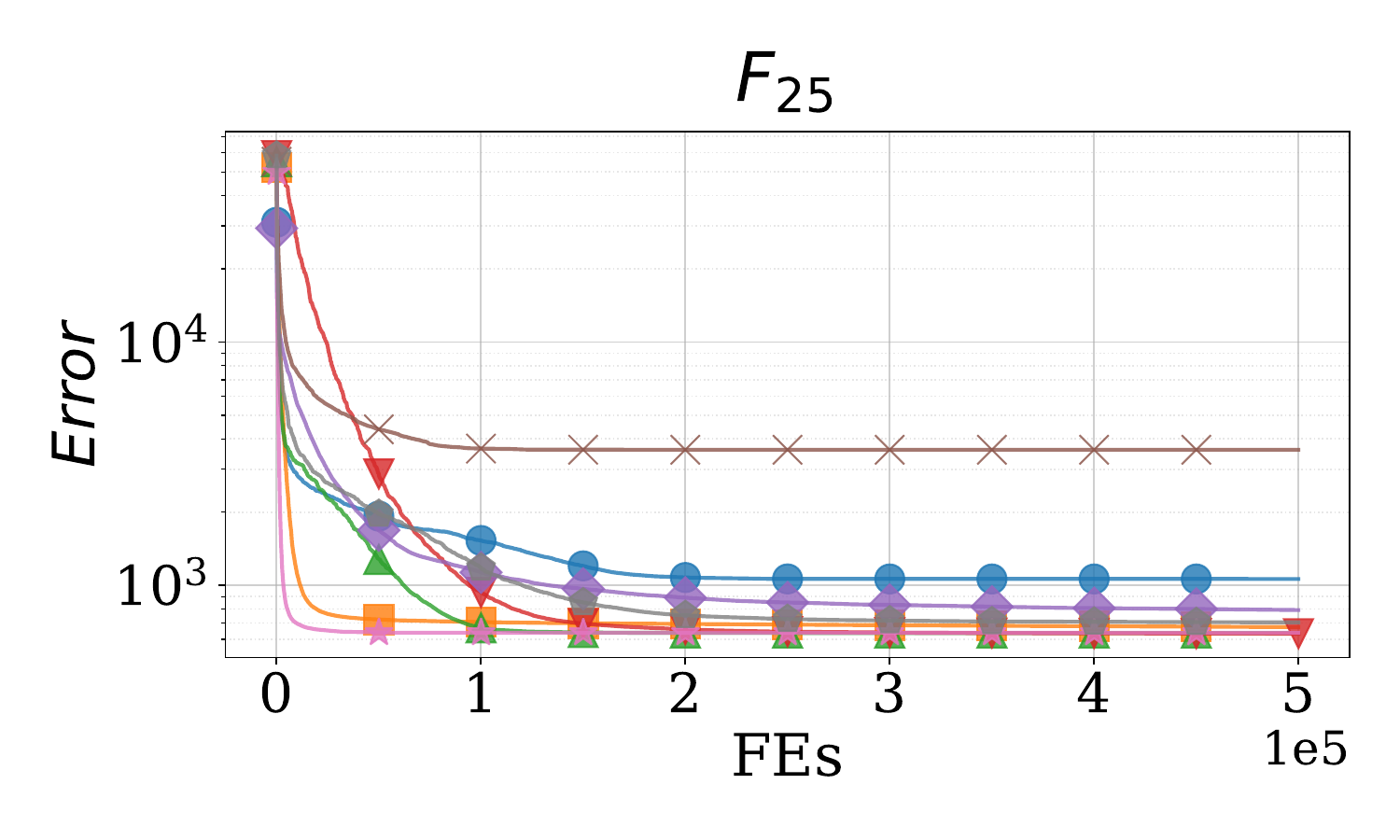}
     \end{subfigure}

      \vspace{2pt}
      \begin{subfigure}[t]{0.24\textwidth}
         \centering
         \includegraphics[width=\linewidth, trim=0 15 0 13, clip]{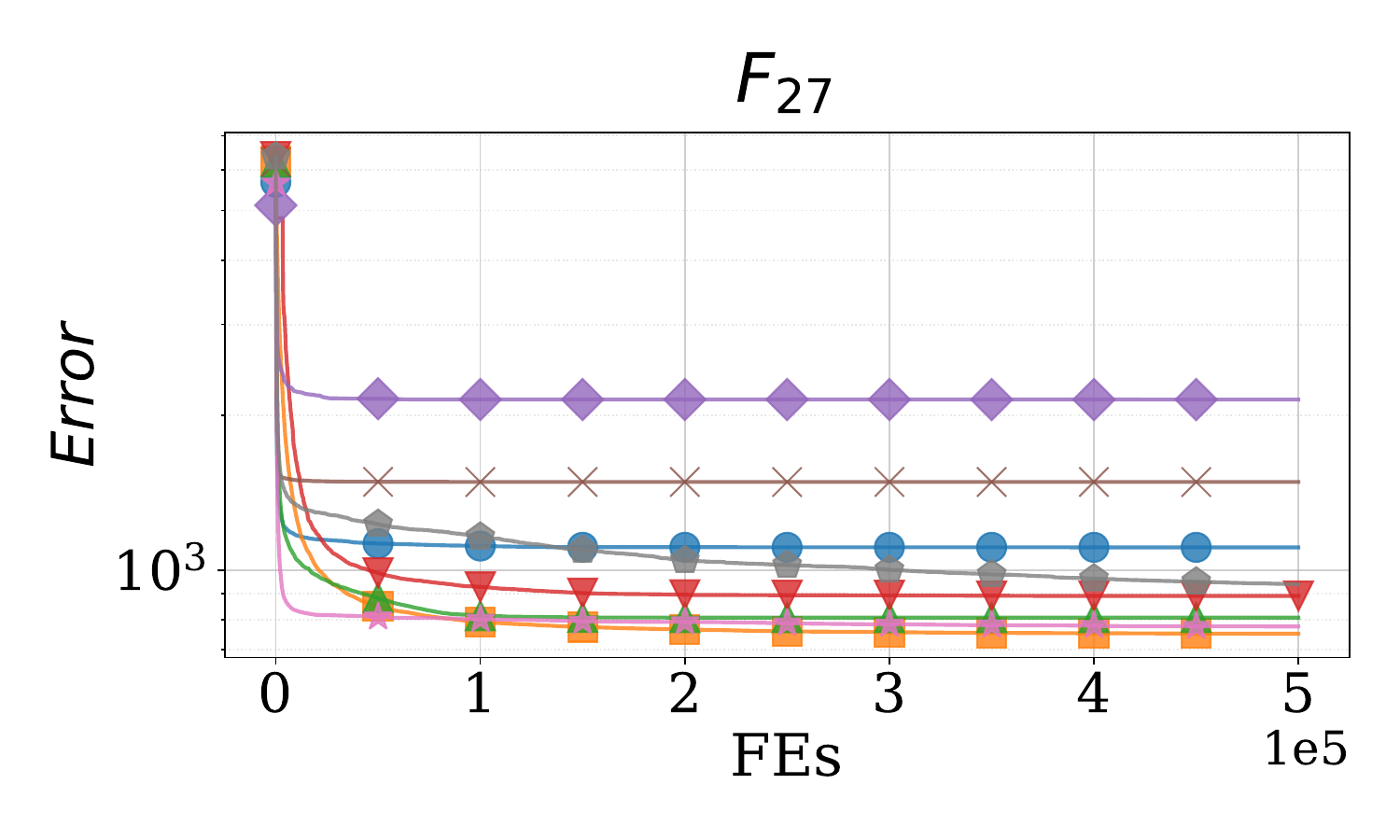}
     \end{subfigure}
     \begin{subfigure}[t]{0.24\textwidth}
         \centering
         \includegraphics[width=\linewidth, trim=0 15 0 13, clip]{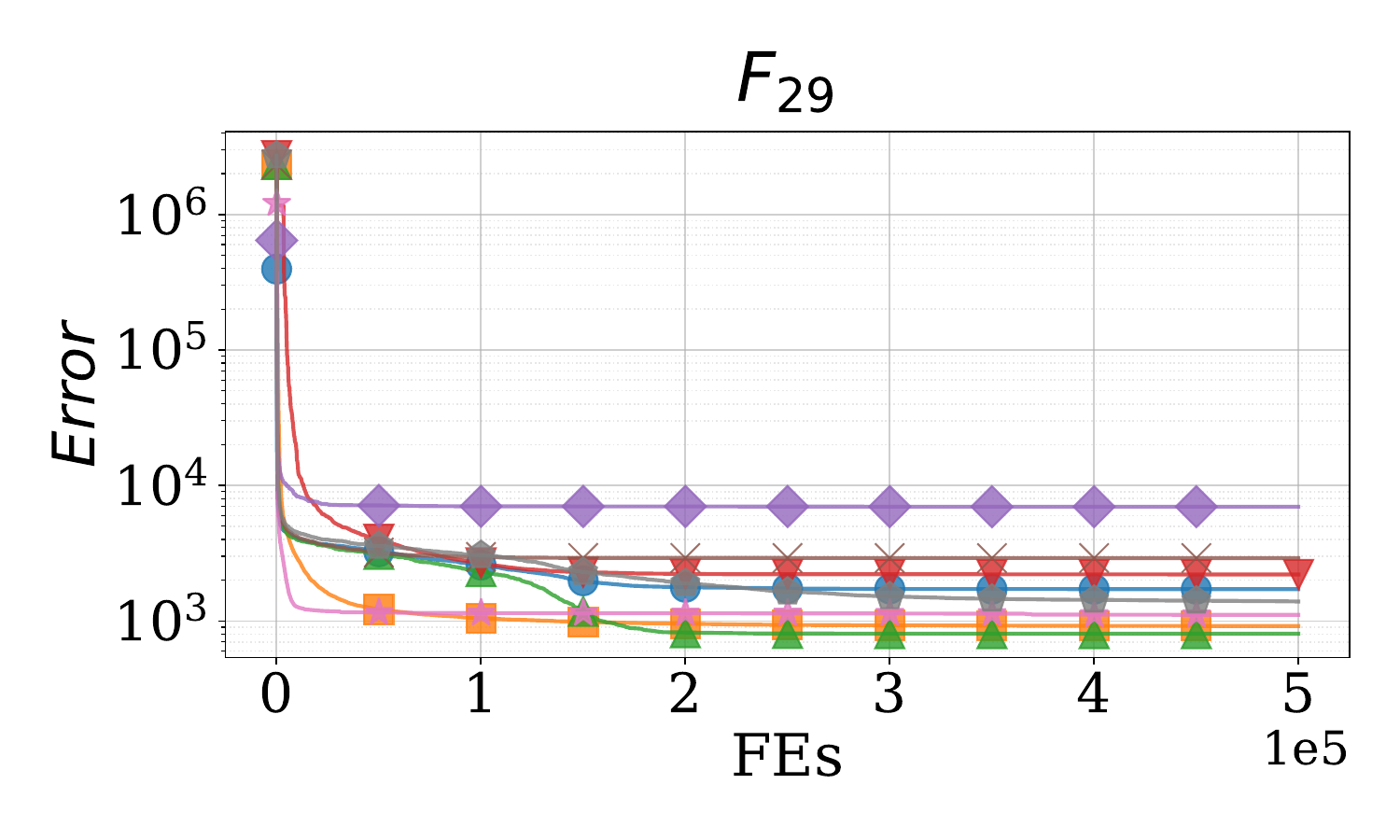}
     \end{subfigure}
     \begin{subfigure}[t]{0.23\textwidth}
         \centering
         \includegraphics[width=\linewidth, trim=0 15 0 13, clip]{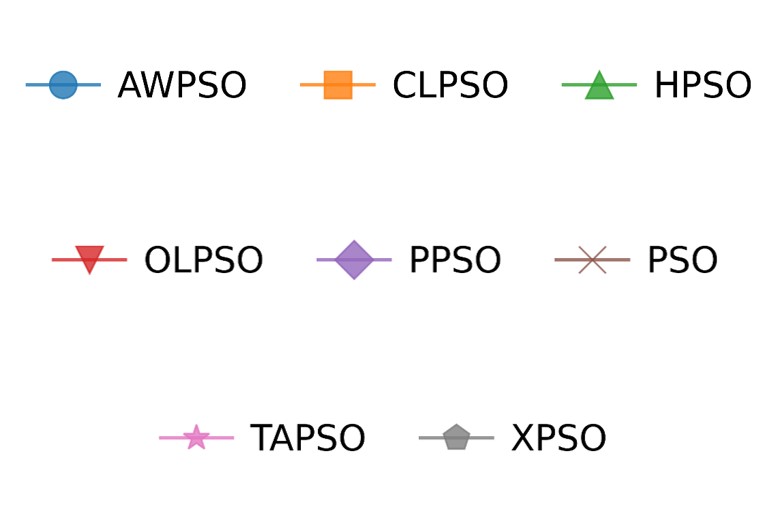}
     \end{subfigure}\hfill
     
     \caption{Mean convergence curves on the remaining 50D problems. $Error$ represents the logarithmic of mean error to the best value.}
     \label{fig_curve_appendix}
\end{figure*}

This section presents the convergence curves of the eight PSO variants for the remaining 50D functions, as shown in Fig. \ref{fig_curve_appendix}. 

\section{Population diversity curves of HPSO and HPSO-F}
\label{appendix_C}
\begin{figure*}[htbp]
     \centering

     \begin{subfigure}[t]{0.19\textwidth}
         \centering
         \includegraphics[width=\linewidth, trim=0 0 50 0, clip]{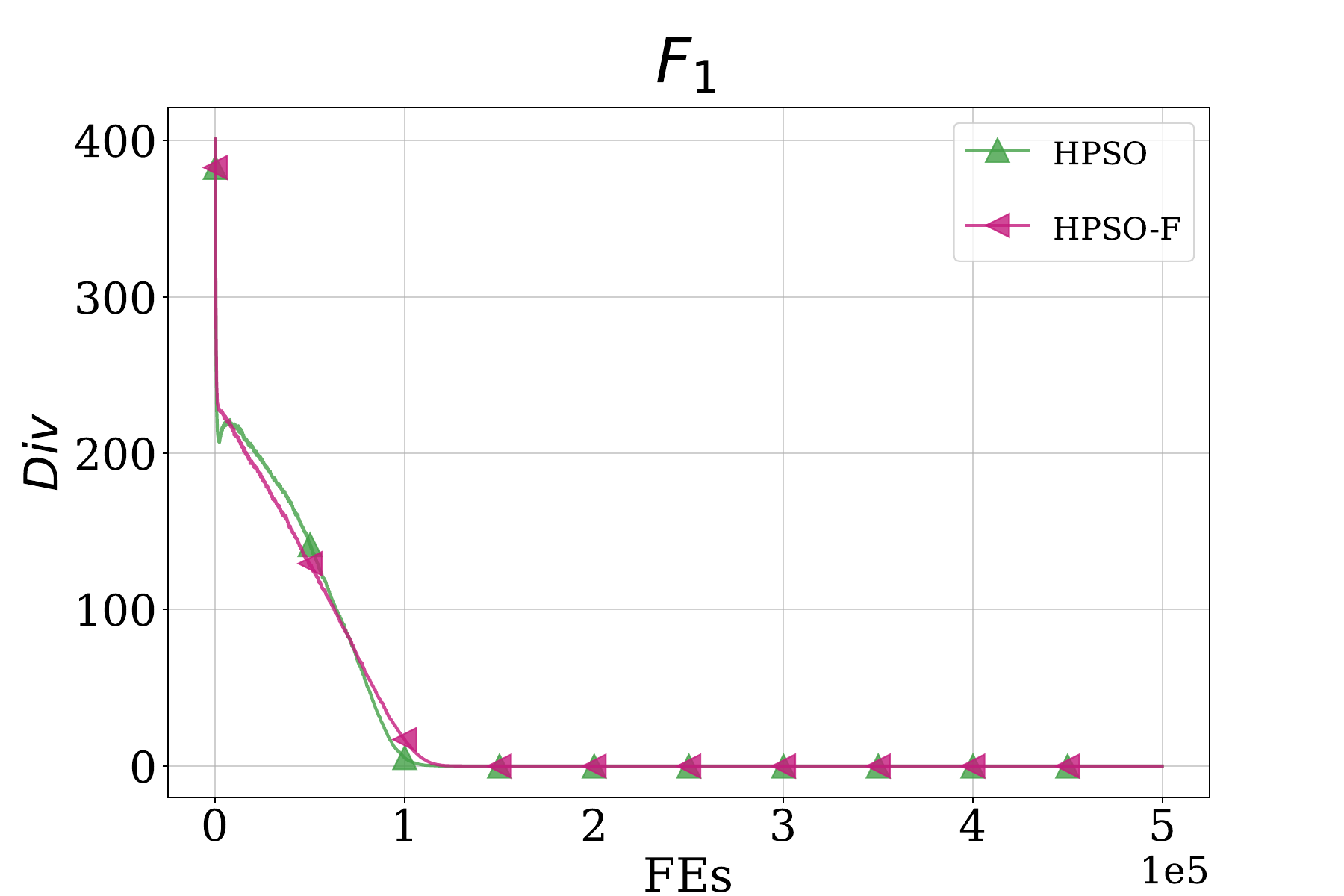}
     \end{subfigure}
     \hfill 
     \begin{subfigure}[t]{0.19\textwidth}
         \centering
         \includegraphics[width=\linewidth, trim=0 0 50 0, clip]{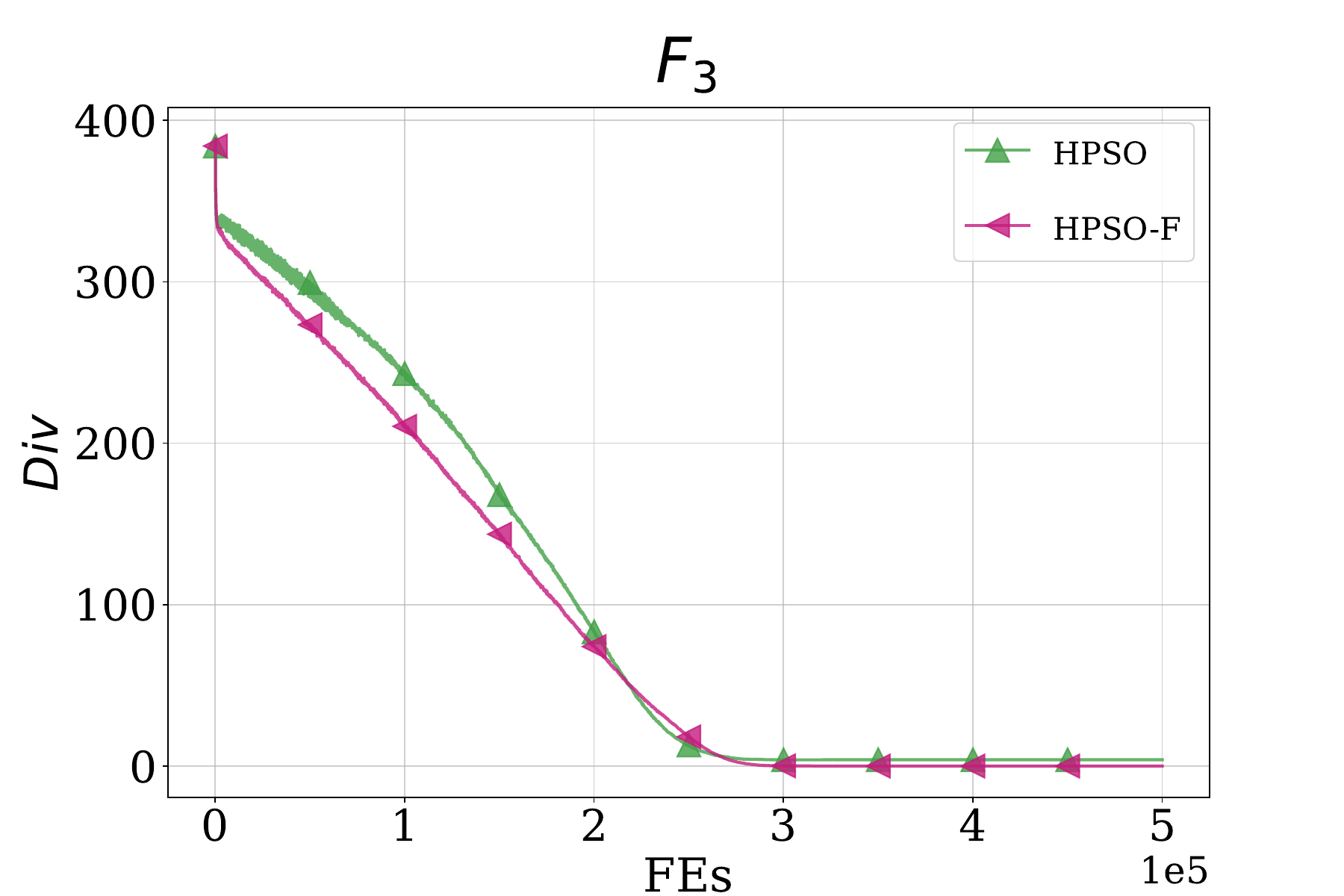}
     \end{subfigure}
     \hfill
     \begin{subfigure}[t]{0.19\textwidth}
         \centering
         \includegraphics[width=\linewidth, trim=0 0 50 0, clip]{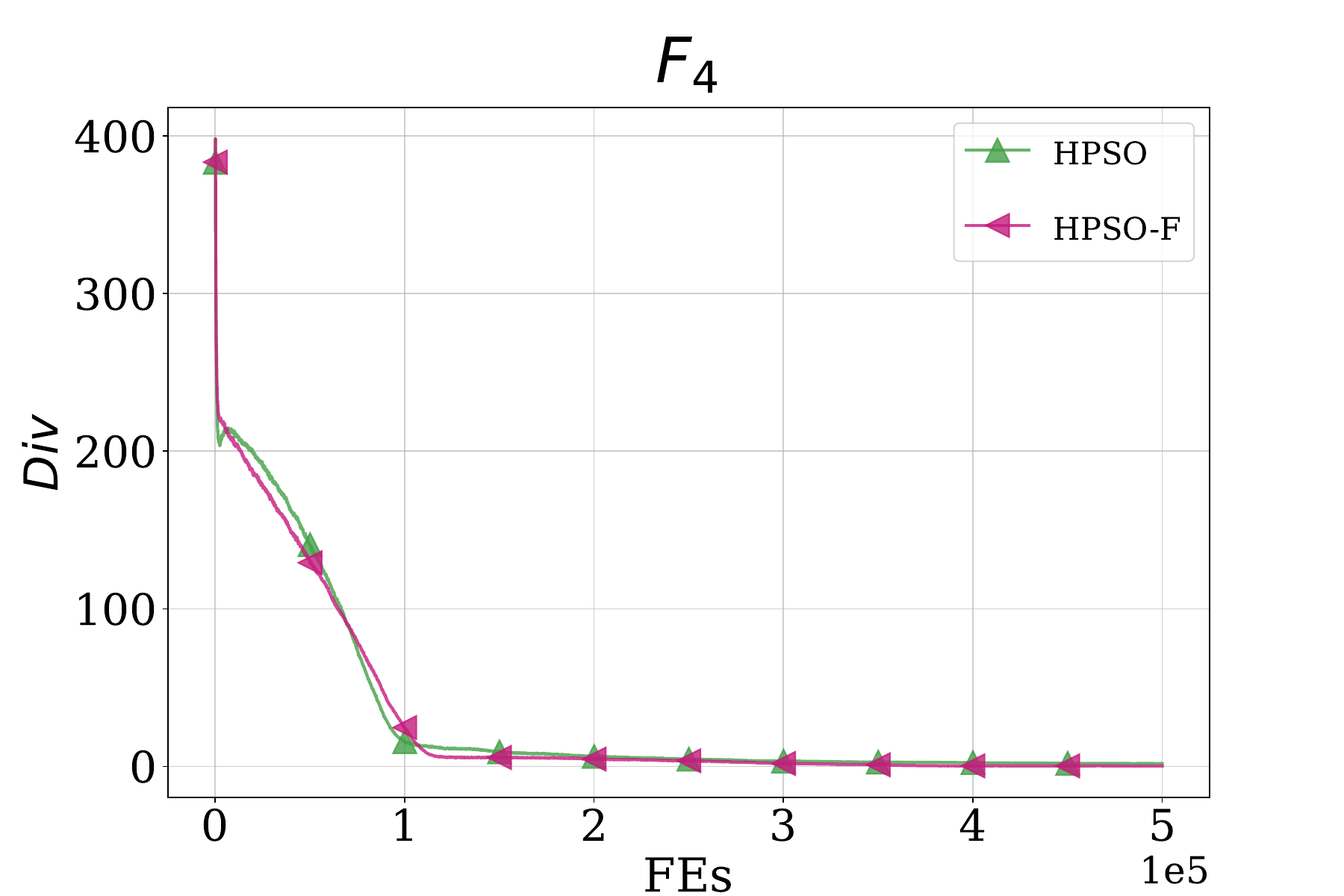}
     \end{subfigure}
     \hfill
     \begin{subfigure}[t]{0.19\textwidth}
         \centering
         \includegraphics[width=\linewidth, trim=0 0 50 0, clip]{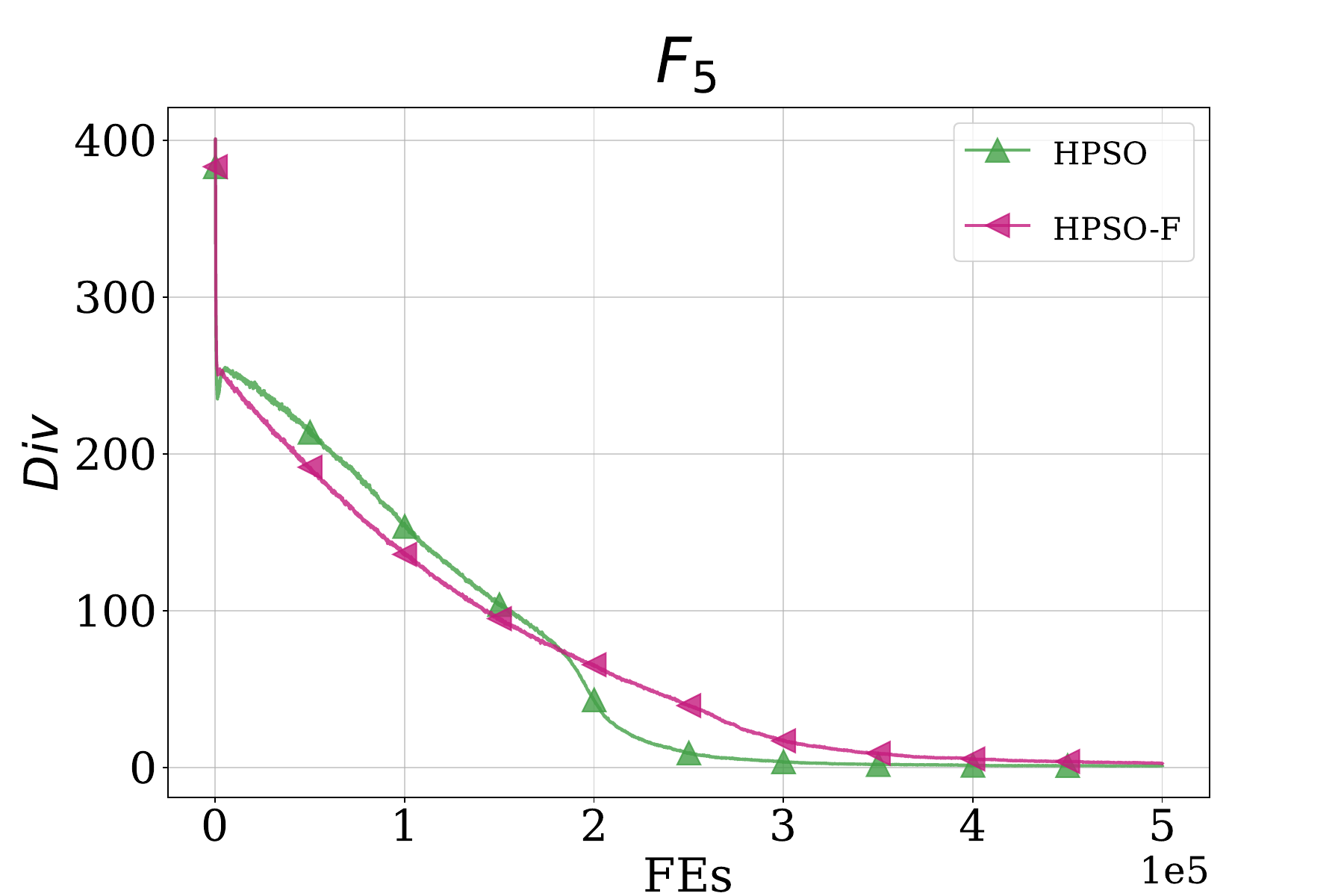}
     \end{subfigure}\hfill
      \begin{subfigure}[t]{0.19\textwidth}
         \centering
         \includegraphics[width=\linewidth, trim=0 0 50 0, clip]{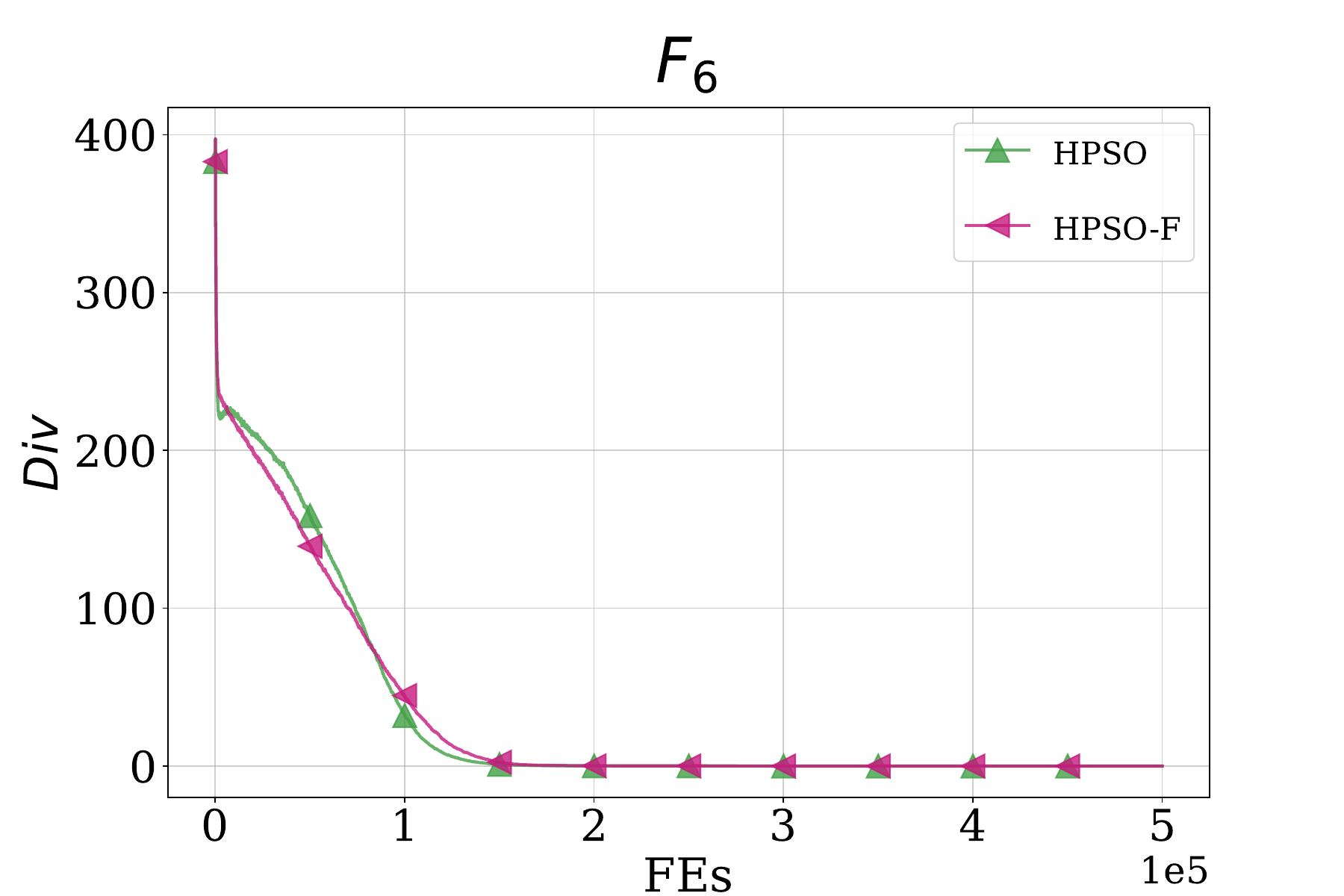}
     \end{subfigure}
     
     \vspace{2pt}
     \begin{subfigure}[t]{0.19\textwidth}
         \centering
         \includegraphics[width=\linewidth, trim=0 0 50 0, clip]{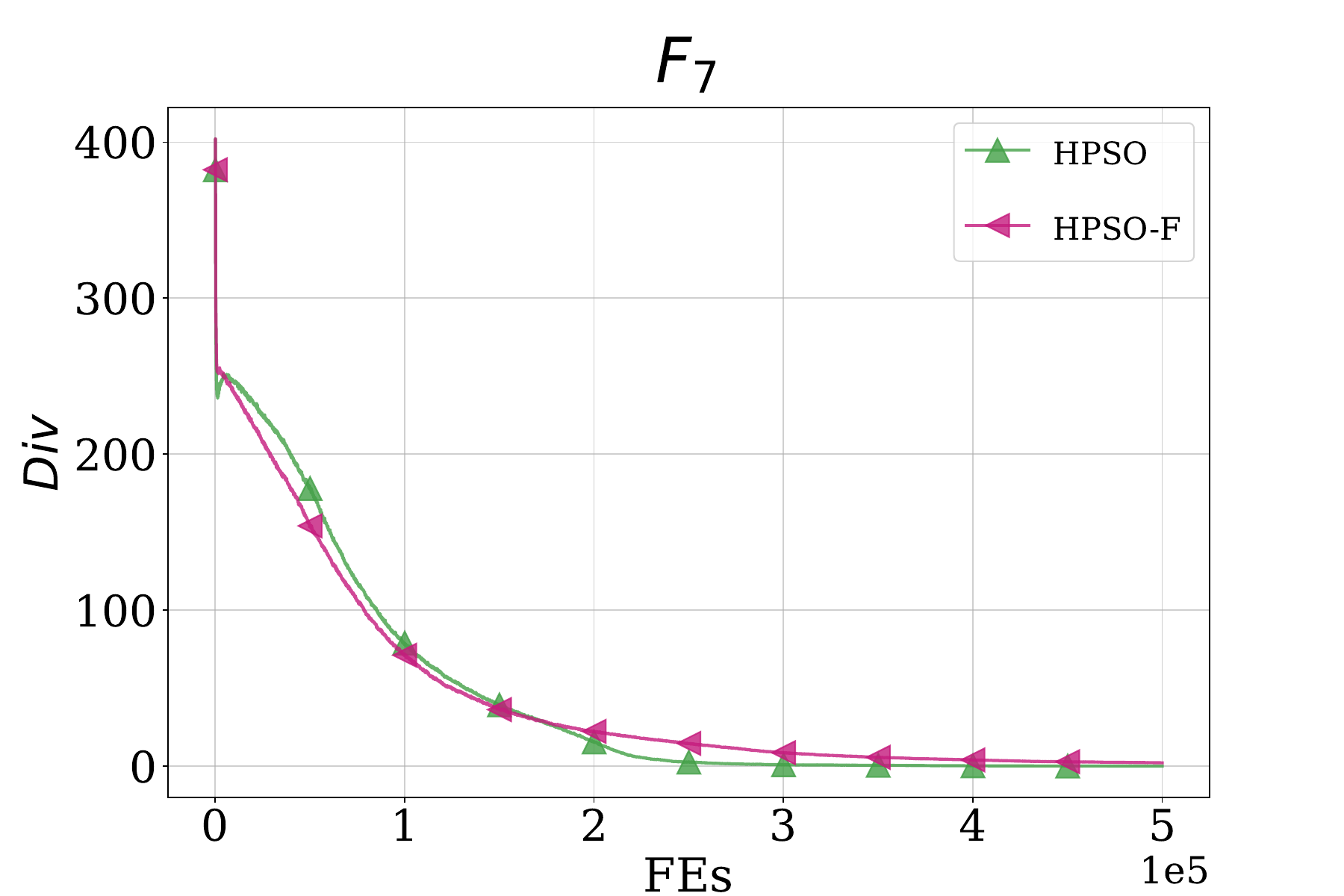}
     \end{subfigure}\hfill
     \begin{subfigure}[t]{0.19\textwidth}
         \centering
         \includegraphics[width=\linewidth, trim=0 0 50 0, clip]{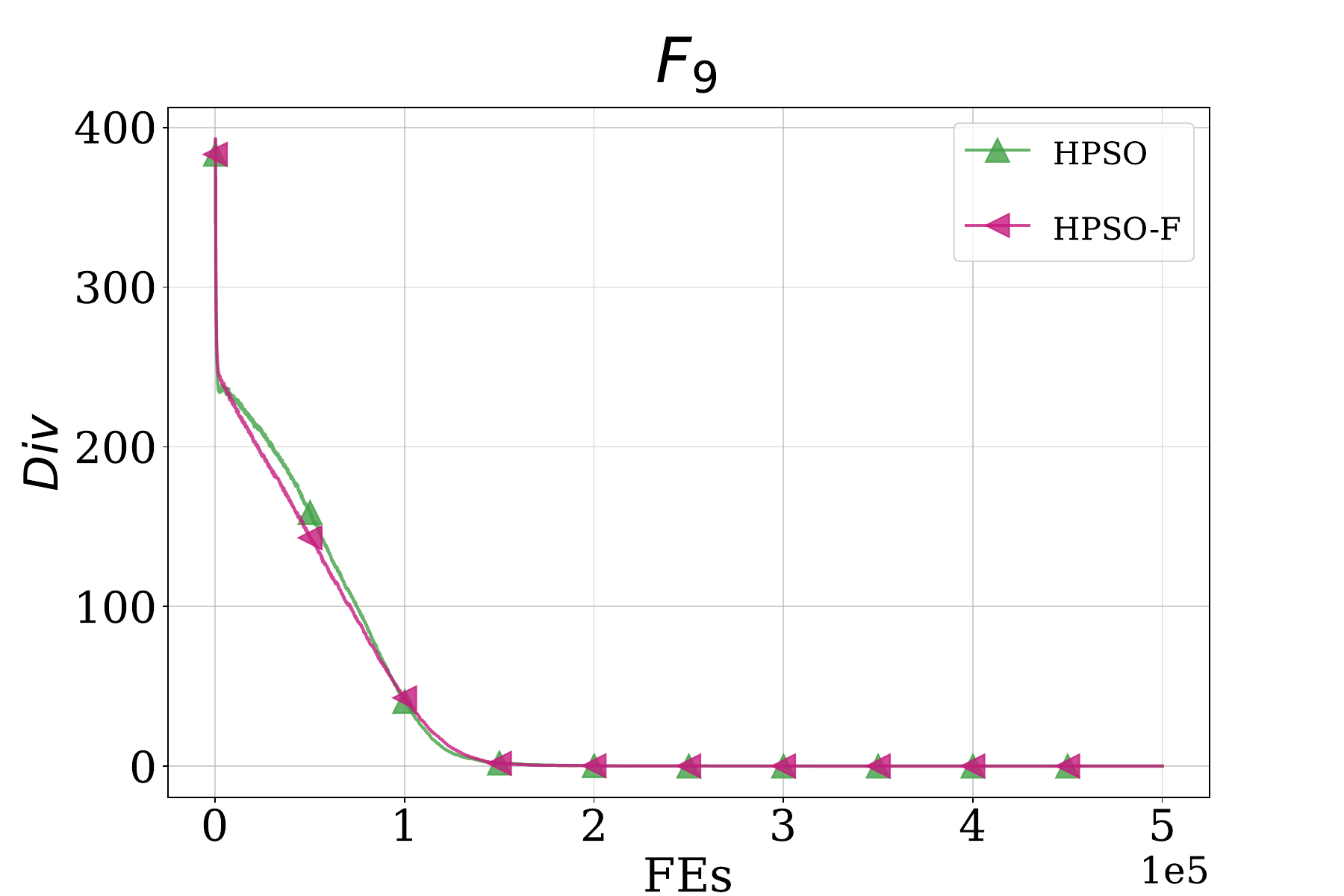}
     \end{subfigure}\hfill
     \begin{subfigure}[t]{0.19\textwidth}
         \centering
         \includegraphics[width=\linewidth, trim=0 0 50 0, clip]{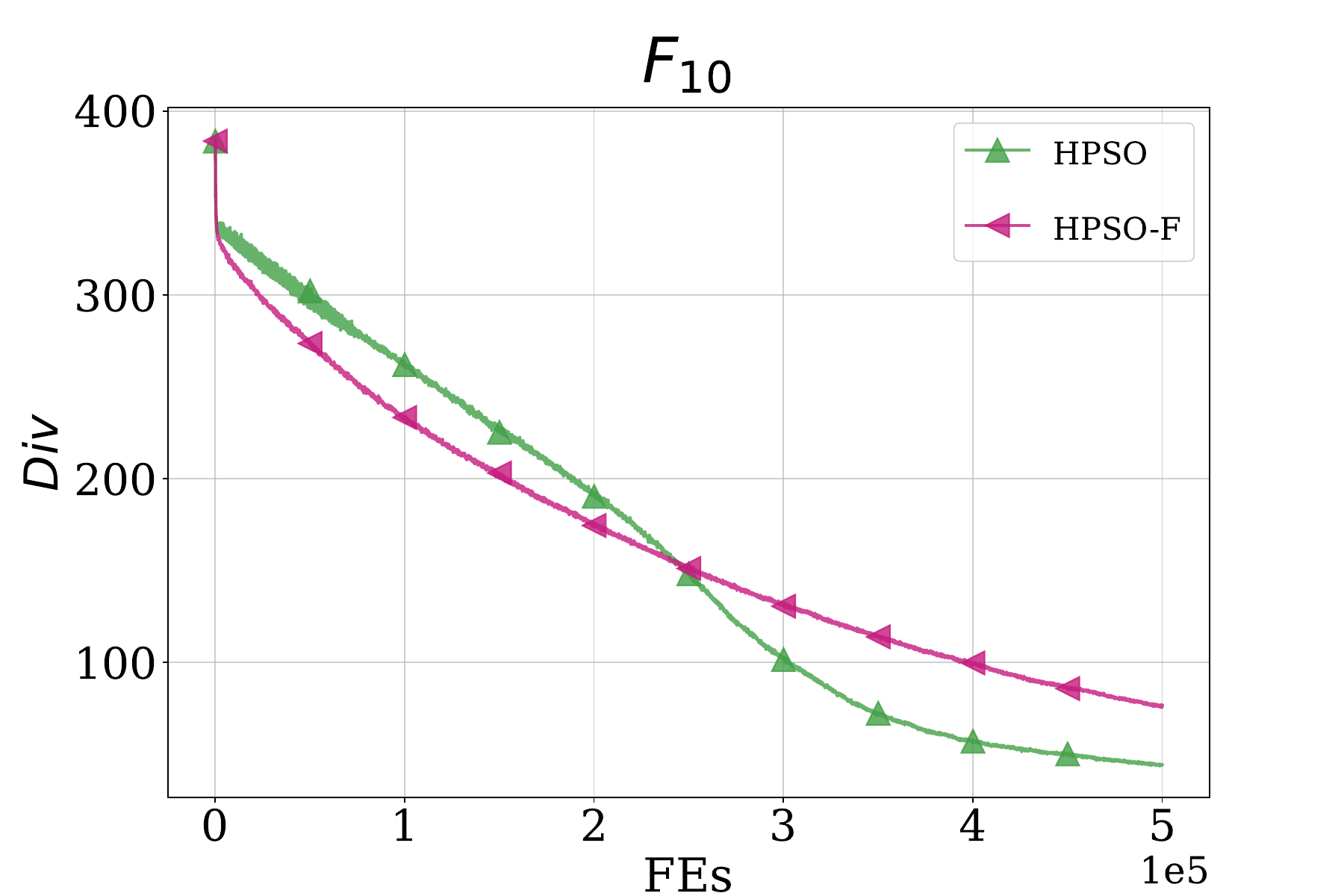}
     \end{subfigure}\hfill
     \begin{subfigure}[t]{0.19\textwidth}
         \centering
         \includegraphics[width=\linewidth, trim=0 0 50 0, clip]{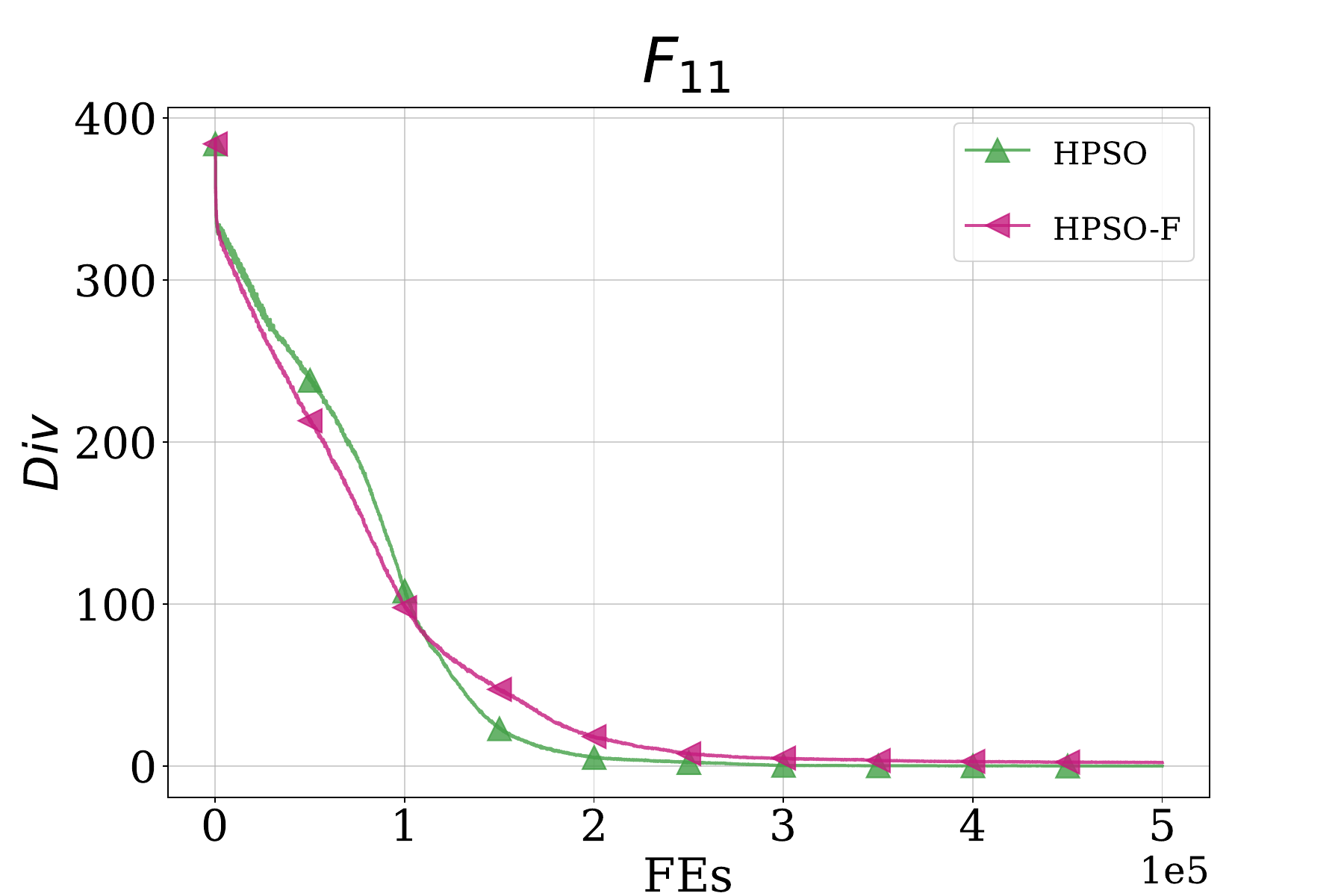}
     \end{subfigure}\hfill
     \begin{subfigure}[t]{0.19\textwidth}
         \centering
         \includegraphics[width=\linewidth, trim=0 0 50 0, clip]{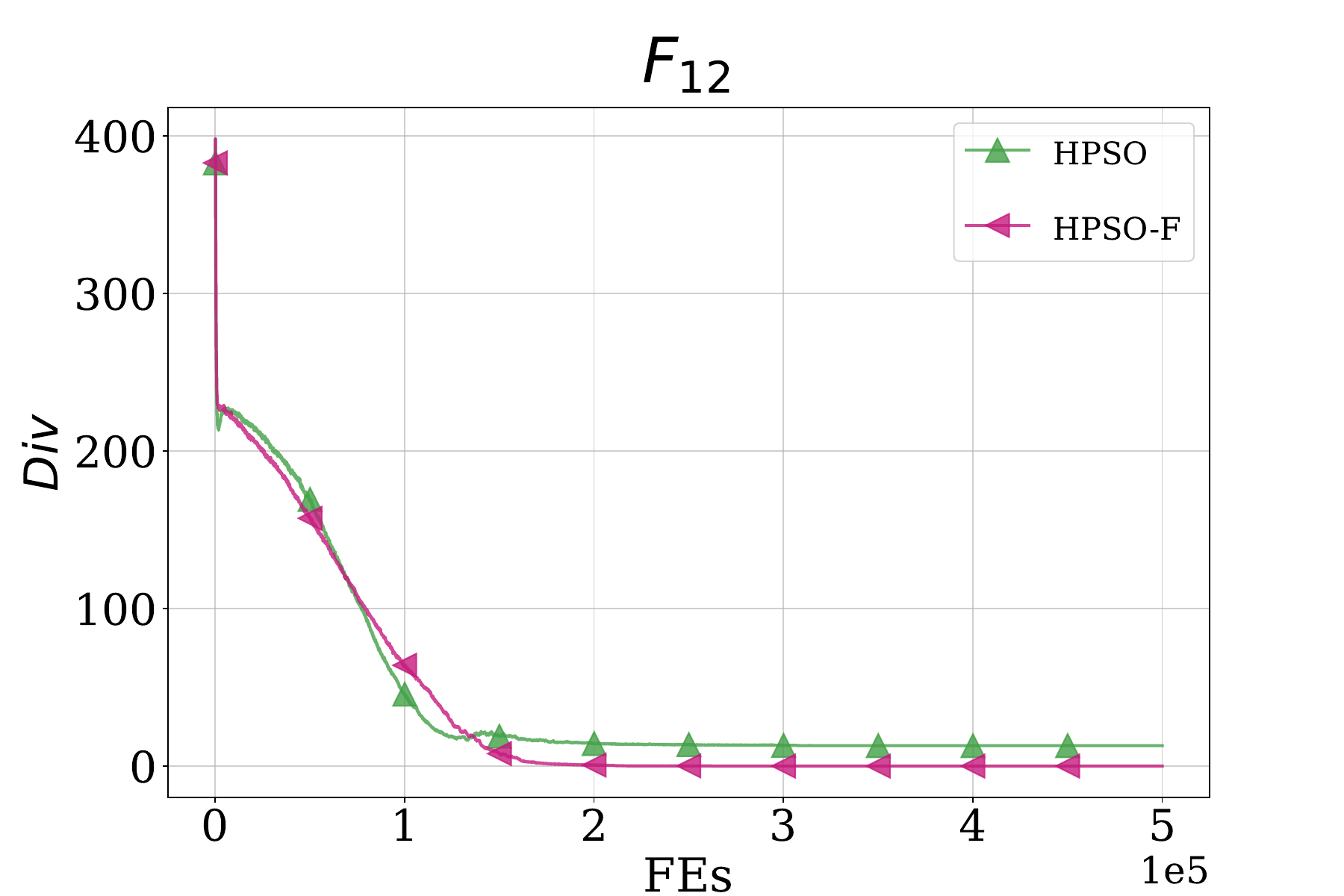}
     \end{subfigure}
     
    \vspace{2pt}
     \begin{subfigure}[t]{0.19\textwidth}
         \centering
         \includegraphics[width=\linewidth, trim=0 0 50 0, clip]{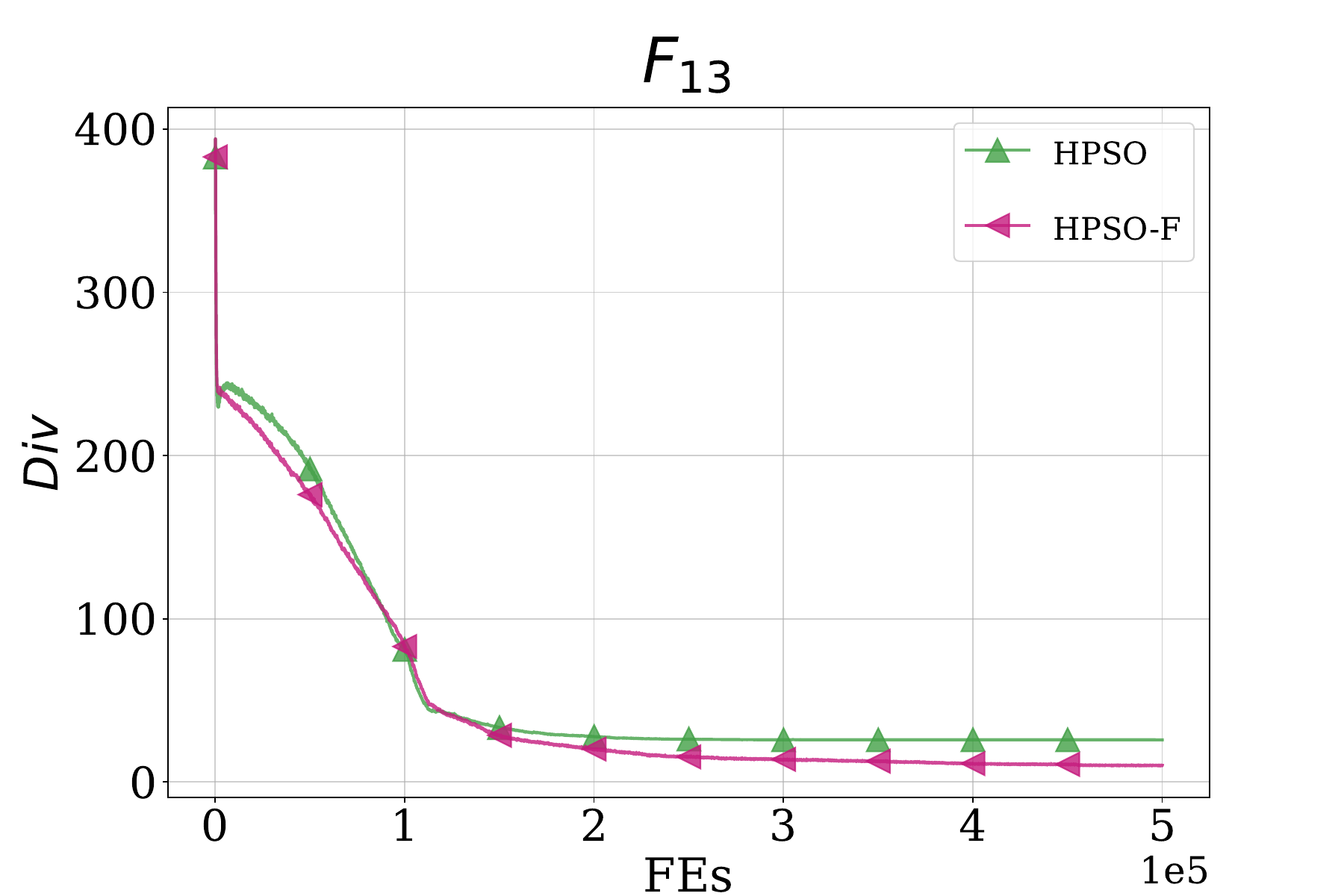}
     \end{subfigure}\hfill
     \begin{subfigure}[t]{0.19\textwidth}
         \centering
         \includegraphics[width=\linewidth, trim=0 0 50 0, clip]{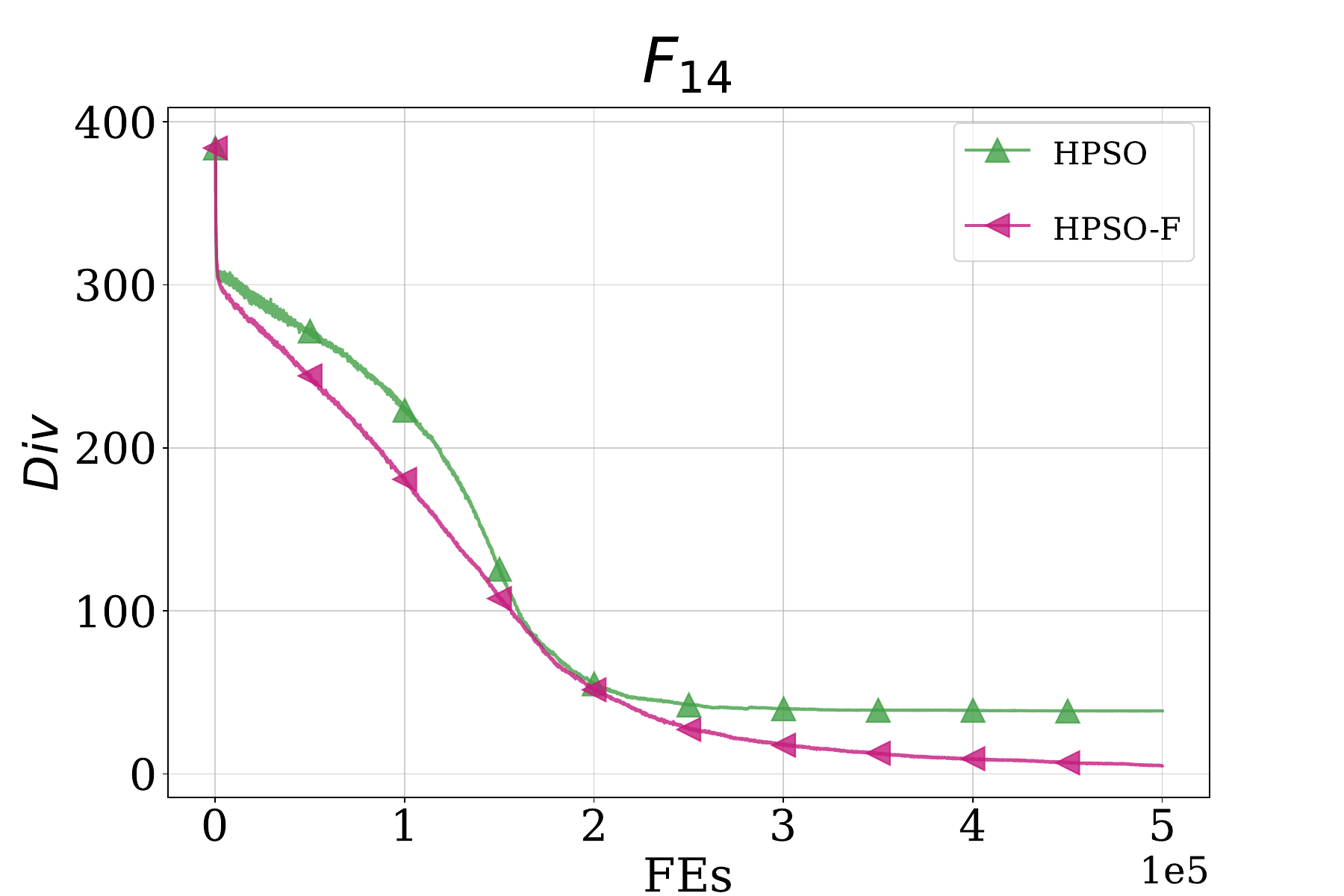}
     \end{subfigure}\hfill
     \begin{subfigure}[t]{0.19\textwidth}
         \centering
         \includegraphics[width=\linewidth, trim=0 0 50 0, clip]{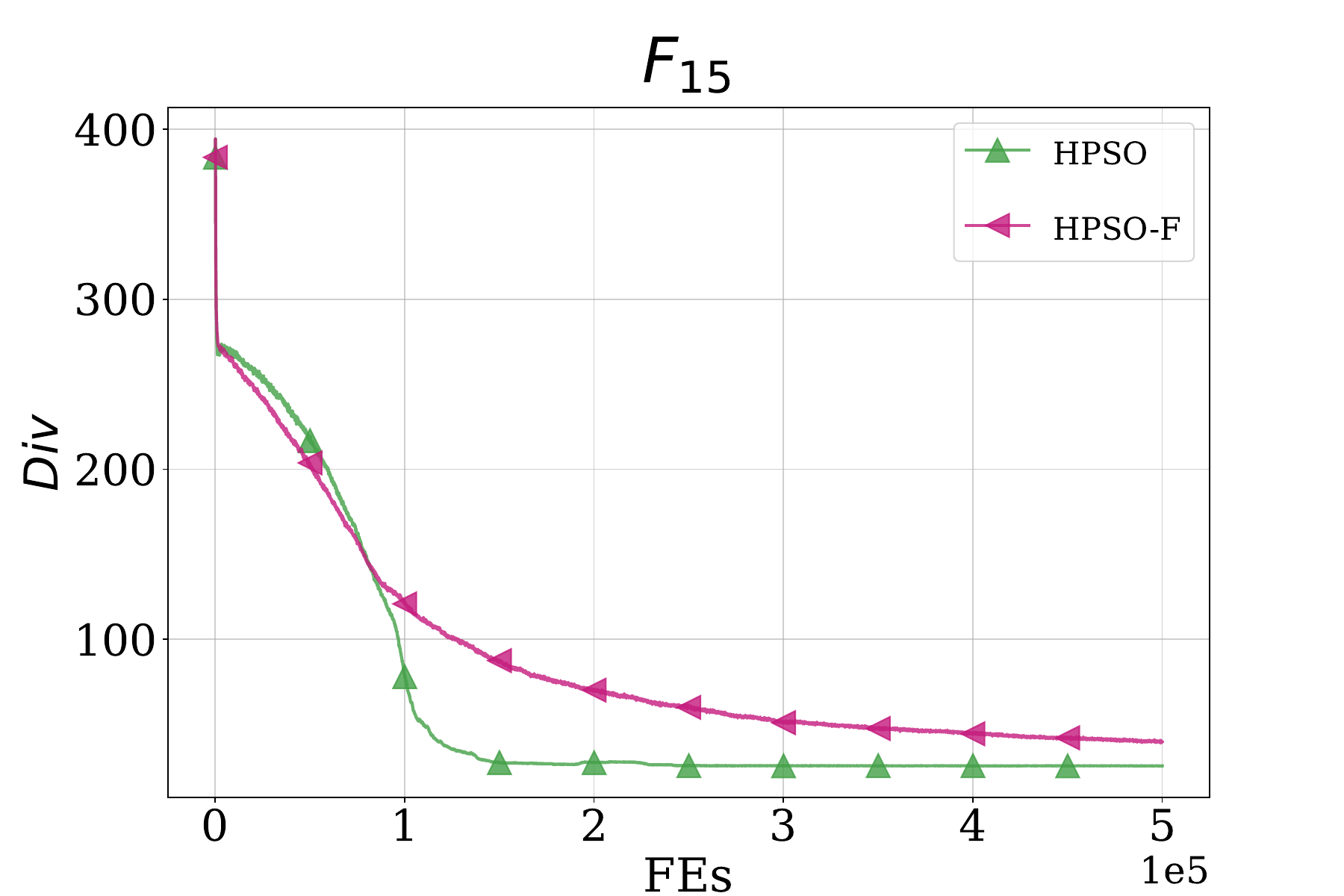}
     \end{subfigure}\hfill
     \begin{subfigure}[t]{0.19\textwidth}
         \centering
         \includegraphics[width=\linewidth, trim=0 0 50 0, clip]{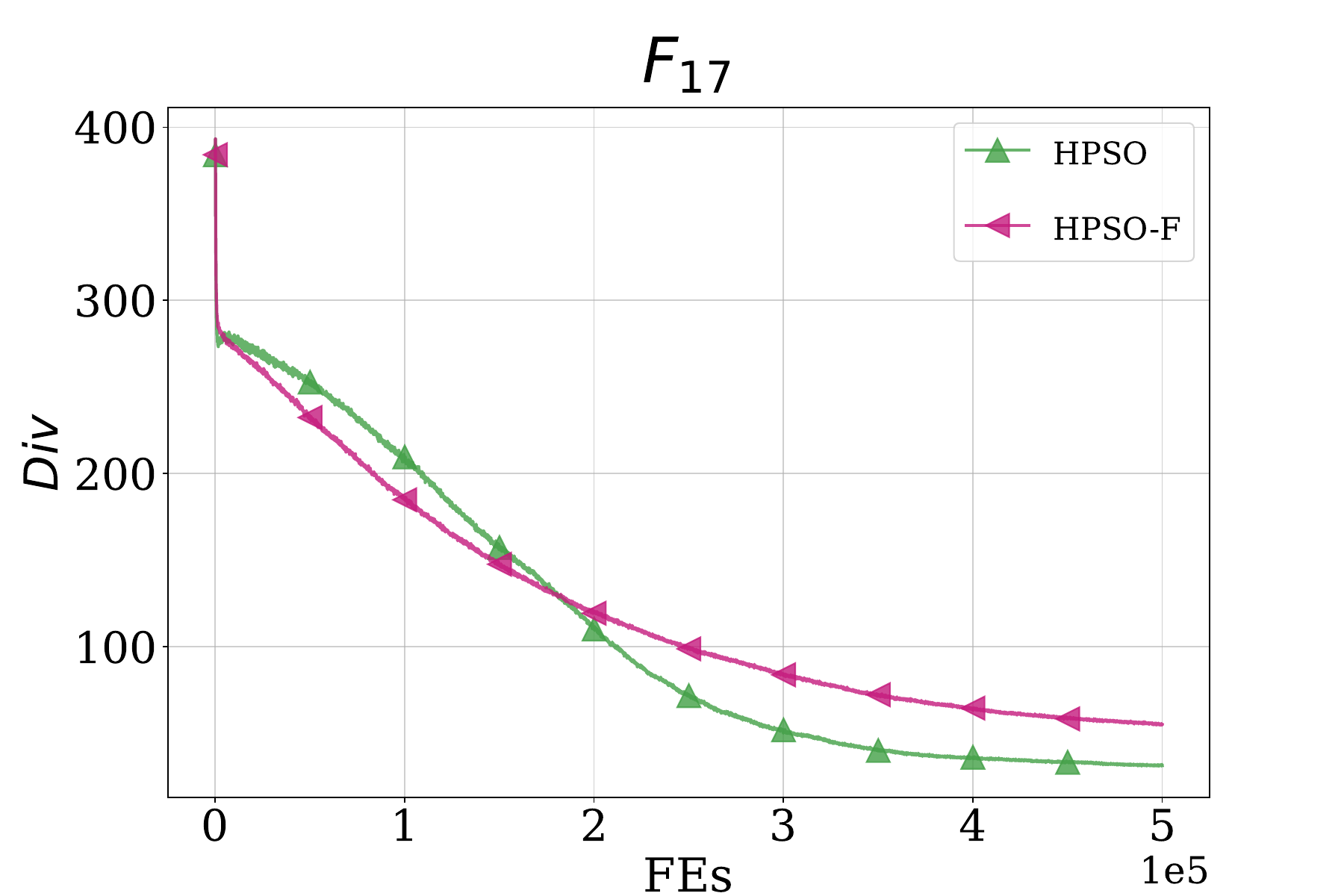}
     \end{subfigure}\hfill
     \begin{subfigure}[t]{0.19\textwidth}
         \centering
         \includegraphics[width=\linewidth, trim=0 0 50 0, clip]{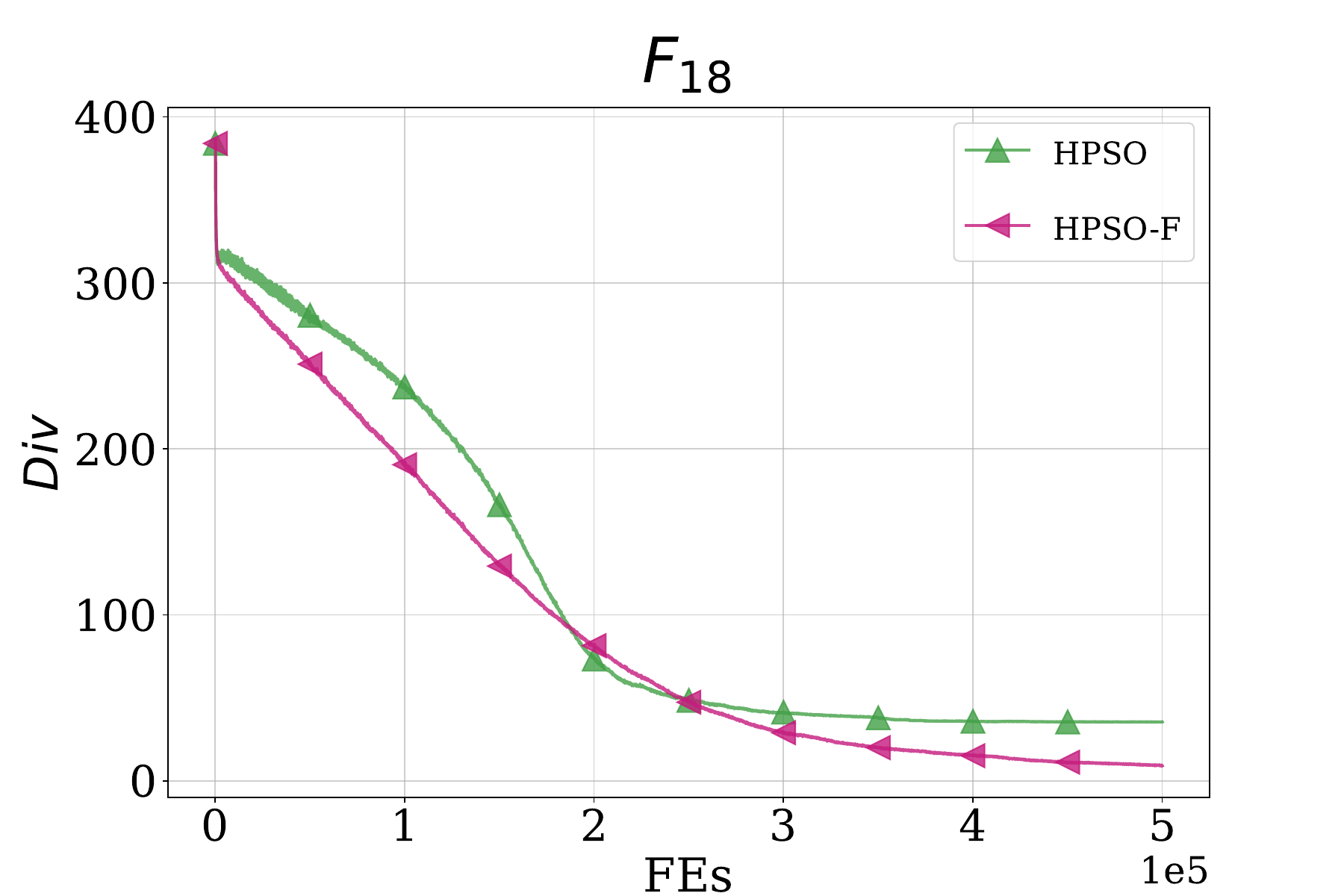}
     \end{subfigure}
     
     \vspace{2pt}
     \begin{subfigure}[t]{0.19\textwidth}
         \centering
         \includegraphics[width=\linewidth, trim=0 0 50 0, clip]{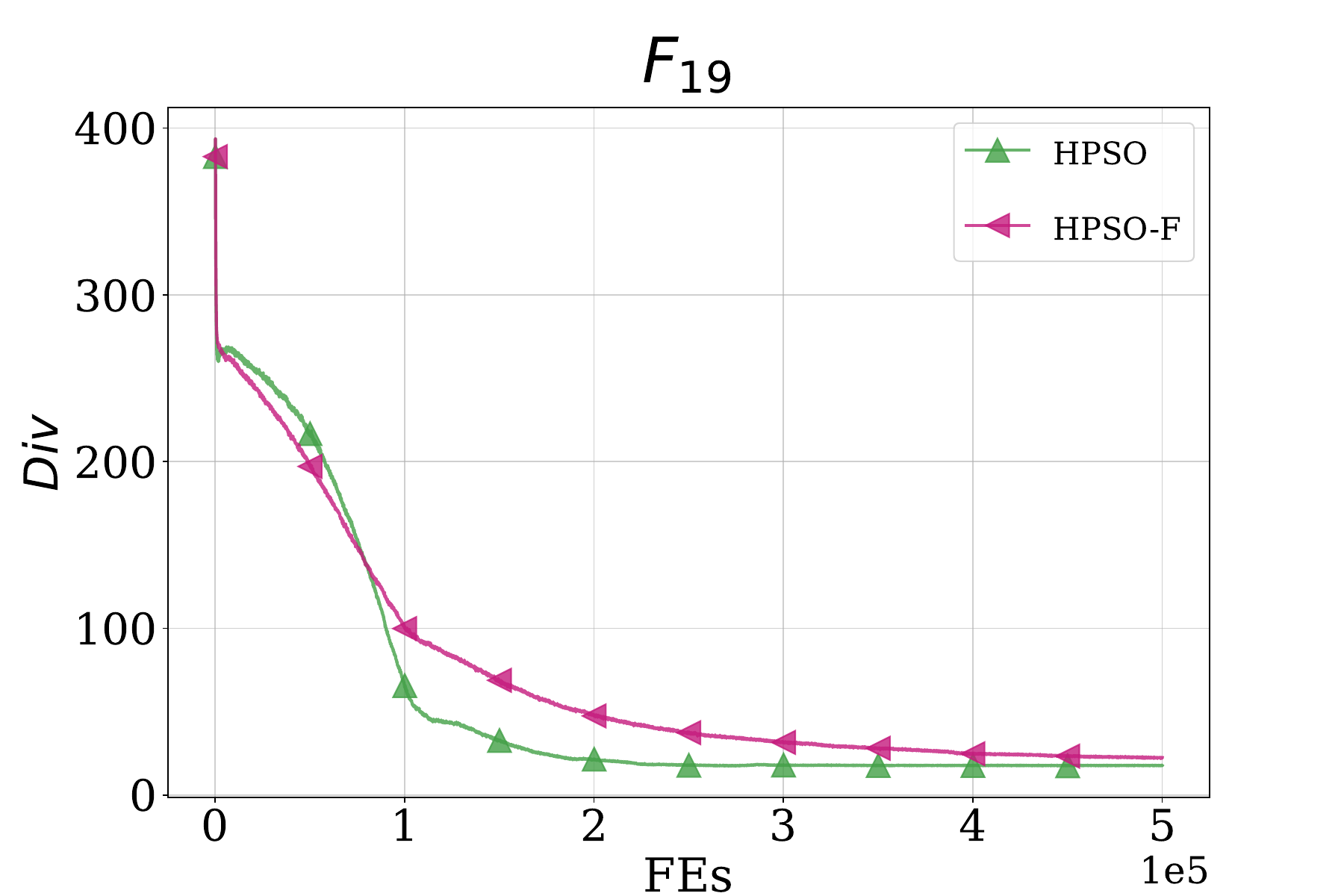}
     \end{subfigure}\hfill
     \begin{subfigure}[t]{0.19\textwidth}
         \centering
         \includegraphics[width=\linewidth, trim=0 0 50 0, clip]{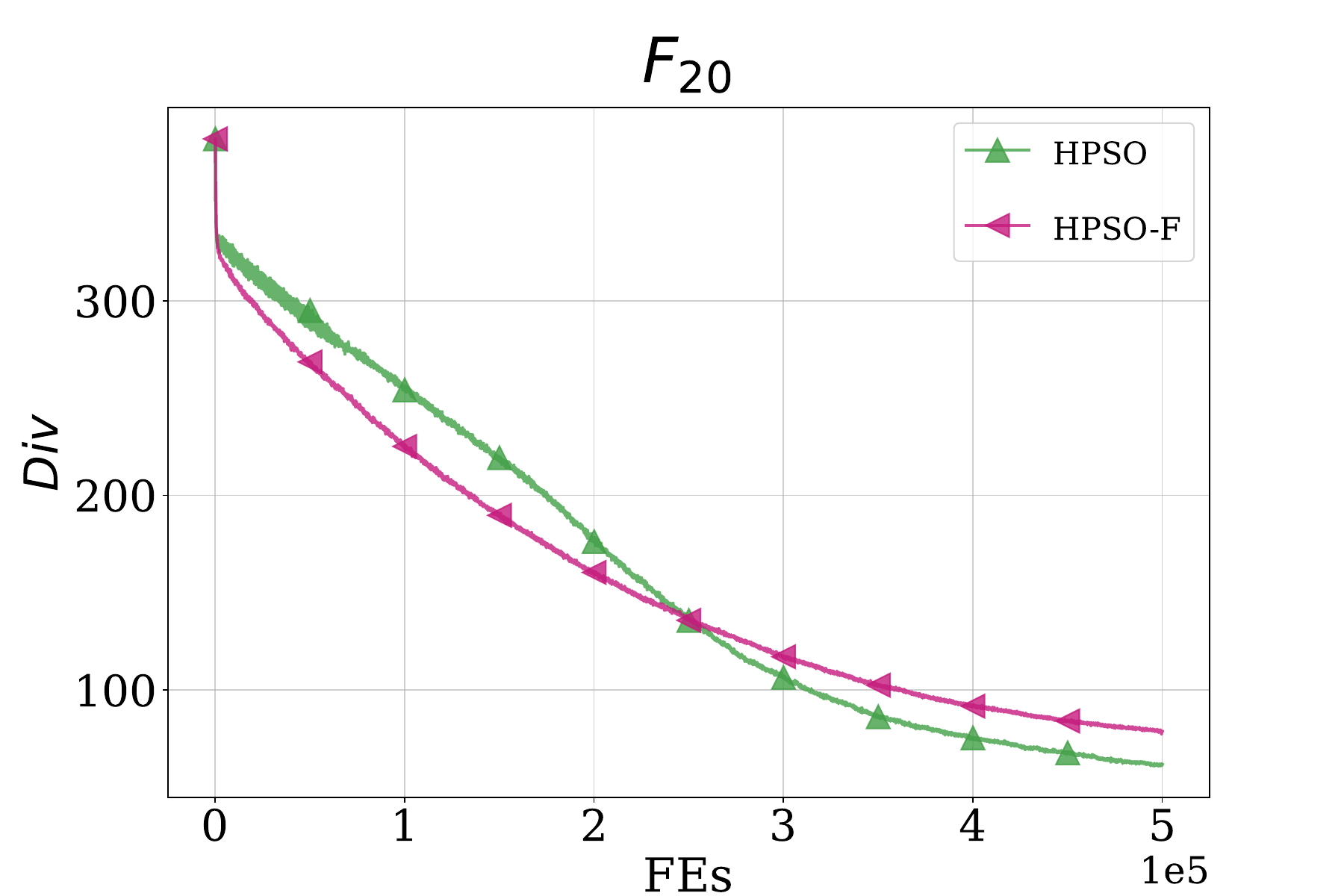}
     \end{subfigure}\hfill
     \begin{subfigure}[t]{0.19\textwidth}
         \centering
         \includegraphics[width=\linewidth, trim=0 0 50 0, clip]{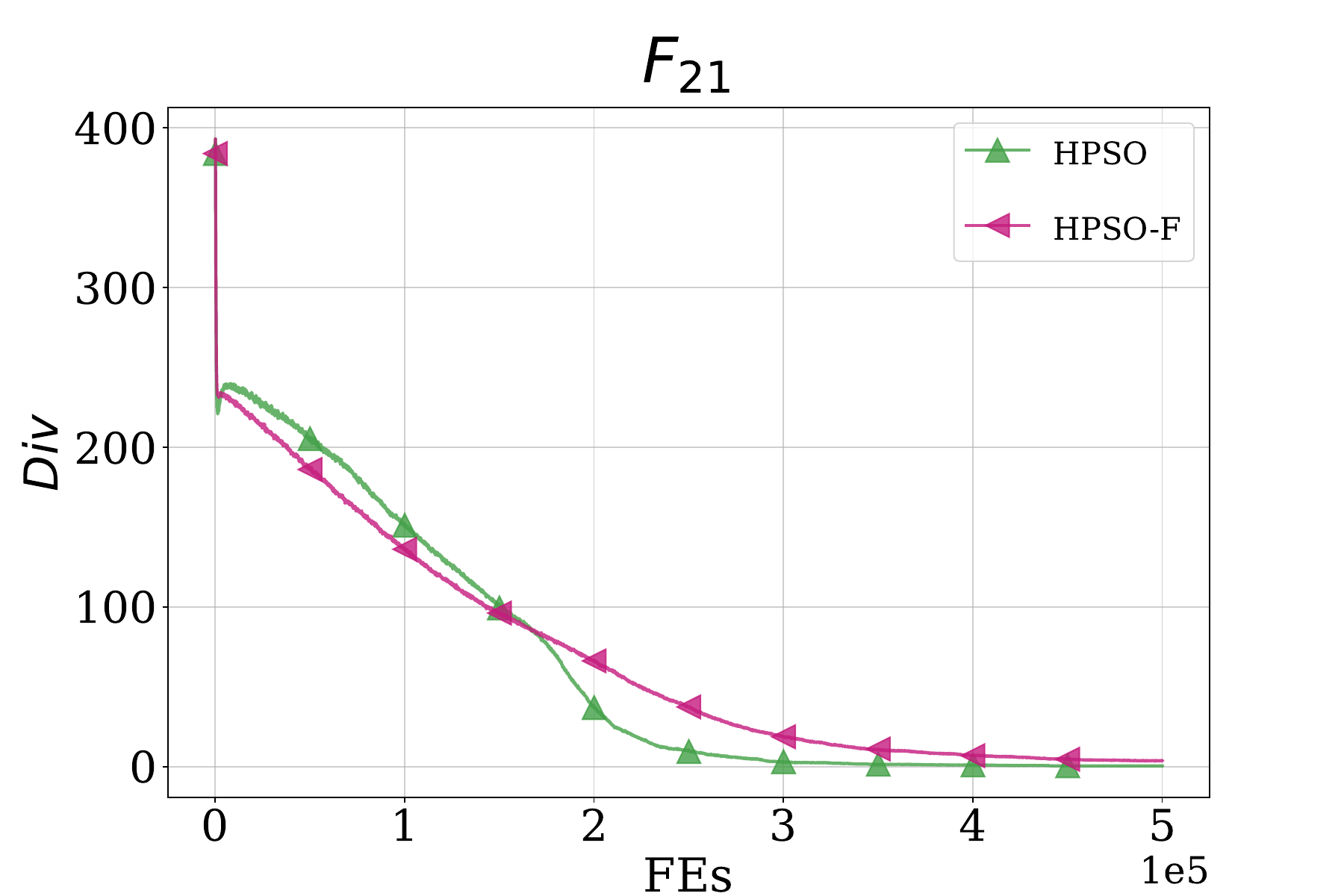}
     \end{subfigure}\hfill
     \begin{subfigure}[t]{0.19\textwidth}
         \centering
         \includegraphics[width=\linewidth, trim=0 0 50 0, clip]{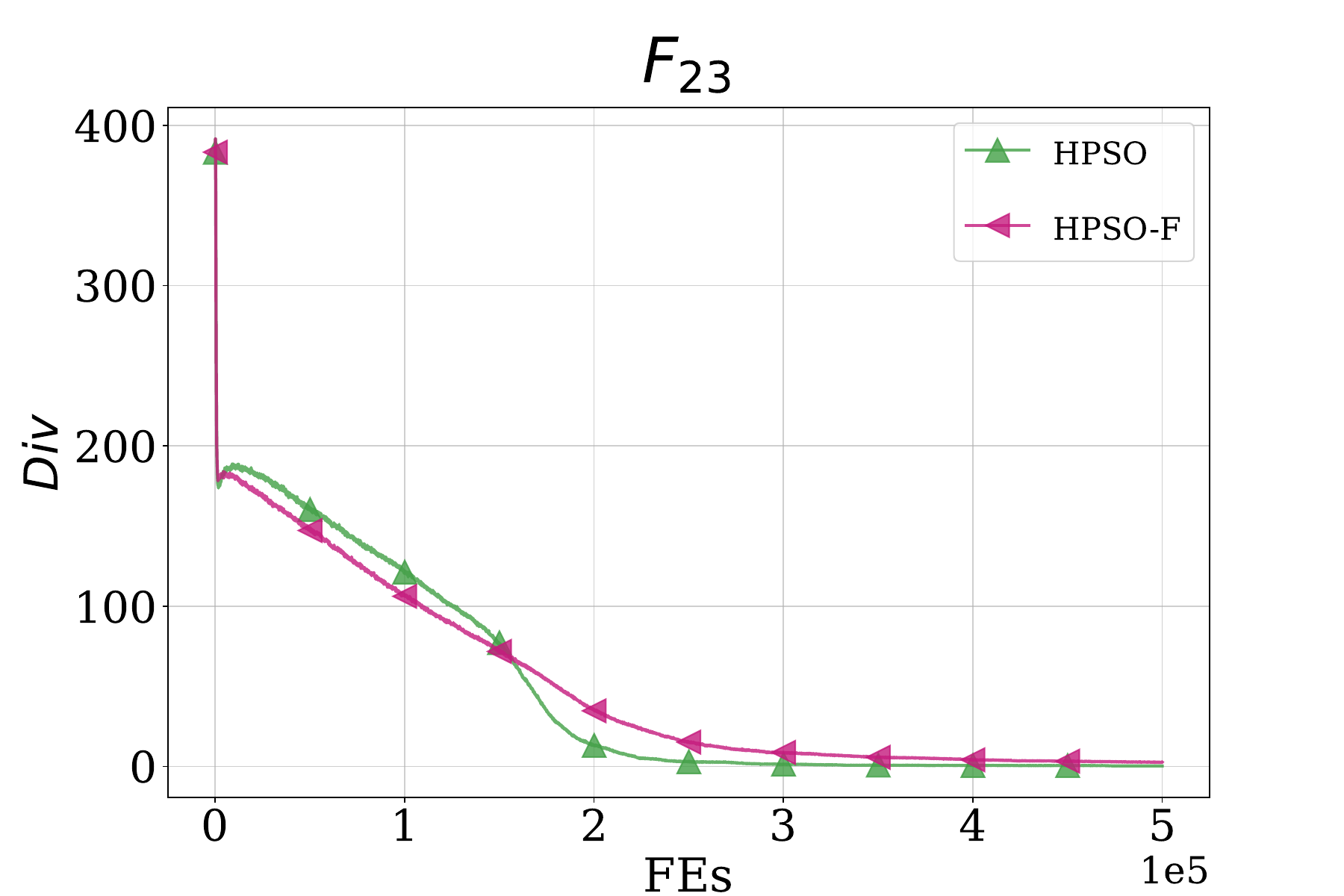}
     \end{subfigure}\hfill
     \begin{subfigure}[t]{0.19\textwidth}
         \centering
         \includegraphics[width=\linewidth, trim=0 0 50 0, clip]{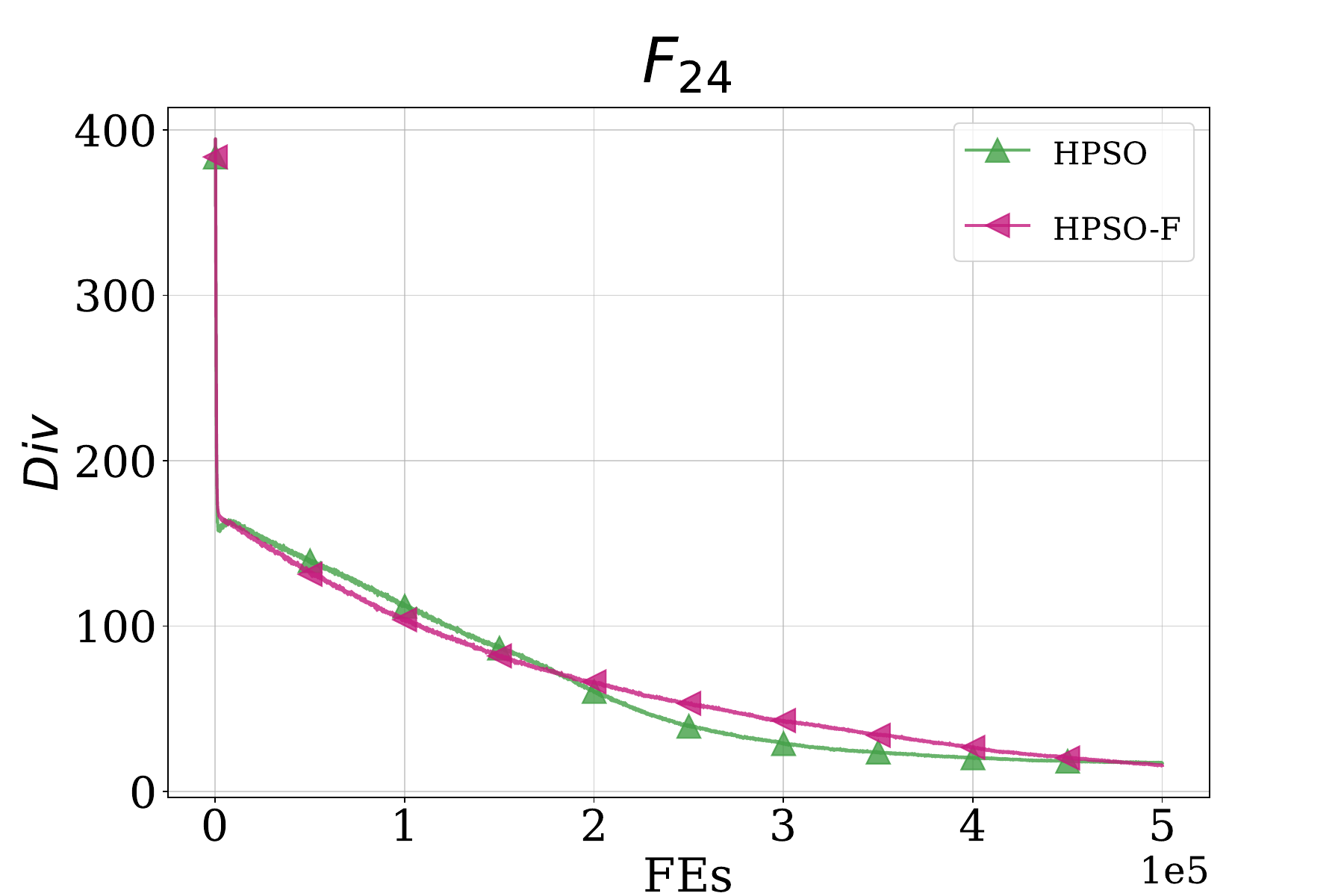}
     \end{subfigure}

     \vspace{2pt}
     \begin{subfigure}[t]{0.19\textwidth}
         \centering
         \includegraphics[width=\linewidth, trim=0 0 50 0, clip]{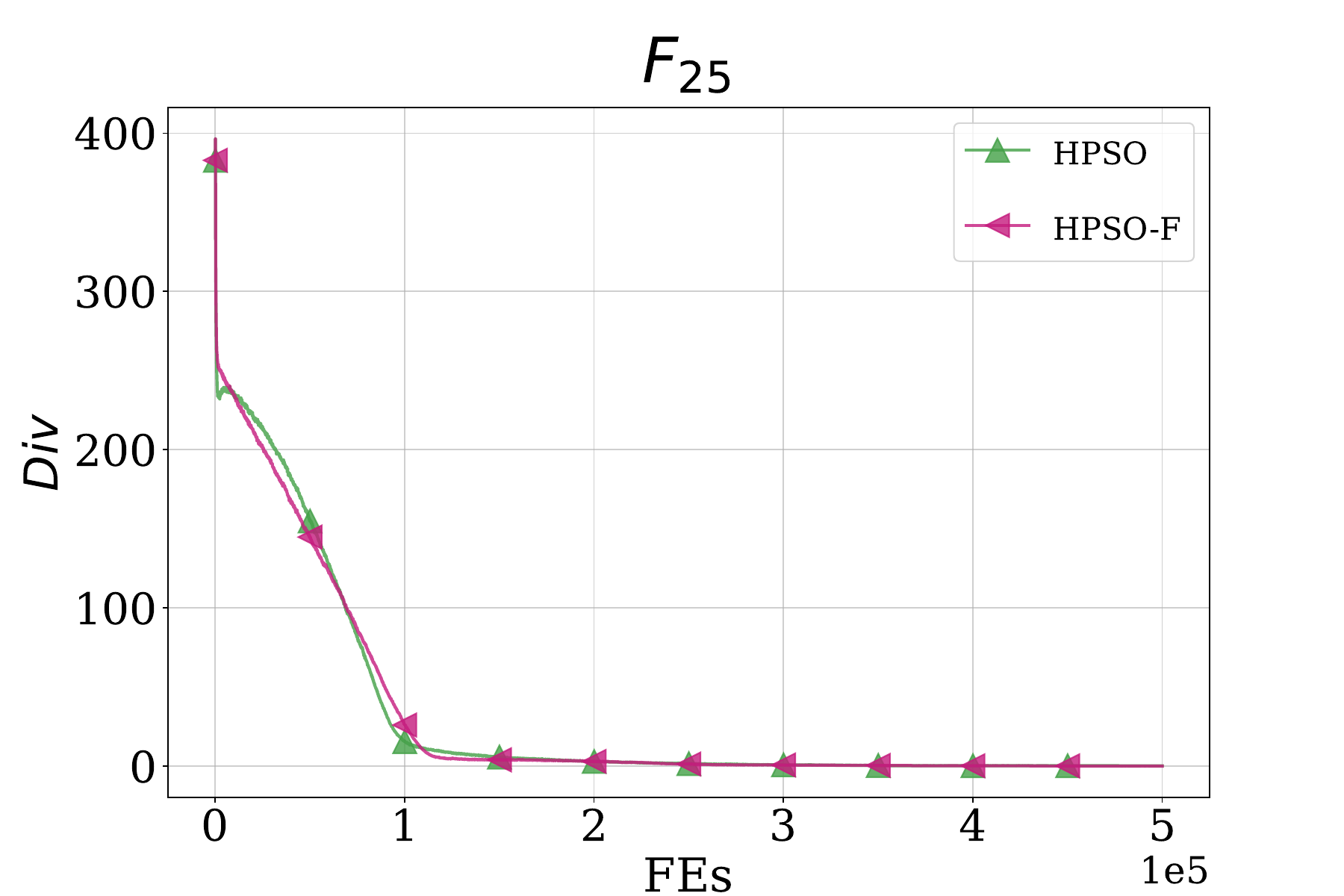}
     \end{subfigure}\hfill
     \begin{subfigure}[t]{0.19\textwidth}
         \centering
         \includegraphics[width=\linewidth, trim=0 0 50 0, clip]{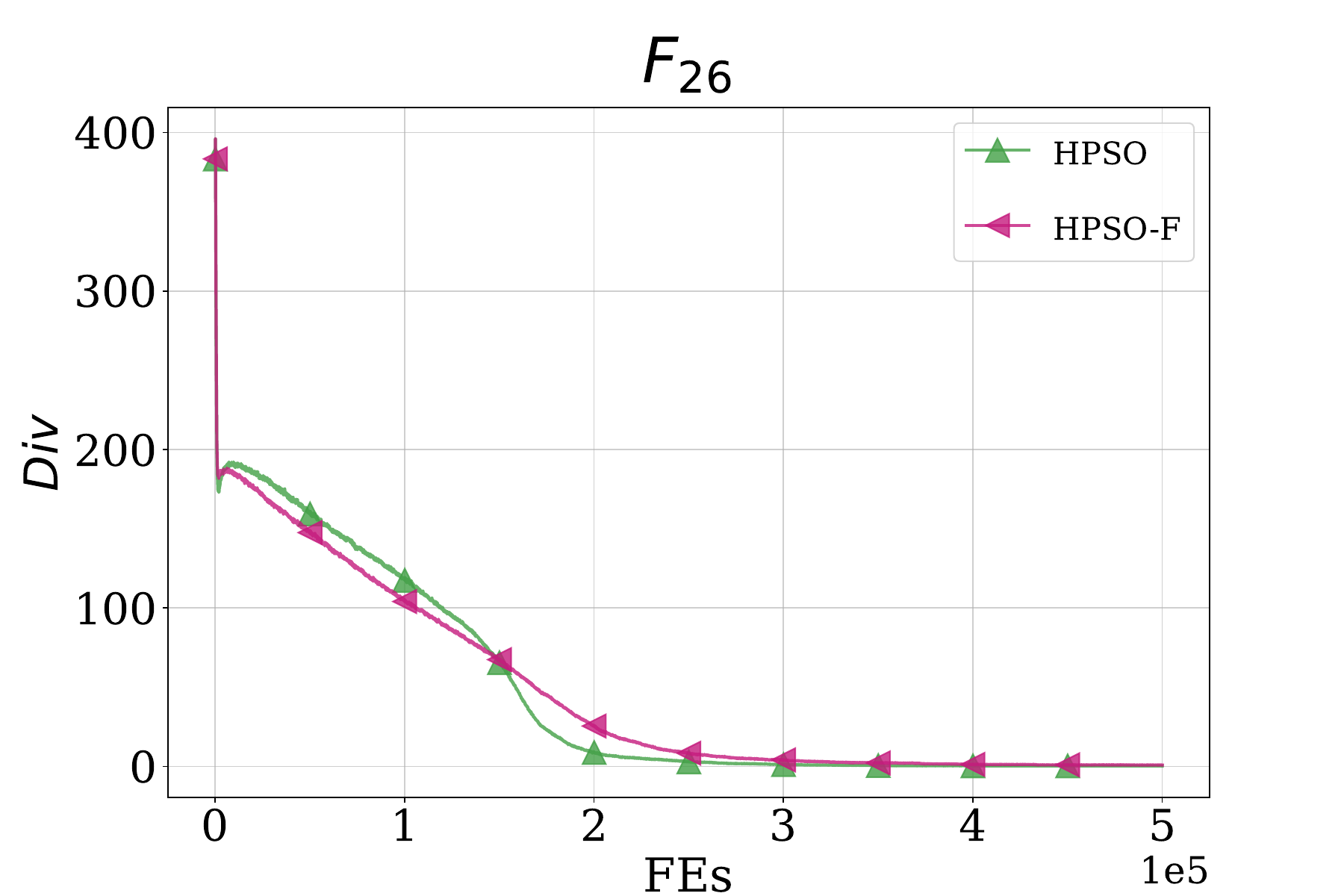}
     \end{subfigure}\hfill
     \begin{subfigure}[t]{0.19\textwidth}
         \centering
         \includegraphics[width=\linewidth, trim=0 0 50 0, clip]{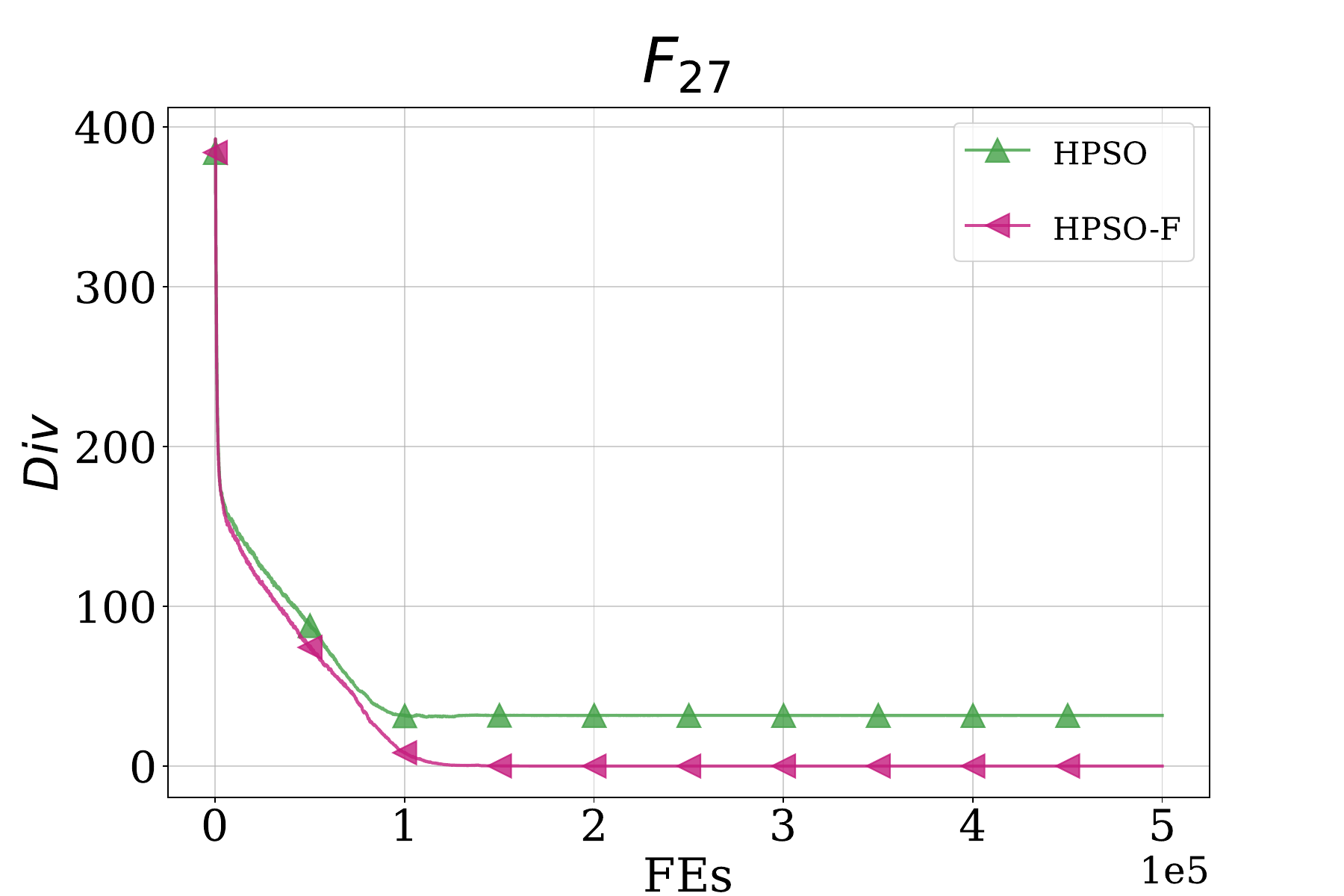}
     \end{subfigure}\hfill
     \begin{subfigure}[t]{0.19\textwidth}
         \centering
         \includegraphics[width=\linewidth, trim=0 0 50 0, clip]{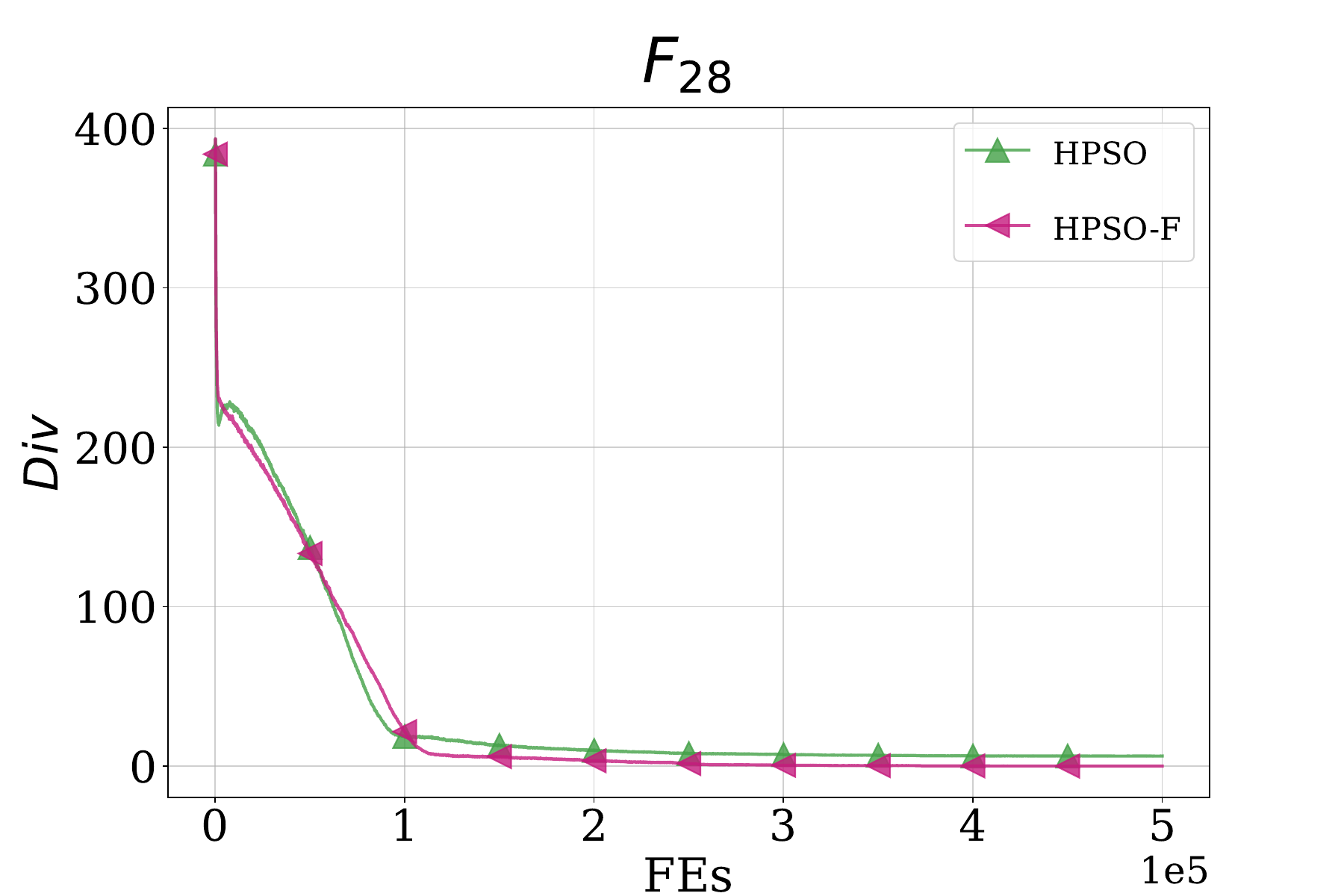}
     \end{subfigure}\hfill
     \begin{subfigure}[t]{0.19\textwidth}
         \centering
         \includegraphics[width=\linewidth, trim=0 0 50 0, clip]{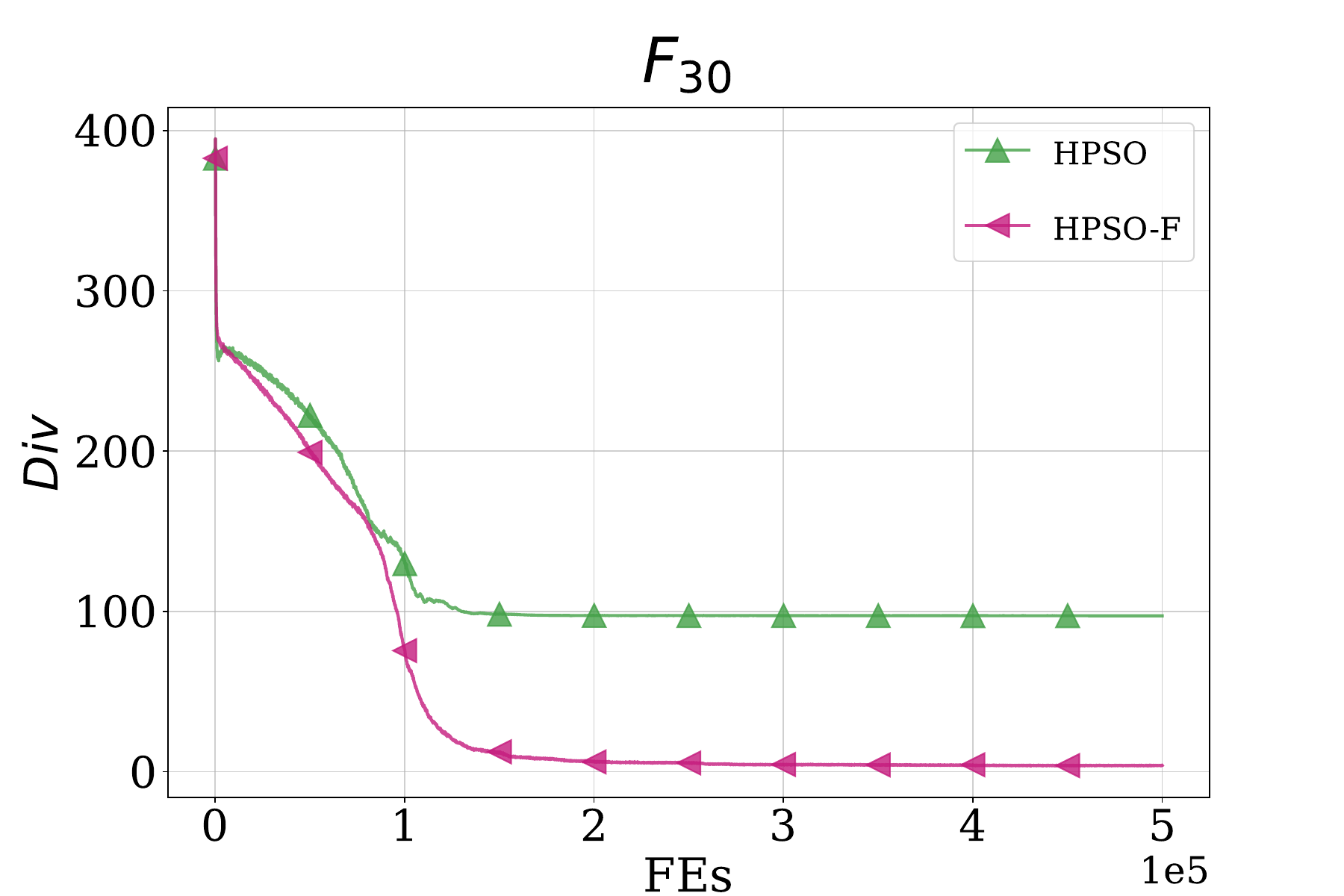}
     \end{subfigure}\hfill
     
     \caption{Population diversity curves of HPSO and HPSO-F on the remaining 50D problems. }
\label{fig_div_appendix}
\end{figure*}

This section presents the population diversity curves of HPSO and HPSO-F on the remaining 50D functions, as shown in Fig. \ref{fig_div_appendix}.

\clearpage

\begin{table*}[]
\centering
\caption{Results for the \textbf{five unimodal and five simple multimodal} functions of the CEC'17 benchmark suite (30D)}
\setlength{\tabcolsep}{6pt}
\renewcommand\arraystretch{1}
\label{30D-1}
\begin{minipage}{1\textwidth} 
    \tiny
    \centering  
  \resizebox{\textwidth}{!}{
\begin{tabular}{ccllllllll}
\hline
\textbf{}                          & \textbf{}     & \textbf{PSO} & \textbf{CLPSO}      & \textbf{OLPSO} & \textbf{PPSO} & \textbf{XPSO} & \textbf{TAPSO}      & \textbf{AWPSO} & \textbf{HPSO}     \\ \hline
\multirow{3}{*}{\textbf{$F_1$}}    & \textbf{Mean} & 1.60E+11 +   & \textbf{1.21E+03 -} & 5.67E+03 =     & 1.84E+09 +    & 6.03E+03 =    & 4.45E+03 =          & 2.42E+10 +     & 4.78E+03          \\
                                   & \textbf{Std}  & 7.98E+10     & \textbf{1.08E+03}   & 5.12E+03       & 4.91E+09      & 5.80E+03      & 3.88E+03            & 2.80E+10       & 4.51E+03          \\
                                   & \textbf{Ranking} & 8            & \textbf{1}          & 4              & 6             & 5             & 2                   & 7              & 3                 \\
\multirow{3}{*}{\textbf{$F_3$}}    & \textbf{Mean} & 1.30E+04 =   & 1.15E+04 +          & 2.32E+05 +     & 4.54E+04 +    & 3.01E+02 -    & \textbf{3.00E+02 -} & 3.28E+03 -     & 7.94E+03          \\
                                   & \textbf{Std}  & 1.97E+04     & 1.86E+03            & 4.03E+04       & 1.80E+04      & 2.53E+00      & \textbf{0.00E+00}   & 6.52E+03       & 3.43E+03          \\
                                   & \textbf{Ranking} & 6            & 5                   & 8              & 7             & 2             & \textbf{1}          & 3              & 4                 \\
\multirow{3}{*}{\textbf{$F_4$}}    & \textbf{Mean} & 1.81E+03 +   & 4.86E+02 -          & 4.92E+02 =     & 5.51E+02 +    & 5.31E+02 +    & \textbf{4.37E+02 -} & 7.65E+02 +     & 4.98E+02          \\
                                   & \textbf{Std}  & 1.12E+03     & 2.54E+01            & 3.28E+00       & 5.30E+01      & 1.39E+01      & \textbf{2.99E+01}   & 2.67E+02       & 1.40E+01          \\
                                   & \textbf{Ranking} & 8            & 2                   & 3              & 6             & 5             & \textbf{1}          & 7              & 4                 \\
\multirow{3}{*}{\textbf{$F_5$}}    & \textbf{Mean} & 6.40E+02 +   & 5.51E+02 +          & 5.95E+02 +     & 7.39E+02 +    & 5.51E+02 +    & 5.49E+02 +          & 5.89E+02 +     & \textbf{5.11E+02} \\
                                   & \textbf{Std}  & 3.04E+01     & 8.51E+00            & 2.26E+01       & 4.88E+01      & 1.54E+01      & 1.15E+01            & 2.12E+01       & \textbf{3.40E+00} \\
                                   & \textbf{Ranking} & 7            & 3                   & 6              & 8             & 4             & 2                   & 5              & \textbf{1}        \\
\multirow{3}{*}{\textbf{$F_6$}}    & \textbf{Mean} & 6.41E+02 +   & 6.19E+02 +          & 6.20E+02 +     & 6.84E+02 +    & 6.04E+02 +    & 6.07E+02 +          & 6.18E+02 +     & \textbf{6.01E+02} \\
                                   & \textbf{Std}  & 1.20E+01     & 4.30E+00            & 5.67E+00       & 1.39E+01      & 3.17E+00      & 4.09E+00            & 5.48E+00       & \textbf{8.89E-01} \\
                                   & \textbf{Ranking} & 7            & 5                   & 6              & 8             & 2             & 3                   & 4              & \textbf{1}        \\
\multirow{3}{*}{\textbf{$F_7$}}    & \textbf{Mean} & 9.61E+02 +   & 7.83E+02 +          & 7.83E+02 +     & 1.36E+03 +    & 8.10E+02 +    & 7.86E+02 +          & 8.27E+02 +     & \textbf{7.43E+02} \\
                                   & \textbf{Std}  & 1.20E+02     & 9.11E+00            & 1.48E+01       & 1.20E+02      & 2.08E+01      & 1.38E+01            & 4.01E+01       & \textbf{3.51E+00} \\
                                   & \textbf{Ranking} & 7            & 3                   & 2              & 8             & 5             & 4                   & 6              & \textbf{1}        \\
\multirow{3}{*}{\textbf{$F_8$}}    & \textbf{Mean} & 9.41E+02 +   & 8.53E+02 +          & 8.90E+02 +     & 9.93E+02 +    & 8.55E+02 +    & 8.56E+02 +          & 8.85E+02 +     & \textbf{8.12E+02} \\
                                   & \textbf{Std}  & 3.13E+01     & 7.60E+00            & 1.52E+01       & 5.09E+01      & 2.74E+01      & 1.71E+01            & 2.21E+01       & \textbf{4.44E+00} \\
                                   & \textbf{Ranking} & 7            & 2                   & 6              & 8             & 3             & 4                   & 5              & \textbf{1}        \\
\multirow{3}{*}{\textbf{$F_9$}}    & \textbf{Mean} & 3.75E+03 +   & 9.22E+02 +          & 1.27E+03 +     & 7.10E+03 +    & 9.11E+02 +    & 9.45E+02 +          & 1.33E+03 +     & \textbf{9.02E+02} \\
                                   & \textbf{Std}  & 1.47E+03     & 2.23E+01            & 4.90E+02       & 1.72E+03      & 1.33E+01      & 4.87E+01            & 4.74E+02       & \textbf{1.48E+00} \\
                                   & \textbf{Ranking} & 7            & 3                   & 5              & 8             & 2             & 4                   & 6              & \textbf{1}        \\
\multirow{3}{*}{\textbf{$F_{10}$}} & \textbf{Mean} & 4.49E+03 +   & 3.84E+03 +          & 5.22E+03 +     & 6.67E+03 +    & 4.60E+03 +    & \textbf{3.48E+03 =} & 4.33E+03 +     & 3.52E+03          \\
                                   & \textbf{Std}  & 5.90E+02     & 3.39E+02            & 5.31E+02       & 1.17E+03      & 1.06E+03      & \textbf{6.50E+02}   & 5.30E+02       & 6.33E+02          \\
                                   & \textbf{Ranking} & 5            & 3                   & 7              & 8             & 6             & \textbf{1}          & 4              & 2                 \\ \hline
\end{tabular}
}
\end{minipage}
\end{table*}

\begin{table*}[]
\centering
\caption{Results for the \textbf{ten hybrid functions} of the CEC'17 benchmark suite (30D)}
\setlength{\tabcolsep}{6pt}
\renewcommand\arraystretch{1}
\label{30D-2}
\begin{minipage}{1\textwidth}  
    \tiny
    \centering  
  \resizebox{\textwidth}{!}{
\begin{tabular}{ccllllllll}
\hline
\textbf{}                          & \textbf{}     & \textbf{PSO} & \textbf{CLPSO}      & \textbf{OLPSO} & \textbf{PPSO} & \textbf{XPSO}       & \textbf{TAPSO}      & \textbf{AWPSO} & \textbf{HPSO}     \\ \hline
\multirow{3}{*}{\textbf{$F_{11}$}} & \textbf{Mean} & 1.71E+03 +   & 1.16E+03 =          & 1.25E+03 +     & 1.38E+03 +    & 1.18E+03 +          & 1.16E+03 =          & 1.34E+03 +     & \textbf{1.15E+03} \\
                                   & \textbf{Std}  & 5.48E+02     & 2.32E+01            & 3.33E+01       & 1.13E+02      & 3.74E+01            & 3.04E+01            & 6.64E+01       & \textbf{3.11E+01} \\
                                   & \textbf{Ranking} & 8            & 3                   & 5              & 7             & 4                   & 2                   & 6              & \textbf{1}        \\
\multirow{3}{*}{\textbf{$F_{12}$}} & \textbf{Mean} & 7.58E+09 +   & 5.66E+05 =          & 2.12E+07 +     & 1.56E+08 +    & 1.22E+06 +          & \textbf{2.83E+04 -} & 4.41E+08 +     & 7.98E+05          \\
                                   & \textbf{Std}  & 9.67E+09     & 3.70E+05            & 4.09E+07       & 2.20E+08      & 2.73E+06            & \textbf{2.26E+04}   & 6.51E+08       & 7.77E+05          \\
                                   & \textbf{Ranking} & 8            & 2                   & 5              & 6             & 4                   & \textbf{1}          & 7              & 3                 \\
\multirow{3}{*}{\textbf{$F_{13}$}} & \textbf{Mean} & 6.36E+09 +   & \textbf{1.20E+04 =} & 1.48E+06 =     & 1.75E+06 +    & 1.92E+04 =          & 2.23E+04 =          & 4.83E+08 +     & 2.48E+04          \\
                                   & \textbf{Std}  & 8.56E+09     & \textbf{4.41E+03}   & 8.00E+06       & 3.63E+06      & 1.95E+04            & 2.00E+04            & 1.93E+09       & 2.00E+04          \\
                                   & \textbf{Ranking} & 8            & \textbf{1}          & 5              & 6             & 2                   & 3                   & 7              & 4                 \\
\multirow{3}{*}{\textbf{$F_{14}$}} & \textbf{Mean} & 4.52E+04 +   & 3.65E+04 +          & 4.95E+04 +     & 1.06E+05 +    & \textbf{1.00E+04 =} & 1.26E+04 -          & 5.49E+04 =     & 1.60E+04          \\
                                   & \textbf{Std}  & 6.38E+04     & 3.85E+04            & 4.00E+04       & 1.06E+05      & \textbf{6.67E+03}   & 2.38E+04            & 8.57E+04       & 1.60E+04          \\
                                   & \textbf{Ranking} & 5            & 4                   & 6              & 8             & \textbf{1}          & 2                   & 7              & 3                 \\
\multirow{3}{*}{\textbf{$F_{15}$}} & \textbf{Mean} & 1.37E+05 +   & \textbf{1.92E+03 -} & 1.37E+04 +     & 1.61E+05 +    & 1.35E+04 +          & 9.17E+03 =          & 4.47E+04 +     & 5.41E+03          \\
                                   & \textbf{Std}  & 9.24E+04     & \textbf{5.28E+02}   & 1.29E+04       & 5.63E+05      & 1.13E+04            & 1.10E+04            & 4.38E+04       & 4.09E+03          \\
                                   & \textbf{Ranking} & 7            & \textbf{1}          & 5              & 8             & 4                   & 3                   & 6              & 2                 \\
\multirow{3}{*}{\textbf{$F_{16}$}} & \textbf{Mean} & 2.82E+03 +   & 2.22E+03 +          & 2.86E+03 +     & 3.64E+03 +    & 2.17E+03 +          & 2.38E+03 +          & 2.54E+03 +     & \textbf{1.92E+03} \\
                                   & \textbf{Std}  & 5.08E+02     & 1.49E+02            & 2.30E+02       & 6.68E+02      & 3.27E+02            & 2.88E+02            & 2.52E+02       & \textbf{1.77E+02} \\
                                   & \textbf{Ranking} & 6            & 3                   & 7              & 8             & 2                   & 4                   & 5              & \textbf{1}        \\
\multirow{3}{*}{\textbf{$F_{17}$}} & \textbf{Mean} & 2.39E+03 +   & \textbf{1.81E+03 =} & 2.22E+03 +     & 2.84E+03 +    & 1.84E+03 =          & 2.03E+03 +          & 2.05E+03 +     & 1.81E+03          \\
                                   & \textbf{Std}  & 3.17E+02     & \textbf{5.96E+01}   & 1.84E+02       & 3.11E+02      & 9.79E+01            & 1.90E+02            & 1.67E+02       & 6.73E+01          \\
                                   & \textbf{Ranking} & 7            & \textbf{1}          & 6              & 8             & 3                   & 4                   & 5              & 2                 \\
\multirow{3}{*}{\textbf{$F_{18}$}} & \textbf{Mean} & 7.14E+05 =   & 1.17E+05 -          & 9.34E+05 +     & 1.75E+06 +    & 1.61E+05 -          & \textbf{6.62E+04 -} & 2.97E+05 =     & 2.84E+05          \\
                                   & \textbf{Std}  & 7.47E+05     & 5.11E+04            & 6.24E+05       & 2.22E+06      & 1.11E+05            & \textbf{9.85E+04}   & 3.84E+05       & 1.65E+05          \\
                                   & \textbf{Ranking} & 6            & 2                   & 7              & 8             & 3                   & \textbf{1}          & 5              & 4                 \\
\multirow{3}{*}{\textbf{$F_{19}$}} & \textbf{Mean} & 3.30E+08 +   & \textbf{3.14E+03 -} & 2.73E+04 +     & 4.70E+06 +    & 2.02E+04 +          & 1.17E+04 =          & 1.56E+06 +     & 6.29E+03          \\
                                   & \textbf{Std}  & 6.23E+08     & \textbf{1.39E+03}   & 3.29E+04       & 1.39E+07      & 1.60E+04            & 1.20E+04            & 4.91E+06       & 4.64E+03          \\
                                   & \textbf{Ranking} & 8            & \textbf{1}          & 5              & 7             & 4                   & 3                   & 6              & 2                 \\
\multirow{3}{*}{\textbf{$F_{20}$}} & \textbf{Mean} & 2.42E+03 +   & 2.21E+03 +          & 2.43E+03 +     & 2.90E+03 +    & 2.16E+03 =          & 2.31E+03 +          & 2.28E+03 +     & \textbf{2.12E+03} \\
                                   & \textbf{Std}  & 1.36E+02     & 6.53E+01            & 1.38E+02       & 2.14E+02      & 7.22E+01            & 1.93E+02            & 1.33E+02       & \textbf{6.95E+01} \\
                                   & \textbf{Ranking} & 6            & 3                   & 7              & 8             & 2                   & 5                   & 4              & \textbf{1}        \\ \hline
\end{tabular}
}
\end{minipage}
\end{table*}

\begin{table*}[]
\centering
\caption{Results for the \textbf{ten composition functions} of the CEC'17 benchmark suite (30D)}
\setlength{\tabcolsep}{6pt}
\renewcommand\arraystretch{1}
\label{30D-3}
\begin{minipage}{1\textwidth} 
    \tiny
    \centering  
  \resizebox{\textwidth}{!}{
\begin{tabular}{ccllllllll}
\hline
\textbf{}                          & \textbf{}     & \textbf{PSO} & \textbf{CLPSO}      & \textbf{OLPSO} & \textbf{PPSO} & \textbf{XPSO}       & \textbf{TAPSO}      & \textbf{AWPSO} & \textbf{HPSO}     \\ \hline
\multirow{3}{*}{\textbf{$F_{21}$}} & \textbf{Mean} & 2.46E+03 +   & 2.36E+03 +          & 2.41E+03 +     & 2.56E+03 +    & 2.35E+03 +          & 2.36E+03 +          & 2.40E+03 +     & \textbf{2.32E+03} \\
                                   & \textbf{Std}  & 3.47E+01     & 6.98E+00            & 1.81E+01       & 5.34E+01      & 1.50E+01            & 2.33E+01            & 2.40E+01       & \textbf{3.35E+00} \\
                                   & \textbf{Ranking} & 7            & 3                   & 6              & 8             & 2                   & 4                   & 5              & \textbf{1}        \\
\multirow{3}{*}{\textbf{$F_{22}$}} & \textbf{Mean} & 5.57E+03 +   & \textbf{2.34E+03 -} & 5.20E+03 +     & 7.38E+03 +    & 2.70E+03 =          & 3.89E+03 +          & 3.70E+03 +     & 2.54E+03          \\
                                   & \textbf{Std}  & 1.27E+03     & \textbf{9.18E+01}   & 1.85E+03       & 1.42E+03      & 1.28E+03            & 1.59E+03            & 1.32E+03       & 6.87E+02          \\
                                   & \textbf{Ranking} & 7            & \textbf{1}          & 6              & 8             & 3                   & 5                   & 4              & 2                 \\
\multirow{3}{*}{\textbf{$F_{23}$}} & \textbf{Mean} & 2.94E+03 +   & 2.71E+03 +          & 2.80E+03 +     & 3.21E+03 +    & 2.71E+03 +          & 2.72E+03 +          & 2.88E+03 +     & \textbf{2.70E+03} \\
                                   & \textbf{Std}  & 5.55E+01     & 9.20E+00            & 1.46E+01       & 1.70E+02      & 1.69E+01            & 1.79E+01            & 8.59E+01       & \textbf{2.02E+01} \\
                                   & \textbf{Ranking} & 7            & 3                   & 5              & 8             & 2                   & 4                   & 6              & \textbf{1}        \\
\multirow{3}{*}{\textbf{$F_{24}$}} & \textbf{Mean} & 3.14E+03 +   & 2.91E+03 +          & 3.03E+03 +     & 3.38E+03 +    & \textbf{2.89E+03 =} & 2.91E+03 +          & 3.05E+03 +     & 2.89E+03          \\
                                   & \textbf{Std}  & 7.43E+01     & 1.82E+01            & 2.46E+01       & 2.00E+02      & \textbf{2.76E+01}   & 2.59E+01            & 8.56E+01       & 2.13E+01          \\
                                   & \textbf{Ranking} & 7            & 4                   & 5              & 8             & \textbf{1}          & 3                   & 6              & 2                 \\
\multirow{3}{*}{\textbf{$F_{25}$}} & \textbf{Mean} & 3.29E+03 +   & 2.89E+03 =          & 2.89E+03 -     & 2.97E+03 +    & 2.90E+03 +          & 2.89E+03 =          & 2.98E+03 +     & 2.89E+03          \\
                                   & \textbf{Std}  & 4.50E+02     & 8.89E-01            & 4.51E-01       & 4.02E+01      & 9.20E+00            & 7.62E+00            & 1.34E+02       & 6.67E+00          \\
                                   & \textbf{Ranking} & 8            & 3                   & 2              & 6             & 5                   & 1                   & 7              & 4                 \\
\multirow{3}{*}{\textbf{$F_{26}$}} & \textbf{Mean} & 6.49E+03 +   & 4.30E+03 +          & 4.98E+03 +     & 7.45E+03 +    & 4.15E+03 +          & 4.23E+03 +          & 4.88E+03 +     & \textbf{3.99E+03} \\
                                   & \textbf{Std}  & 7.39E+02     & 2.14E+02            & 1.82E+02       & 2.16E+03      & 3.12E+02            & 4.64E+02            & 5.63E+02       & \textbf{1.48E+02} \\
                                   & \textbf{Ranking} & 7            & 4                   & 6              & 8             & 2                   & 3                   & 5              & \textbf{1}        \\
\multirow{3}{*}{\textbf{$F_{27}$}} & \textbf{Mean} & 3.35E+03 +   & \textbf{3.22E+03 =} & 3.22E+03 +     & 3.61E+03 +    & 3.24E+03 +          & 3.22E+03 =          & 3.31E+03 +     & 3.22E+03          \\
                                   & \textbf{Std}  & 6.44E+01     & \textbf{5.05E+00}   & 7.78E+00       & 1.47E+02      & 1.66E+01            & 1.02E+01            & 5.64E+01       & 1.41E+01          \\
                                   & \textbf{Ranking} & 7            & \textbf{1}          & 4              & 8             & 5                   & 2                   & 6              & 3                 \\
\multirow{3}{*}{\textbf{$F_{28}$}} & \textbf{Mean} & 4.99E+03 +   & 3.21E+03 -          & 3.68E+03 +     & 3.35E+03 +    & 3.29E+03 +          & \textbf{3.16E+03 -} & 3.42E+03 +     & 3.24E+03          \\
                                   & \textbf{Std}  & 1.22E+03     & 8.81E+00            & 5.08E+02       & 5.87E+01      & 5.29E+01            & \textbf{6.42E+01}   & 1.76E+02       & 5.47E+01          \\
                                   & \textbf{Ranking} & 8            & 2                   & 7              & 5             & 4                   & \textbf{1}          & 6              & 3                 \\
\multirow{3}{*}{\textbf{$F_{29}$}} & \textbf{Mean} & 4.21E+03 +   & 3.50E+03 +          & 3.99E+03 +     & 5.76E+03 +    & 3.60E+03 +          & 3.68E+03 +          & 3.72E+03 +     & \textbf{3.46E+03} \\
                                   & \textbf{Std}  & 3.79E+02     & 5.20E+01            & 1.74E+02       & 7.26E+02      & 1.68E+02            & 1.58E+02            & 2.54E+02       & \textbf{1.04E+02} \\
                                   & \textbf{Ranking} & 7            & 2                   & 6              & 8             & 3                   & 4                   & 5              & \textbf{1}        \\
\multirow{3}{*}{\textbf{$F_{30}$}} & \textbf{Mean} & 1.74E+07 +   & 2.77E+04 +          & 7.28E+05 +     & 3.88E+06 +    & 1.72E+04 +          & \textbf{9.46E+03 -} & 3.38E+06 +     & 1.44E+04          \\
                                   & \textbf{Std}  & 1.94E+07     & 1.55E+04            & 1.31E+06       & 5.64E+06      & 1.22E+04            & \textbf{4.57E+03}   & 7.40E+06       & 1.36E+04          \\
                                   & \textbf{Ranking} & 8            & 4                   & 5              & 7             & 3                   & \textbf{1}          & 6              & 2                 \\ \hline
\end{tabular}
}
\end{minipage}
\end{table*}

\begin{table*}[]
\centering
\caption{Results for the \textbf{five unimodal and five simple multimodal functions} of the CEC'17 benchmark suite (100D)}
\setlength{\tabcolsep}{6pt}
\renewcommand\arraystretch{1}
\label{100D-1}
\begin{minipage}{1\textwidth}  
    \tiny
    \centering  
  \resizebox{\textwidth}{!}{
\begin{tabular}{ccllllllll}
\hline
\textbf{}                          & \textbf{}     & \textbf{PSO} & \textbf{CLPSO}      & \textbf{OLPSO} & \textbf{PPSO} & \textbf{XPSO}       & \textbf{TAPSO}      & \textbf{AWPSO} & \textbf{HPSO}     \\ \hline
\multirow{3}{*}{\textbf{$F_1$}}    & \textbf{Mean} & 1.83E+12 +   & \textbf{4.54E+03 -} & 2.67E+04 -     & 1.86E+10 +    & 3.46E+07 +          & 8.65E+03 -          & 6.94E+11 +     & 1.10E+07          \\
                                   & \textbf{Std}  & 2.83E+11     & \textbf{3.23E+03}   & 2.86E+04       & 2.92E+10      & 1.34E+08            & 1.15E+04            & 1.83E+11       & 3.69E+07          \\
                                   & \textbf{Ranking} & 8            & \textbf{1}          & 3              & 6             & 5                   & 2                   & 7              & 4                 \\
\multirow{3}{*}{\textbf{$F_3$}}    & \textbf{Mean} & 3.24E+05 +   & 1.43E+05 -          & 9.96E+05 +     & 3.66E+05 +    & \textbf{4.32E+04 -} & 4.76E+04 -          & 1.86E+05 =     & 1.86E+05          \\
                                   & \textbf{Std}  & 1.17E+05     & 1.33E+04            & 1.14E+05       & 6.62E+04      & \textbf{9.91E+03}   & 2.26E+04            & 6.48E+04       & 3.04E+04          \\
                                   & \textbf{Ranking} & 6            & 3                   & 8              & 7             & \textbf{1}          & 2                   & 4              & 5                 \\
\multirow{3}{*}{\textbf{$F_4$}}    & \textbf{Mean} & 3.68E+04 +   & 6.73E+02 =          & 6.67E+02 -     & 1.26E+03 +    & 9.79E+02 +          & \textbf{5.67E+02 -} & 8.24E+03 +     & 7.06E+02          \\
                                   & \textbf{Std}  & 1.23E+04     & 5.37E+01            & 1.42E+01       & 4.34E+02      & 5.81E+01            & \textbf{4.28E+01}   & 3.89E+03       & 1.17E+02          \\
                                   & \textbf{Ranking} & 8            & 3                   & 2              & 6             & 5                   & \textbf{1}          & 7              & 4                 \\
\multirow{3}{*}{\textbf{$F_5$}}    & \textbf{Mean} & 1.53E+03 +   & 1.05E+03 +          & 9.98E+02 +     & 1.53E+03 +    & 7.82E+02 +          & 8.53E+02 +          & 1.07E+03 +     & \textbf{6.14E+02} \\
                                   & \textbf{Std}  & 1.13E+02     & 2.47E+01            & 6.11E+01       & 1.08E+02      & 5.52E+01            & 4.99E+01            & 6.45E+01       & \textbf{3.09E+01} \\
                                   & \textbf{Ranking} & 8            & 5                   & 4              & 7             & 2                   & 3                   & 6              & \textbf{1}        \\
\multirow{3}{*}{\textbf{$F_6$}}    & \textbf{Mean} & 6.80E+02 +   & 6.38E+02 +          & 6.43E+02 +     & 7.01E+02 +    & 6.26E+02 +          & 6.33E+02 +          & 6.49E+02 +     & \textbf{6.11E+02} \\
                                   & \textbf{Std}  & 8.68E+00     & 4.24E+00            & 6.91E+00       & 9.40E+00      & 6.06E+00            & 5.36E+00            & 7.65E+00       & \textbf{2.31E+00} \\
                                   & \textbf{Ranking} & 7            & 4                   & 5              & 8             & 2                   & 3                   & 6              & \textbf{1}        \\
\multirow{3}{*}{\textbf{$F_7$}}    & \textbf{Mean} & 4.73E+03 +   & 1.36E+03 +          & 1.08E+03 +     & 4.24E+03 +    & 1.52E+03 +          & 1.29E+03 +          & 1.90E+03 +     & \textbf{9.57E+02} \\
                                   & \textbf{Std}  & 7.40E+02     & 2.51E+01            & 4.42E+01       & 2.87E+02      & 1.63E+02            & 6.97E+01            & 2.62E+02       & \textbf{3.94E+01} \\
                                   & \textbf{Ranking} & 8            & 4                   & 2              & 7             & 5                   & 3                   & 6              & \textbf{1}        \\
\multirow{3}{*}{\textbf{$F_8$}}    & \textbf{Mean} & 1.93E+03 +   & 1.34E+03 +          & 1.30E+03 +     & 1.96E+03 +    & 1.09E+03 +          & 1.15E+03 +          & 1.42E+03 +     & \textbf{9.01E+02} \\
                                   & \textbf{Std}  & 1.45E+02     & 3.02E+01            & 5.43E+01       & 1.36E+02      & 5.39E+01            & 5.55E+01            & 8.38E+01       & \textbf{2.43E+01} \\
                                   & \textbf{Ranking} & 7            & 5                   & 4              & 8             & 2                   & 3                   & 6              & \textbf{1}        \\
\multirow{3}{*}{\textbf{$F_9$}}    & \textbf{Mean} & 3.68E+04 +   & 1.50E+04 +          & 7.58E+03 +     & 4.89E+04 +    & 3.23E+03 +          & 7.04E+03 +          & 2.82E+04 +     & \textbf{1.23E+03} \\
                                   & \textbf{Std}  & 1.06E+04     & 2.86E+03            & 2.25E+03       & 7.97E+03      & 1.35E+03            & 2.82E+03            & 1.85E+04       & \textbf{2.11E+02} \\
                                   & \textbf{Ranking} & 7            & 5                   & 4              & 8             & 2                   & 3                   & 6              & \textbf{1}        \\
\multirow{3}{*}{\textbf{$F_{10}$}} & \textbf{Mean} & 1.74E+04 +   & 2.14E+04 +          & 1.91E+04 +     & 2.89E+04 +    & 2.06E+04 +          & \textbf{1.31E+04 -} & 1.56E+04 =     & 1.54E+04          \\
                                   & \textbf{Std}  & 1.60E+03     & 6.18E+02            & 1.35E+03       & 1.74E+03      & 6.01E+03            & \textbf{1.17E+03}   & 1.13E+03       & 2.19E+03          \\
                                   & \textbf{Ranking} & 4            & 7                   & 5              & 8             & 6                   & \textbf{1}          & 3              & 2                 \\ \hline
\end{tabular}
}
\end{minipage}
\end{table*}

\begin{table*}[]
\centering
\caption{Results for the \textbf{ten hybrid functions} of the CEC'17 benchmark suite (100D)}
\setlength{\tabcolsep}{6pt}
\renewcommand\arraystretch{1}
\label{100D-2}
\begin{minipage}{1\textwidth}  
    \tiny
    \centering  
  \resizebox{\textwidth}{!}{
\begin{tabular}{ccllllllll}
\hline
\textbf{}                          & \textbf{}     & \textbf{PSO} & \textbf{CLPSO}      & \textbf{OLPSO} & \textbf{PPSO} & \textbf{XPSO} & \textbf{TAPSO}      & \textbf{AWPSO} & \textbf{HPSO}     \\ \hline
\multirow{3}{*}{\textbf{$F_{11}$}} & \textbf{Mean} & 5.66E+04 +   & 1.75E+03 -          & 5.91E+04 +     & 2.04E+04 +    & 3.08E+03 +    & \textbf{1.50E+03 -} & 1.86E+04 +     & 2.54E+03          \\
                                   & \textbf{Std}  & 4.13E+04     & 1.06E+02            & 3.71E+04       & 1.70E+04      & 4.28E+02      & \textbf{4.96E+02}   & 1.47E+04       & 2.29E+02          \\
                                   & \textbf{Ranking} & 7            & 2                   & 8              & 6             & 4             & \textbf{1}          & 5              & 3                 \\
\multirow{3}{*}{\textbf{$F_{12}$}} & \textbf{Mean} & 5.44E+11 +   & 2.78E+06 -          & 7.29E+08 +     & 1.21E+10 +    & 1.84E+08 +    & \textbf{5.90E+05 -} & 1.25E+11 +     & 1.93E+07          \\
                                   & \textbf{Std}  & 2.21E+11     & 9.51E+05            & 1.07E+09       & 1.60E+10      & 4.21E+08      & \textbf{3.95E+05}   & 5.66E+10       & 1.47E+07          \\
                                   & \textbf{Ranking} & 8            & 2                   & 5              & 6             & 4             & \textbf{1}          & 7              & 3                 \\
\multirow{3}{*}{\textbf{$F_{13}$}} & \textbf{Mean} & 1.12E+11 +   & \textbf{5.67E+03 -} & 6.31E+06 +     & 1.15E+09 +    & 4.56E+06 +    & 5.79E+03 -          & 2.48E+10 +     & 2.30E+04          \\
                                   & \textbf{Std}  & 5.22E+10     & \textbf{2.20E+03}   & 2.39E+07       & 3.90E+09      & 2.30E+07      & 5.52E+03            & 1.62E+10       & 2.58E+04          \\
                                   & \textbf{Ranking} & 8            & \textbf{1}          & 5              & 6             & 4             & 2                   & 7              & 3                 \\
\multirow{3}{*}{\textbf{$F_{14}$}} & \textbf{Mean} & 1.74E+07 +   & 9.22E+05 =          & 1.37E+07 +     & 4.43E+06 +    & 2.13E+06 =    & \textbf{8.87E+04 -} & 3.24E+06 =     & 1.53E+06          \\
                                   & \textbf{Std}  & 2.61E+07     & 2.48E+05            & 6.70E+06       & 2.23E+06      & 2.86E+06      & \textbf{6.96E+04}   & 4.80E+06       & 1.26E+06          \\
                                   & \textbf{Ranking} & 8            & 2                   & 7              & 6             & 4             & \textbf{1}          & 5              & 3                 \\
\multirow{3}{*}{\textbf{$F_{15}$}} & \textbf{Mean} & 3.62E+10 +   & \textbf{2.46E+03 -} & 2.39E+04 +     & 2.60E+07 +    & 1.46E+06 +    & 5.00E+03 =          & 8.33E+09 +     & 4.53E+03          \\
                                   & \textbf{Std}  & 1.88E+10     & \textbf{6.58E+02}   & 1.62E+04       & 1.01E+08      & 5.46E+06      & 3.52E+03            & 8.36E+09       & 3.58E+03          \\
                                   & \textbf{Ranking} & 8            & \textbf{1}          & 4              & 6             & 5             & 3                   & 7              & 2                 \\
\multirow{3}{*}{\textbf{$F_{16}$}} & \textbf{Mean} & 9.37E+03 +   & 4.87E+03 +          & 7.72E+03 +     & 1.28E+04 +    & 5.45E+03 +    & 5.29E+03 +          & 6.41E+03 +     & \textbf{4.10E+03} \\
                                   & \textbf{Std}  & 1.10E+03     & 3.95E+02            & 5.70E+02       & 4.35E+03      & 7.53E+02      & 5.48E+02            & 7.77E+02       & \textbf{6.14E+02} \\
                                   & \textbf{Ranking} & 7            & 2                   & 6              & 8             & 4             & 3                   & 5              & \textbf{1}        \\
\multirow{3}{*}{\textbf{$F_{17}$}} & \textbf{Mean} & 5.11E+04 +   & 4.08E+03 +          & 6.77E+03 +     & 2.62E+04 +    & 4.42E+03 +    & 4.30E+03 +          & 6.61E+03 +     & \textbf{3.78E+03} \\
                                   & \textbf{Std}  & 1.09E+05     & 1.84E+02            & 4.76E+02       & 2.67E+04      & 6.19E+02      & 5.52E+02            & 9.95E+02       & \textbf{5.09E+02} \\
                                   & \textbf{Ranking} & 8            & 2                   & 6              & 7             & 4             & 3                   & 5              & \textbf{1}        \\
\multirow{3}{*}{\textbf{$F_{18}$}} & \textbf{Mean} & 1.53E+07 +   & 1.78E+06 =          & 2.03E+07 +     & 7.35E+06 +    & 4.45E+06 =    & \textbf{1.26E+05 -} & 4.86E+06 =     & 2.36E+06          \\
                                   & \textbf{Std}  & 1.61E+07     & 4.74E+05            & 9.32E+06       & 4.71E+06      & 4.69E+06      & \textbf{7.31E+04}   & 7.57E+06       & 2.26E+06          \\
                                   & \textbf{Ranking} & 7            & 2                   & 8              & 6             & 4             & \textbf{1}          & 5              & 3                 \\
\multirow{3}{*}{\textbf{$F_{19}$}} & \textbf{Mean} & 3.45E+10 +   & \textbf{3.12E+03 -} & 4.41E+06 +     & 1.27E+09 +    & 8.02E+05 =    & 6.97E+03 =          & 6.53E+09 +     & 5.65E+03          \\
                                   & \textbf{Std}  & 2.49E+10     & \textbf{9.63E+02}   & 1.76E+07       & 2.52E+09      & 3.65E+06      & 5.98E+03            & 7.29E+09       & 3.98E+03          \\
                                   & \textbf{Ranking} & 8            & \textbf{1}          & 5              & 6             & 4             & 3                   & 7              & 2                 \\
\multirow{3}{*}{\textbf{$F_{20}$}} & \textbf{Mean} & 5.09E+03 +   & 4.39E+03 +          & 5.66E+03 +     & 6.86E+03 +    & 4.76E+03 +    & 4.52E+03 +          & 4.87E+03 +     & \textbf{3.78E+03} \\
                                   & \textbf{Std}  & 4.81E+02     & 2.96E+02            & 4.38E+02       & 6.27E+02      & 9.18E+02      & 5.38E+02            & 4.47E+02       & \textbf{5.49E+02} \\
                                   & \textbf{Ranking} & 6            & 2                   & 7              & 8             & 4             & 3                   & 5              & \textbf{1}        \\ \hline
\end{tabular}
}
\end{minipage}
\end{table*}

\begin{table*}[]
\centering
\caption{Results for the \textbf{ten composition functions} of the CEC'17 benchmark suite (100D)}
\setlength{\tabcolsep}{6pt}
\renewcommand\arraystretch{1}
\label{100D-3}
\begin{minipage}{1\textwidth}  
    \tiny
    \centering 
  \resizebox{\textwidth}{!}{
\begin{tabular}{ccllllllll}
\hline
\textbf{}                          & \textbf{}     & \textbf{PSO} & \textbf{CLPSO}      & \textbf{OLPSO} & \textbf{PPSO} & \textbf{XPSO} & \textbf{TAPSO}      & \textbf{AWPSO} & \textbf{HPSO}     \\ \hline
\multirow{3}{*}{\textbf{$F_{21}$}} & \textbf{Mean} & 3.63E+03 +   & 2.86E+03 +          & 2.83E+03 +     & 3.94E+03 +    & 2.65E+03 +    & 2.71E+03 +          & 3.21E+03 +     & \textbf{2.51E+03} \\
                                   & \textbf{Std}  & 1.55E+02     & 3.18E+01            & 4.69E+01       & 2.14E+02      & 9.70E+01      & 5.21E+01            & 1.37E+02       & \textbf{4.80E+01} \\
                                   & \textbf{Ranking} & 7            & 5                   & 4              & 8             & 2             & 3                   & 6              & \textbf{1}        \\
\multirow{3}{*}{\textbf{$F_{22}$}} & \textbf{Mean} & 1.94E+04 +   & 2.40E+04 +          & 2.19E+04 +     & 3.14E+04 +    & 2.14E+04 +    & \textbf{1.55E+04 -} & 1.83E+04 +     & 1.67E+04          \\
                                   & \textbf{Std}  & 1.55E+03     & 6.77E+02            & 1.14E+03       & 2.83E+03      & 5.99E+03      & \textbf{8.83E+02}   & 1.21E+03       & 2.53E+03          \\
                                   & \textbf{Ranking} & 4            & 7                   & 6              & 8             & 5             & \textbf{1}          & 3              & 2                 \\
\multirow{3}{*}{\textbf{$F_{23}$}} & \textbf{Mean} & 4.86E+03 +   & 3.14E+03 -          & 3.26E+03 =     & 5.74E+03 +    & 3.18E+03 -    & \textbf{3.08E+03 -} & 4.50E+03 +     & 3.26E+03          \\
                                   & \textbf{Std}  & 2.87E+02     & 1.78E+01            & 4.53E+01       & 3.44E+02      & 7.09E+01      & \textbf{3.80E+01}   & 2.52E+02       & 1.20E+02          \\
                                   & \textbf{Ranking} & 7            & 2                   & 4              & 8             & 3             & \textbf{1}          & 6              & 5                 \\
\multirow{3}{*}{\textbf{$F_{24}$}} & \textbf{Mean} & 6.46E+03 +   & 3.75E+03 -          & 3.97E+03 =     & 8.49E+03 +    & 3.77E+03 -    & \textbf{3.73E+03 -} & 5.84E+03 +     & 4.06E+03          \\
                                   & \textbf{Std}  & 4.54E+02     & 2.83E+01            & 5.03E+01       & 1.49E+03      & 1.11E+02      & \textbf{5.78E+01}   & 4.92E+02       & 2.21E+02          \\
                                   & \textbf{Ranking} & 7            & 2                   & 4              & 8             & 3             & \textbf{1}          & 6              & 5                 \\
\multirow{3}{*}{\textbf{$F_{25}$}} & \textbf{Mean} & 2.11E+04 +   & 3.37E+03 =          & 3.41E+03 +     & 3.71E+03 +    & 3.72E+03 +    & \textbf{3.27E+03 -} & 6.52E+03 +     & 3.36E+03          \\
                                   & \textbf{Std}  & 7.82E+03     & 4.02E+01            & 3.11E+01       & 1.16E+02      & 1.14E+02      & \textbf{6.01E+01}   & 1.78E+03       & 6.42E+01          \\
                                   & \textbf{Ranking} & 8            & 3                   & 4              & 5             & 6             & \textbf{1}          & 7              & 2                 \\
\multirow{3}{*}{\textbf{$F_{26}$}} & \textbf{Mean} & 3.63E+04 +   & 1.10E+04 +          & 1.40E+04 +     & 3.65E+04 +    & 1.06E+04 =    & 1.07E+04 =          & 2.19E+04 +     & \textbf{1.02E+04} \\
                                   & \textbf{Std}  & 3.57E+03     & 3.91E+02            & 5.46E+02       & 4.66E+03      & 9.65E+02      & 8.62E+02            & 4.19E+03       & \textbf{2.00E+03} \\
                                   & \textbf{Ranking} & 7            & 4                   & 5              & 8             & 2             & 3                   & 6              & \textbf{1}        \\
\multirow{3}{*}{\textbf{$F_{27}$}} & \textbf{Mean} & 5.37E+03 +   & \textbf{3.46E+03 =} & 3.63E+03 +     & 6.71E+03 +    & 3.72E+03 +    & 3.51E+03 =          & 4.28E+03 +     & 3.48E+03          \\
                                   & \textbf{Std}  & 5.31E+02     & \textbf{2.59E+01}   & 7.23E+01       & 1.36E+03      & 1.18E+02      & 7.44E+01            & 3.59E+02       & 6.32E+01          \\
                                   & \textbf{Ranking} & 7            & \textbf{1}          & 4              & 8             & 5             & 3                   & 6              & 2                 \\
\multirow{3}{*}{\textbf{$F_{28}$}} & \textbf{Mean} & 2.57E+04 +   & 3.47E+03 =          & 1.60E+04 +     & 4.22E+03 +    & 4.19E+03 +    & \textbf{3.35E+03 -} & 1.32E+04 +     & 3.55E+03          \\
                                   & \textbf{Std}  & 3.39E+03     & 3.53E+01            & 2.83E+02       & 6.57E+02      & 2.33E+03      & \textbf{3.42E+01}   & 3.25E+03       & 4.08E+02          \\
                                   & \textbf{Ranking} & 8            & 2                   & 7              & 5             & 4             & \textbf{1}          & 6              & 3                 \\
\multirow{3}{*}{\textbf{$F_{29}$}} & \textbf{Mean} & 1.35E+04 +   & 6.29E+03 +          & 8.18E+03 +     & 2.36E+04 +    & 6.73E+03 +    & 6.14E+03 +          & 7.96E+03 +     & \textbf{5.56E+03} \\
                                   & \textbf{Std}  & 5.72E+03     & 2.95E+02            & 4.45E+02       & 8.64E+03      & 6.05E+02      & 4.54E+02            & 7.31E+02       & \textbf{4.78E+02} \\
                                   & \textbf{Ranking} & 7            & 3                   & 6              & 8             & 4             & 2                   & 5              & \textbf{1}        \\
\multirow{3}{*}{\textbf{$F_{30}$}} & \textbf{Mean} & 6.47E+10 +   & 3.95E+04 +          & 2.18E+07 +     & 6.62E+08 +    & 1.15E+05 +    & \textbf{9.82E+03 -} & 1.74E+10 +     & 2.20E+04          \\
                                   & \textbf{Std}  & 3.48E+10     & 2.78E+04            & 2.57E+07       & 1.65E+09      & 1.58E+05      & \textbf{3.92E+03}   & 1.83E+10       & 2.13E+04          \\
                                   & \textbf{Ranking} & 8            & 3                   & 5              & 6             & 4             & \textbf{1}          & 7              & 2                 \\ \hline
\end{tabular}
}
\end{minipage}
\end{table*}

\end{document}